\documentclass{article}

\usepackage{arxiv}

\usepackage[utf8]{inputenc} 
\usepackage[T1]{fontenc}    
\usepackage{hyperref}       
\usepackage{url}            
\usepackage{booktabs}       
\usepackage{amsfonts}       
\usepackage{nicefrac}       
\usepackage{microtype}      
\usepackage{lipsum}
\usepackage{graphicx}
\graphicspath{ {./images/} }

\usepackage{moreverb,url}
\usepackage{subcaption}
\usepackage[justification=centering]{caption}
\usepackage{mathtools}
\usepackage{amsmath,amsfonts,amssymb}
\usepackage{soul}

\usepackage{enumitem, hyperref}
\makeatletter
\def\namedlabel#1#2{\begingroup
    #2%
    \def\@currentlabel{#2}%
    \phantomsection\label{#1}\endgroup
}
\makeatother

\usepackage[]{acronym}

\title{Wobble control of a pendulum actuated spherical robot}

\author{
 Animesh Singhal \\
  Systems and Control Engineering Dept.\\
  Indian Institute of Technology Bombay\\
  Mumbai, India\\
  \texttt{animeshsinghal.iitb@gmail.com} \\ 
   \And
Sahil Modi  \\
  General Electric Research\\
  Bangalore, India\\
  \texttt{sdsahil12@gmail.com} \\
  \hspace{4.2cm}
  \And
 \hspace{0.7cm}Abhishek Gupta \\
  \hspace{0.7cm}Mechanical Engineering Dept.\\
  \hspace{0.7cm}Indian Institute of Technology Bombay\\
  \hspace{0.7cm}Mumbai, India \\
  \hspace{0.7cm}\texttt{abhi.gupta@iitb.ac.in} \\
  \And
 \hspace{0.1cm}Leena Vachhani \\
  Systems and Control Engineering Dept.\\
  Indian Institute of Technology Bombay\\
  Mumbai, India \\
  \texttt{leena.vachhani@iitb.ac.in} \\  
}

\begin{document}
\maketitle
\begin{abstract}
Spherical robots can conduct surveillance in hostile, cluttered environments without being damaged, as their protective shell can safely house sensors such as cameras. However, lateral oscillations, also known as wobble, occur when these sphere-shaped robots operate at low speeds, leading to shaky camera feedback. These oscillations in a pendulum-actuated spherical robot are caused by the coupling between the forward and steering motions due to nonholonomic constraints. Designing a controller to limit wobbling in these robots is challenging due to their underactuated nature. We propose a model-based controller to navigate a pendulum-actuated spherical robot using wobble-free turning maneuvers consisting of circular arcs and straight lines. The model is developed using Lagrange-D'Alembert equations and accounts for the coupled forward and steering motions. The model is further analyzed to derive expressions for radius of curvature, precession rate, wobble amplitude, and wobble frequency during circular motions. Finally, we design an input-output feedback linearization-based controller to control the robot's heading direction and wobble. Overall, the proposed controller enables a teleoperator to command a specific forward velocity and pendulum angle as per the desired turning radius while limiting the robot's lateral oscillations to enhance the quality of camera feedback.
\end{abstract}


\section*{Nomenclature}
\renewcommand{\baselinestretch}{1}\normalsize
\renewcommand{\aclabelfont}[1]{\textsc{\acsfont{#1}}}
\begin{acronym}[long]

\acro{}[$T_s$]{Rolling torque that rotates the hull with respect to the yoke resulting in a forward motion of the robot}
\acro{}[$T_p$]{Pendulum torque that rotates the pendulum with respect to the yoke resulting in sideways motion of the robot}
\acro{}[$r_p$]{Distance between the pendulum's centre of mass and hull's geometric centre}
\acro{}[$r_h$]{Radius of the sphere}
\acro{}[$r_y$]{Distance between the yoke's centre of mass and hull's geometric centre}
\acro{}[$\mathbf{G}$]{Global inertial frame fixed to the ground}
\acro{}[$\mathbf{Y}$]{Frame attached to the yoke}
\acro{}[$\mathbf{P}$]{Frame attached to the pendulum}
\acro{}[$\mathbf{H}$]{Frame attached to the hull}
\acro{}[$\phi$]{Heading angle of the robot measured with respect to the $X$-axis of global frame ${\mathbf{G}}$. This angle characterises the precession of the robot.}
\acro{}[$\theta$]{Lean angle of the robot perpendicular to the heading direction. This angle characterises the lateral oscillations or the wobbling of the robot}
\acro{}[$\psi$]{Forward spin angle of the robot responsible for moving it forward or backwards. This angle characterises the forward rolling of the robot.}
\acro{}[$\beta$]{Pendulum angle relative to the yoke}
\acro{}[${{^{\mathbf{A}}}{\mathbf{R}}_{\mathbf{B}}}$]{Rotation matrix that maps vectors from frame \textbf{B} to frame \textbf{A}}
\acro{}[$\vec{{^{\mathbf{{\mathbf{A}}}}}\omega_{\mathbf{A}}}$]{The angular velocity vector of frame \textbf{A} expressed in frame \textbf{A}}
\acro{}[$\vec{{}^{{\mathbf{A}}}_{{\mathbf{B}}}r_{{\mathbf{P}}}}$]{Position vector of a point \textbf{P} starting from the origin of coordinate frame \textbf{B} expressed in frame \textbf{A}}
\acro{}[${\mathbf{Y}}_c$]{Position vector corresponding to the center of mass of yoke}
\acro{}[${\mathbf{H}}_c$]{Position vector corresponding to the center of mass of hull}
\acro{}[${\mathbf{P}}_c$]{Position vector corresponding to the pendulum's centre of mass}
\acro{}[${\mathbf{P}}_o$]{Position vector corresponding to the origin of pendulum frame}
\acro{}[$X, Z$]{X and Z co-ordinates of the hull centre along the global frame ${\mathbf{G}}(OXYZ)$}
\acro{}[$\vec{{^{\mathbf{A}}}v_{\mathbf{P}}}$]{Linear velocity vector of point \textbf{P} expressed in frame \textbf{A}}
\acro{}[$K$]{Kinetic energy of the system}
\acro{}[$V$]{Potential energy of the system}
\acro{}[$m_B$]{mass of a body B}
\acro{}[${^{\mathbf{B}}}I_{\mathbf{B}}$]{The mass moment of inertia matrix of a body B calculated in its own frame of reference}
\acro{}[$\lambda_i$]{Lagrange multipliers}
\acro{}[$L$]{Lagrangian}
\acro{}[${\mathbf{x}}$]{State vector representing robot's motion}
\acro{}[$u$]{Control input}
\acro{}[$q$]{Generalized coordinates}
\acro{}[$Q$]{Generalized forces}
\acro{}[$A$]{Amplitude of lateral oscillations}
\acro{}[$\omega$]{Frequency of lateral oscillations}
\acro{}[$\rho$]{Radius of curvature for a spherical robot moving in a circular trajectory with pure rolling}
\acro{}[$y$]{Control output}
\acro{}[$K_{p,\dot{\psi}}$]{Speed control gain}
\acro{}[$K_{p,\beta}$]{Proportional component of pendulum control gain}
\acro{}[$K_{d,\beta}$]{Derivative component of pendulum control gain}
\acro{}[$K_{p,\dot{\theta}}$]{Wobble control gain}
\acro{}[$T_{p,\dot{\theta}}$]{Wobble control component of pendulum torque}
\acro{}[$T_{p,\beta}$]{Pendulum control component of pendulum torque}
\acro{}[$\dot{\theta}_{des}$]{Desired value of the rate of change of lean angle}
\acro{}[$\dot{\psi}_{des}$]{Desired value of forward speed}
\acro{}[$\beta_{des}$]{Desired value of the pendulum's angle}
\acro{}[$\gamma, \delta$]{Coefficients of a linear combination}

\end{acronym}
\renewcommand{\baselinestretch}{1}\normalsize

\section{Introduction}

Spherical robots can shield all electro-mechanical components within their spherical shell from harsh environments and external impacts. Due to their ball shape, spherical robots have the remarkable ability to rebound from collisions with obstacles and avoid becoming wedged in corners. These robots cannot topple over as they can regain their equilibrium after a disturbance due to their spherical shape and heavy pendulum. This ability is beneficial in situations such as falling from low heights or being struck during an operation. Hence, spherical robots are suitable for a wide range of applications, including surveillance, reconnaissance, hazardous environment assessment, search and rescue, and planetary exploration \cite{application1}--\cite{application4}. The driving mechanisms determine the dexterity of a spherical robot based on its capability to roll in multiple directions. From the wheel drive \cite{509415} to the pendulum actuated drives \cite{mahboubi2013design}--\cite{PendWithSphericalJoint}, a variety of such mechanisms have been developed \cite{amphibious}--\cite{tomik2012design}. The majority of active drive designs are based on three fundamental physics principles \cite{physicsOfMotion} for propelling a spherical robot: barycenter offset, outer-shell deformation, and conservation of angular momentum.

Several modeling approaches for representing the dynamics of spherical robots have been proposed in the literature. First-order mathematical models of spherical robots are based on the rolling constraint principle and conservation of angular momentum \cite{897794,844763}. The dynamics of a sphere rolling on a smooth surface have been modeled using the Lagrangian method and Euler angles \cite{rosen,ambloch}. Other studies \cite{ThesisWithSimRobot} have used the Gibbs-Appell equation, Kane equation, and Boltzmann-Hamel equation to model spherical robots.

\begin{figure}[t!]
	\begin{subfigure}{0.5\textwidth}
		\centering
		\includegraphics[width=0.81\linewidth]{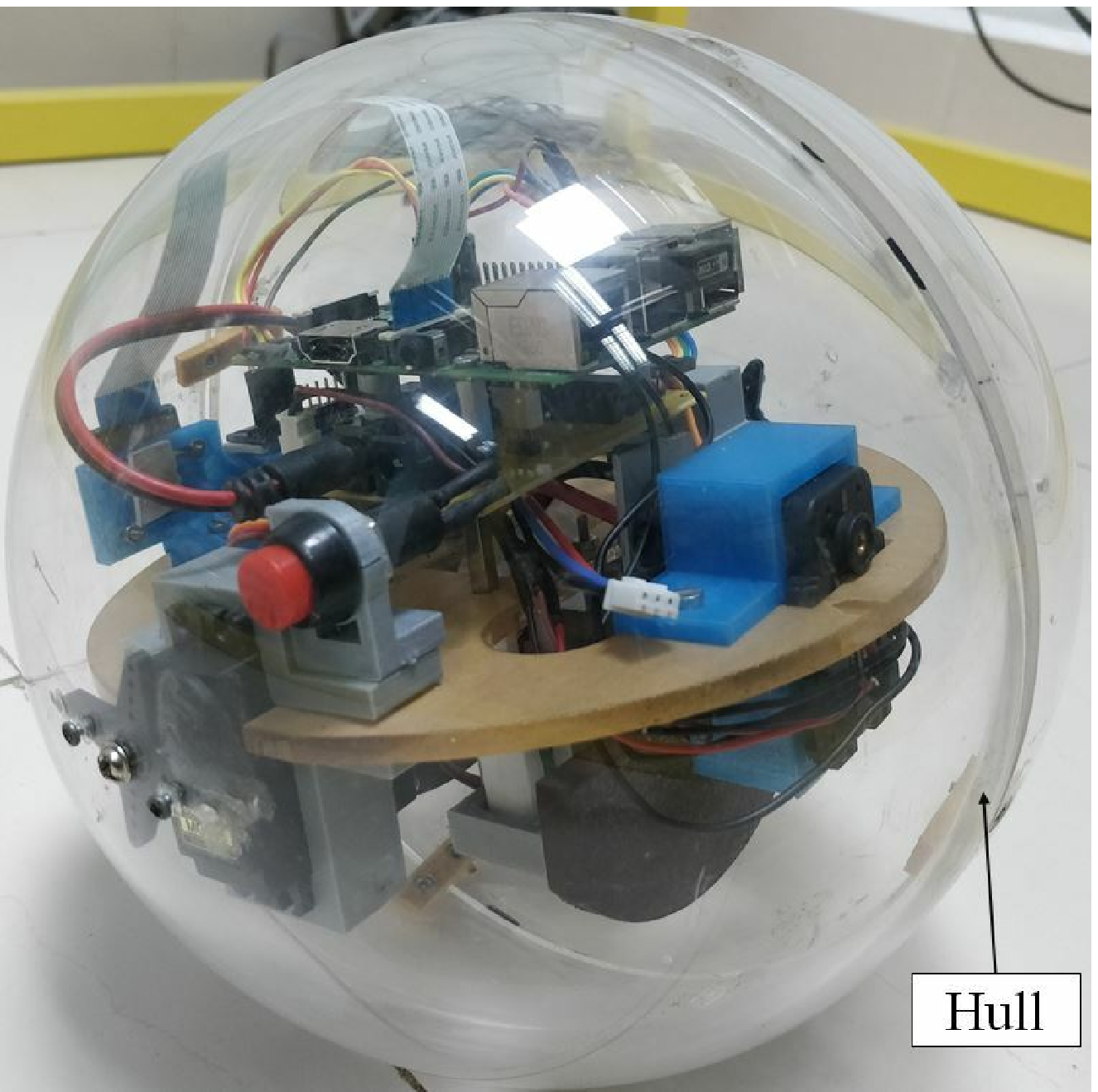}
		\label{fig:bot1}
		\caption{Assembled robot}
	\end{subfigure}
	\begin{subfigure}{0.5\textwidth}
        \centering
        \includegraphics[width=\linewidth]{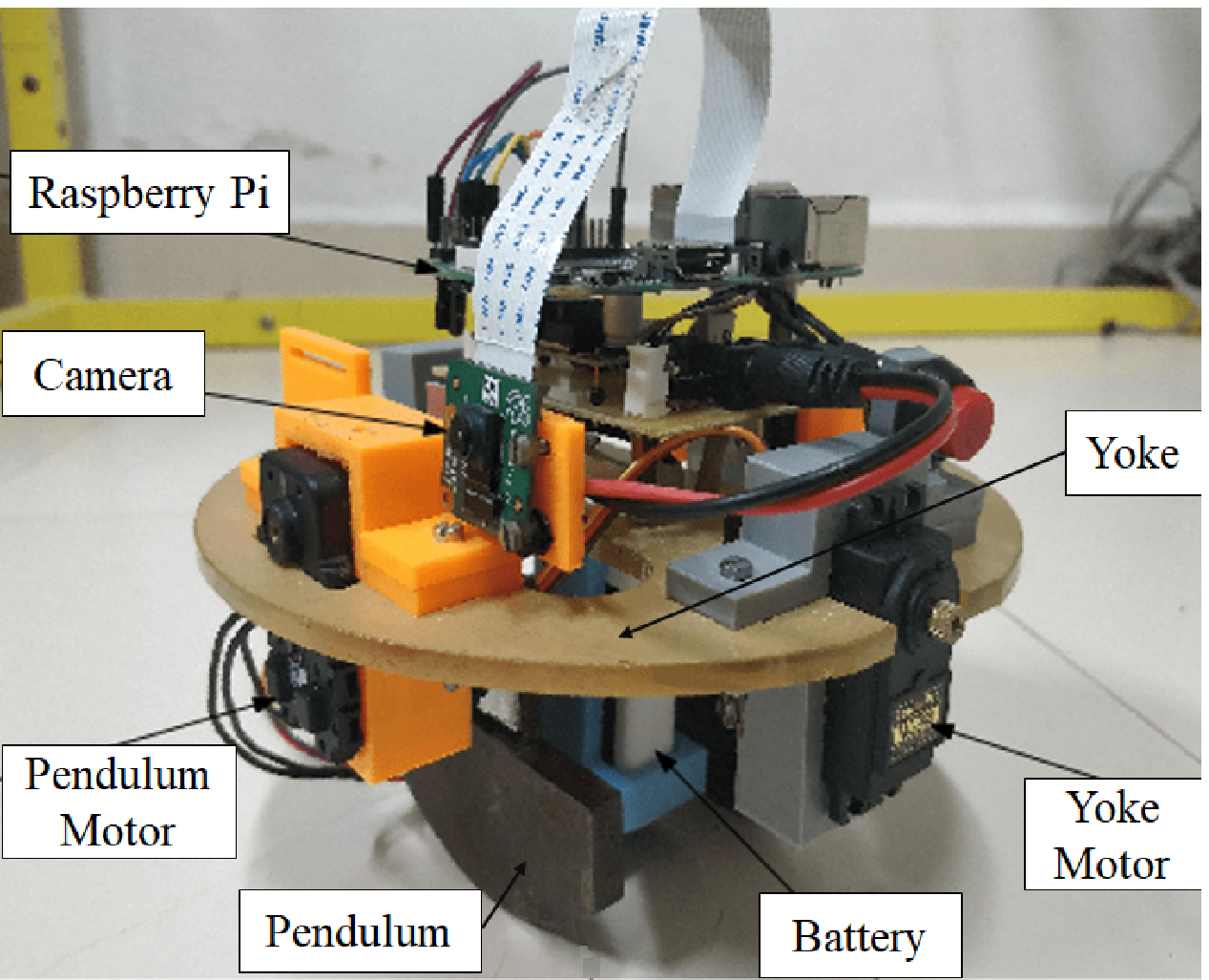}
        \caption{Components of the spherical robot}
    \end{subfigure}
    \caption{The spherical robot \label{fig:robot}}
\end{figure}

This work discusses a spherical robot with a yoke-pendulum design, as depicted in figure \ref{fig:robot}. The robot comprises a spherical shell known as a hull, a platform known as a yoke, and a pendulum. This spherical robot's yoke is an internal platform that can accommodate cameras, various sensors, and the necessary electronics. Two motors drive the robot to generate a forward and steering motion. One of the motors rotates the hull at the required speed, allowing the robot to move forward and backward. The second motor rotates the pendulum with respect to the yoke. This motion perturbs the robot's center of mass to provide steering motion. The pendulum can only swing in the lower hemisphere because the upper hemisphere of the robot contains numerous components, including sensors mounted on the yoke.

Several studies have investigated the dynamics of the sphere with pendulum-based actuation and no-slip conditions. Euler Lagrangian-based dynamic models of just the decoupled forward-driving motion \cite{tomik2012design,liu2008family} exclude the turning motion. Circular motion trajectories can be simulated with a model based on gyroscopic precession \cite{ylikorpi2014gyroscopic}. Other attempts have been made to simultaneously model the forward and steering motions of spherical robots using Euler-Lagrange-based decoupled modeling \cite{decoupled,decoupledModelController}. However, these models have ignored the dynamic interaction between rotations along the lateral and longitudinal axes. In most of these works, mathematical modeling is often followed by a demonstration of straight lines and circular motion executed by spherical bots in simulation \cite{897794, decoupled, cai2012path}. Because of the decoupled modeling approach, the models presented in these works give a bird's-eye view of the motion and do not capture the wobbly nature of the robot observed in practice. As a result, it is worthwhile to investigate the small amplitude lateral oscillations that spherical robots exhibit when moving at low speeds along simple paths such as a straight line or a circle. 

Various experimental results have highlighted the wobbly nature of pendulum-based spherical robots for different pendulum angles \cite{ThesisWithSimRobot,aca_paper,8794742}. Froberg and Smolic \cite{bowling} note the robot's wobbly behavior and suggest using a PID regulator to reverse the robot's incorrect tipping with the pendulum. Schroll \cite{ThesisShowingWobble} modeled the wobbly behavior of a spherical robot driven by a two-degrees-of-freedom pendulum that controls both steering and forward motion. This work demonstrates that when a forward-moving spherical robot is steered by tilting the pendulum at a fixed angle, it moves in a wobbly circle with lateral oscillations and a radius of curvature that oscillates as the robot wobbles into and out of its curved trajectory. Such oscillations have been generally neglected in the literature due to their relatively smaller amplitude. These oscillations become crucial when a robot houses a camera/sensors to capture its surroundings. Specifically, we are interested in the robot's lateral oscillations or sideways fluctuations (perpendicular to the heading direction) as it leads to deviations in the robot's trajectory and shaky video feedback from the mounted camera on the robot's yoke. 

The difficulty of spherical robot's path planning and feedback control problems comes from its non-holonomic, under-actuated, and non-chained properties \cite{controlChallenges1}--\cite{controlChallenges3}. Several innovative attempts have been made in this regard. A  Recurrent Neural Network controller  \cite{RecurrentNN} that accounts for unknown uncertainties and control input saturation has been designed to control the motion of a spherical robot. Another controller has been designed based on Lyapunov's direct method, and neurodynamic technique \cite{neurodynamic} for the two-state trajectory tracking problem of a spherical robot. An alternative controller design for tracking trajectories is based on the back-stepping technique \cite{backStepping}. Two control inputs were used to drive a pendulum-actuated spherical robot to control its heading, and forward speed by Hogan and Forbes \cite{headingAndVelocityControl}. However, this work does not focus on limiting the robot's lateral oscillations. 

Stabilizing the lateral oscillations or wobbling of spherical robots helps get more sharp camera feedback and subsequently enables greater navigational autonomy \cite{navigationalAutonomy} through sensor feedback. Since spherical robots are typically underactuated, the robot's controller can only control a limited number of outputs. Different combinations of control outputs have been used in the past to reduce oscillations when controlling such robots. One of the works \cite{pendControlLaw} chooses the control outputs in such a way that the robot advances with a  constant speed and minimal pendulum oscillations. Despite this controller design, the robot still exhibits small amplitude oscillations, and the pendulum's motion within the spherical shell is unrestricted, which is impractical due to space constraints. A few other attempts are to control forward speed and lean angle, which is responsible for lateral oscillations, using sliding mode controllers \cite{ThesisWithSimRobot, slidingMode1}. However, the limitation of sliding mode control is its tendency to oscillate due to switching around the sliding surface and its sensitivity to controller parameters. Other studies have attempted to stabilize the lean angle by proposing various model-based controller approaches, such as proportional-integral (PI) and Linear Quadratic Regulator (LQR) controllers. The effectiveness of these model-based controller designs is contingent upon the precision of the employed model. The use of a decoupled model \cite{decoupledModelController, LQR-PI-decoupled} and external noise to model the lateral oscillations \cite{systemIdentification} has led to inaccurate modeling of the lateral oscillations. Therefore, it is essential to properly model these oscillations and choose a set of control objectives that can control the speed and direction of the robot's motion while limiting wobbling and maintaining a stable pendulum swing within the available space.

This work makes the following contributions, which are organized into separate sections: 

\begin{itemize}
    \item We model the wobble present in the motion of a spherical robot by developing equations that account for the coupling of its forward and steering motions. Section \ref{modeling} describes the method for modeling the underactuated system using Euler angles and the Lagrange-D'Alembert equations and takes the non-holonomic constraints into account.  
    \item We provide mathematical formulation for the wobble amplitude, wobble frequency, radius of curvature, and the precession rate associated with the robot's circular motion. Section \ref{analysisOfWobblyCircle} derives these formulations by simplifying the robot dynamics. These formulations illustrate the relationship between the aforementioned quantities (characterizing the robot's circular motion) and the robot's speed and pendulum angle. It is observed that the radius of curvature of the robot can be controlled indirectly by adjusting the robot's speed and pendulum angle. The section concludes by comparing the system response generated by the original model with that generated by the simplified model equations, demonstrating their similarity at different speeds.
    \item We propose a feedback linearization-based controller design for controlling a pendulum-actuated spherical robot's heading direction via a turning maneuver while limiting its wobble and maintaining a stable and constrained pendulum motion. The control set-points are determined by calculating the desired robot speed and pendulum angle based on the required turning radius of the turning maneuver. Section \ref{controller} elaborates on the proposed control strategy. 
\end{itemize}
 
\section{Modeling and Dynamics} \label{modeling}

This section discusses the modeling of the pendulum-actuated spherical robot described in this work. The robot's pendulum is mounted on the yoke at the geometric center of the hull through a motor that provides a torque $T_p$. The yoke and the hull are connected through a motor mounted on the yoke. This motor rotates the hull by providing a torque $T_s$. The center of mass of the hull and the yoke are at the geometric center of the hull, while the pendulum's center of mass is at a distance $r_p$ from the geometric center. We assume that the robot rolls without slipping and use the Lagrange D'Alembert formulation to determine the equations of motion for this system.

\subsection{Reference frames and Euler Angles}

This work is based on four reference frames as shown in figure \ref{fig:yxz}: a global inertial frame fixed to the ground ($\mathbf{G}$), and the three frames attached to the yoke ($\mathbf{Y}$), pendulum ($\mathbf{P}$) and the hull ($\mathbf{H}$) respectively with their origins at the geometric center of the hull. The yoke frame $\mathbf{Y}$ is defined to meet the following constraints: (1) The $z$-axis of the yoke is always aligned with the $z$-axis of the hull, and (2) The $x$-axis of the yoke frame lies in the global $XZ$-plane, i.e., it always remains parallel to the ground.

The orientation of the hull at any given instant is characterized by three $YXZ$ Euler angles $\phi$, $\theta$, and $\psi$. The orientation of the pendulum requires another angle $\beta$ due to an additional degree of freedom. The transformation between frames happens as follows :

\begin{figure}[!t]
\centering
\includegraphics[width=0.7\linewidth]{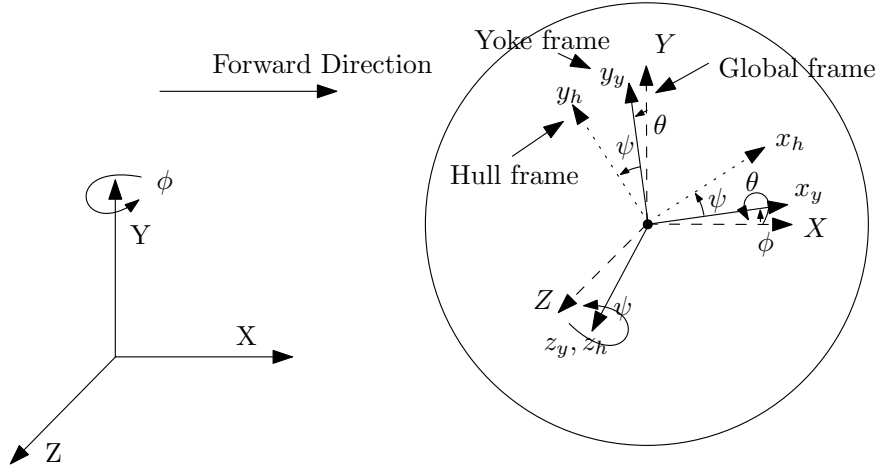}
\caption{Frames and Euler angles (YXZ)}
\label{fig:yxz}
\end{figure}

\begin{figure}[t!]
\centering 
\begin{minipage}[t]{\linewidth}
    \centering 
\begin{subfigure}{0.3\textwidth}
  \includegraphics[width=\linewidth]{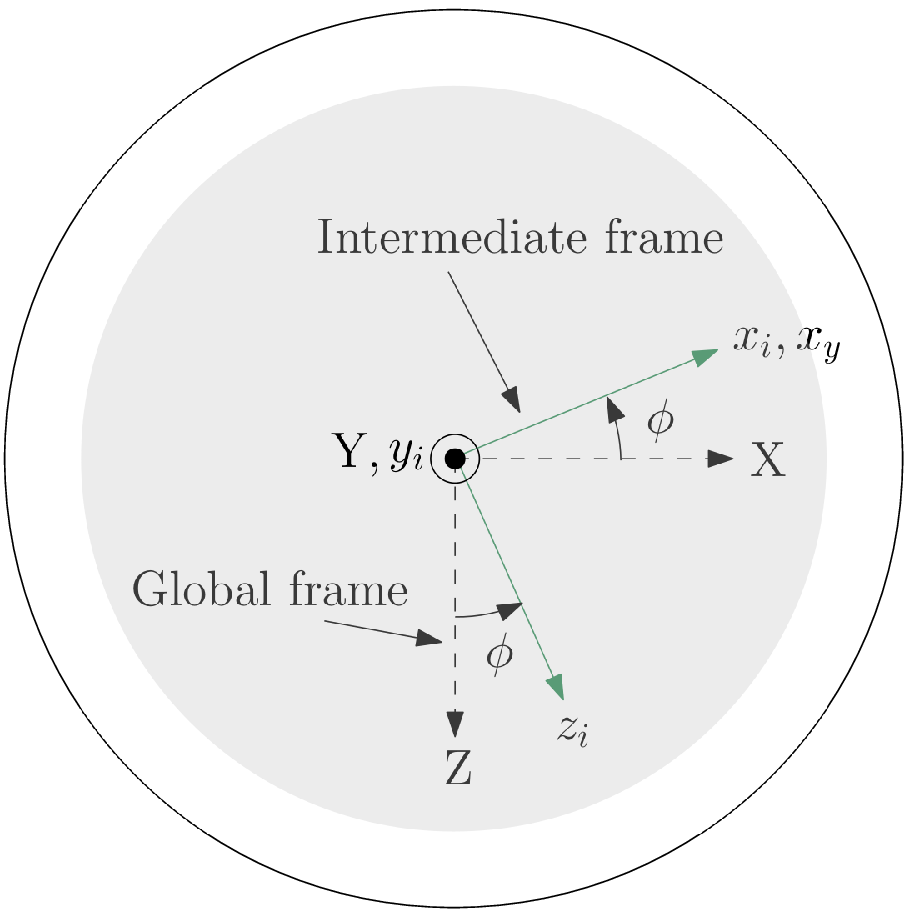}
  \caption{Rotation about Y axis (Top view)}
  \label{fig:fig3a_new}
\end{subfigure}\hfil 
\begin{subfigure}{0.3\textwidth}
  \includegraphics[width=\linewidth]{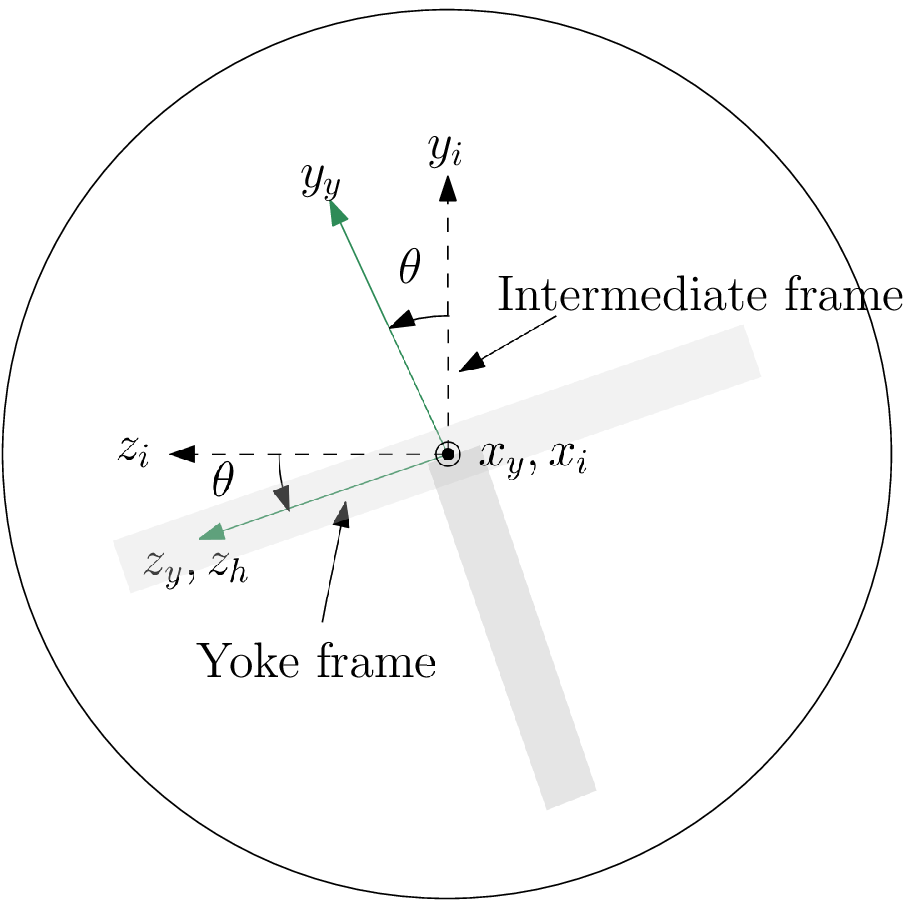}
  \caption{Rotation about $x_i$ axis (Front view)}
  \label{fig:fig3b_new}
\end{subfigure}\hfil 
\end{minipage}
\begin{minipage}[t]{\linewidth}
    \centering 
\begin{subfigure}{0.3\textwidth}
  \includegraphics[width=\linewidth]{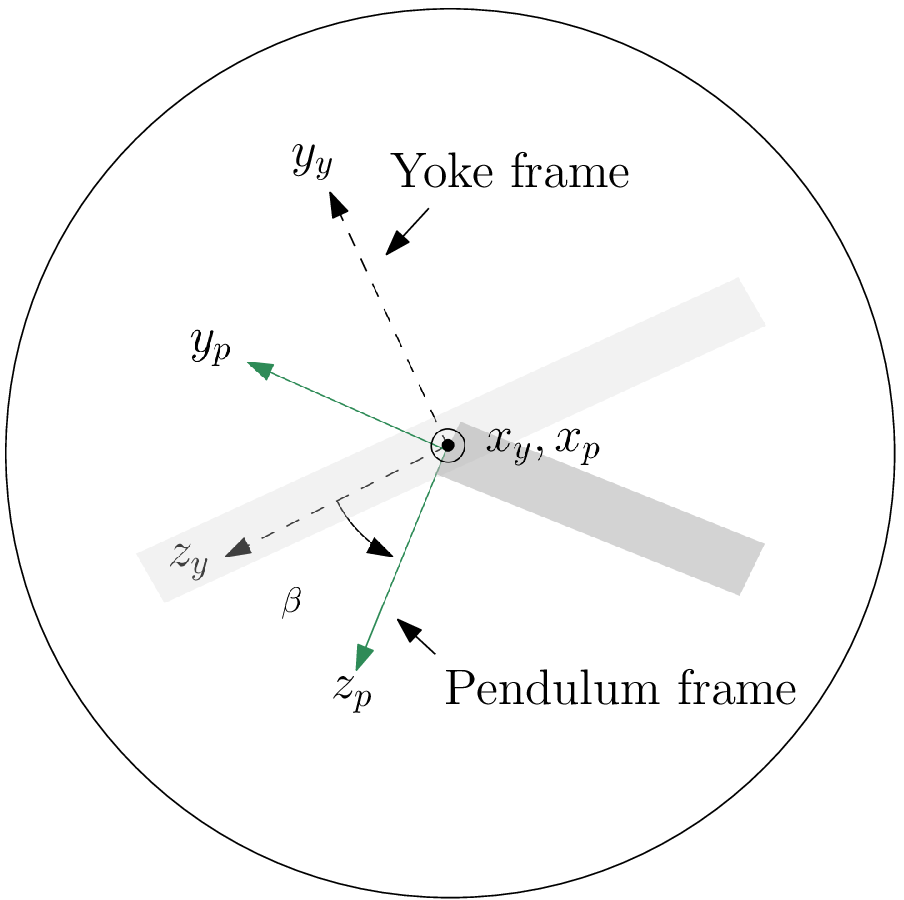}
  \caption{Rotation about $x_y$ axis (Front view)}
  \label{fig:fig3c_new}
\end{subfigure}\hfil 
\begin{subfigure}{0.3\textwidth}
  \includegraphics[width=\linewidth]{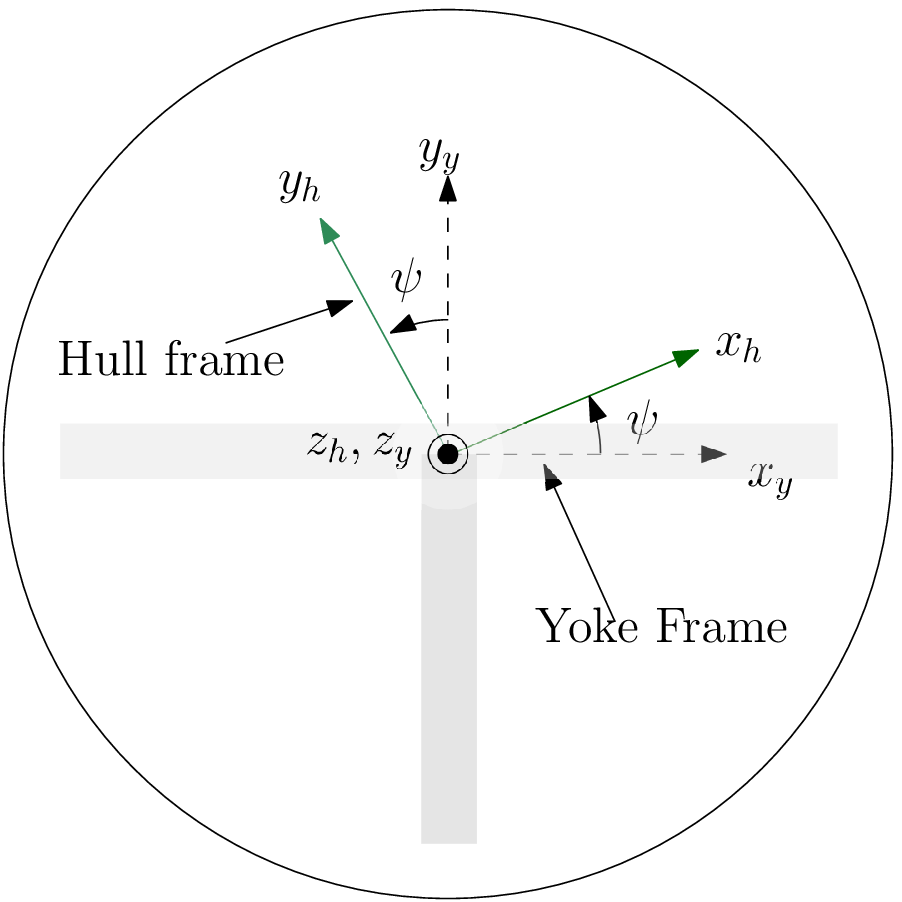}
  \caption{Rotation about $z_y$ axis (Side view)}
  \label{fig:fig3d_new}
\end{subfigure}\hfil 
\end{minipage}
\caption{Steps for obtaining the frames}
\label{fig:fig3_new}
\end{figure}

\begin{itemize}
    \item Intermediate Frame ($\mathbf{I}$): $\mathbf{G}$ is rotated along its $Y$-axis by angle $\phi$ to obtain $\mathbf{I}$ as shown in Figure \ref{fig:fig3a_new}. 
    \item Yoke frame ($\mathbf{Y}$): $\mathbf{I}$ is rotated along its local $x$-axis $x_i$ by angle $\theta$ to obtain $\mathbf{Y}$ as shown in Figure \ref{fig:fig3b_new}. 
    \item Pendulum frame ($\mathbf{P}$): $\mathbf{Y}$ is rotated along its local $x$-axis $x_y$ by angle $\beta$ to obtain $\mathbf{P}$ as shown in Figure \ref{fig:fig3c_new}. 
    Note that the angle made by the pendulum with the vertical axis $Y$ is ($\beta+\theta$).    
    \item Hull frame ($\mathbf{H}$): $\mathbf{Y}$ is rotated along its $z$-axis $z_y$ by angle $\psi$ to obtain $\mathbf{H}$ as shown in Figure \ref{fig:fig3d_new}. 
\end{itemize}

\subsection{Robot kinematics}
Rotation matrices mapping vectors from frames $\mathbf{Y}$, $\mathbf{P}$ and $\mathbf{H}$ to $\mathbf{G}$ are given by, ${{^{\mathbf{G}}}{\mathbf{R}}_{\mathbf{Y}}} = {R_y(\phi)} {R_x(\theta)}$, ${{^{\mathbf{G}}}{\mathbf{R}}_{\mathbf{P}}} = {R_y(\phi)} {R_x(\theta)} {R_x(\beta)}$ and ${{^{\mathbf{G}}}{\mathbf{R}}_{\mathbf{H}}} = {R_y(\phi)} {R_x(\theta)} {R_z(\psi)}$.
 
\noindent The angular velocities of different frames are obtained to be:

\begin{equation}
\vec{{^{\mathbf{{\mathbf{Y}}}}}\omega_{\mathbf{Y}}} = \begin{bmatrix}
\dot{\theta} && \dot{\phi}\,\cos(\theta) && -\dot{\phi}\,\sin(\theta)
\end{bmatrix}^\intercal{}
\end{equation}

\begin{equation}
\vec{{^{\mathbf{P}}}\omega_{\mathbf{P}}} = \begin{bmatrix}
\dot{\beta}+\dot{\theta} &&
\dot{\phi}\,\cos(\beta+\theta) &&
-\dot{\phi}\,\sin(\beta+\theta)
\end{bmatrix}^\intercal{} 
\end{equation}

\begin{equation}
\vec{{^{\mathbf{H}}}\omega_{\mathbf{H}}} = \begin{bmatrix}
\dot{\theta}\,\cos(\psi)\,+\,\dot{\phi}\,\cos(\theta)\,\sin(\psi) \\
\dot{\phi}\,\cos(\psi)\,\cos(\theta)\,-\,\dot{\theta}\,\sin(\psi) \\
\dot{\psi}\,-\,\dot{\phi}\,\sin(\theta)
\end{bmatrix}
\end{equation}

The position vector corresponding to the center of mass of the yoke (${\mathbf{Y}}_c$) and the hull (${\mathbf{H}}_c$) as well as origin of pendulum frame (${\mathbf{P}}_o$) are  
\begin{equation}
\vec{{}^{{\mathbf{G}}}_{{\mathbf{G}}}r_{{\mathbf{Y}}_c}} = \vec{{}^{{\mathbf{G}}}_{{\mathbf{G}}}r_{{\mathbf{H}}_c}} = 
\vec{{}^{{\mathbf{G}}}_{{\mathbf{G}}}r_{{\mathbf{P}}_o}} =
\begin{bmatrix}
X & r_h & Z
\end{bmatrix}^\intercal{}
\label{rHull}
\end{equation}
where $X$ and $Z$ are the coordinates of the hull center along the global frame ${\mathbf{G}}(OXYZ)$ and $r_h$ is the radius of the sphere. Note that the origins of frames $\mathbf{Y, P, H}$ coincide at the center of the robot. The position vector corresponding to the Pendulum's center of mass is given by:

\begin{equation}
\vec{{}^{{\mathbf{G}}}_{{\mathbf{G}}}r_{{\mathbf{P}}_c}} = \vec{{}^{{\mathbf{G}}}_{{\mathbf{P}}}r_{{\mathbf{P}}_c}} + \vec{{}^{{\mathbf{G}}}_{{\mathbf{G}}}r_{{\mathbf{P}}_o}} 
\label{rPendulum}
\end{equation}
\begin{equation}
\vec{{}^{{\mathbf{G}}}_{{\mathbf{P}}}r_{{\mathbf{P}}_c}} = {^{\mathbf{G}}}{\mathbf{R}}_{\mathbf{P}} \vec{{}^{{\mathbf{P}}}_{{\mathbf{P}}}r_{{\mathbf{P}}_c}}   
\text{ and }\vec{{}^{{\mathbf{P}}}_{{\mathbf{P}}}r_{{\mathbf{P}}_c}} = 
\begin{bmatrix}
0 & -r_p & 0
\end{bmatrix}^\intercal{}
\end{equation}

\noindent where ${\mathbf{P}}_c$ stands for pendulum's centre of mass, and $r_p$ is the distance between ${\mathbf{P}}_c$ and ${\mathbf{P}}_o$. 

Differentiating the position vectors, the linear velocities of the center of mass of the hull (${\mathbf{H}}_c$), yoke (${\mathbf{Y}}_c$) and the pendulum (${\mathbf{P}}_c$) are obtained as:
\begin{equation}
\vec{{^{\mathbf{G}}}v_{{\mathbf{H}}_c}} = \vec{{^{\mathbf{G}}}v_{{\mathbf{Y}}_c}} =
\begin{bmatrix}
\dot{X} & 0 & \dot{Z}
\end{bmatrix}^\intercal{}
\label{linVelHull}
\end{equation}
\begin{equation}
\vec{{^{\mathbf{G}}}v_{{\mathbf{P}}_c}} = 
\begin{bmatrix}
\dot{X} \\ 0 \\ \dot{Z}
\end{bmatrix} + \vec{{^{\mathbf{G}}}\omega_{\mathbf{P}}} \times {^{\mathbf{G}}}{\mathbf{R}}_{\mathbf{P}} \begin{bmatrix}
0 \\ -r_p \\ 0
\end{bmatrix} 
\end{equation}

The Kinetic Energy of the system can then be written as:
\begin{multline}
K = {\frac{1}{2}}(m_h \|\vec{{^{\mathbf{G}}}v_{{\mathbf{H}}_c}}\|^2 +
m_y \|\vec{{^{\mathbf{G}}}v_{{\mathbf{Y}}_c}}\|^2 + 
m_p \|\vec{{^{\mathbf{G}}}v_{{\mathbf{P}}_c}}\|^2) + \\
{\frac{1}{2}}(\vec{{^{\mathbf{H}}}\omega^\intercal_{\mathbf{H}}} {^{\mathbf{H}}}I_{\mathbf{H}} \vec{{^{\mathbf{H}}}\omega_{\mathbf{H}}} + 
\vec{{^{\mathbf{Y}}}\omega^\intercal_{\mathbf{Y}}} {^{\mathbf{Y}}}I_{\mathbf{{\mathbf{Y}}}} \vec{{^{\mathbf{Y}}}\omega_{\mathbf{Y}}} + 
\vec{{^{\mathbf{P}}}\omega^\intercal_{\mathbf{P}}} {^{\mathbf{P}}}I_{\mathbf{P}} \vec{{^{\mathbf{P}}}\omega_{\mathbf{P}}})
\end{multline}
where $m_j$ denoted the mass of body ${\mathbf{J}}$ and ${^{\mathbf{J}}}I_{\mathbf{J}}$ denotes the mass moment of inertia matrix of body ${\mathbf{J}}$ calculated in its own frame of reference. Here, ${^{\mathbf{H}}}I_{\mathbf{H}} = diag(I_h,I_h,I_h)$, ${^{\mathbf{Y}}}I_{\mathbf{Y}} = diag(I_y,2 I_y,I_y)$ and ${^{\mathbf{P}}}I_{\mathbf{P}} = diag(I_y,0,I_y)$ , where $I_h = {\frac{2}{3}} m_h r_h^2$, $I_y = {\frac{1}{4}} m_y r_h^2$, $I_p = {\frac{1}{3}} m_p r_p^2$.

The Potential Energy of the system is given by
\begin{equation}
V = m_p g (\vec{{}^{{\mathbf{G}}}_{{\mathbf{P}}}r_{{\mathbf{P}}_c}}\cdot{[0\text{ }1\text{ }0]}^\intercal) + 
m_y g (\vec{{}^{{\mathbf{G}}}_{{\mathbf{Y}}}r_{{\mathbf{Y}}_c}}\cdot{[0\text{ }1\text{ }0]}^\intercal)
\end{equation}
where the datum point for potential energy is chosen as the geometric centre of the sphere. 

\subsection{Non-holonomic Constraints}

The constraint of rolling without slipping is non-holonomic and is given by 

\begin{equation}
\vec{{^{\mathbf{G}}}v_{\mathbf{H}}} = \vec{{^{\mathbf{G}}}\omega_{\mathbf{H}}} \times \vec{{^{\mathbf{G}}}r_{\mathbf{H}}}
\label{pureRolling}  
\end{equation}

Here $\vec{{^{\mathbf{G}}}r_{\mathbf{H}}} = \begin{bmatrix} 0 & r_h & 0
\end{bmatrix}^\intercal$ denotes the vector (written in the global frame) between the stationary point on the hull, which is in contact with the ground and the sphere center. Equation (\ref{pureRolling}) is  simplified further as
\begin{equation}
\dot{X} = r_h (\dot{\theta} \sin(\phi) - \dot{\psi}\cos(\phi)\cos(\theta))
\label{xdot_eq}
\end{equation}
\begin{equation}
\dot{Z} = r_h (\dot{\theta} \cos(\phi) + \dot{\psi}\sin(\phi)\cos(\theta))
\label{zdot_eq}
\end{equation}


These constraint equations are written in the form
\begin{equation}
a_1^\intercal\cdot\dot{q} = 0 \text{ , } a_2^\intercal\cdot\dot{q} = 0 \text{; where generalized coordinates } q = \begin{bmatrix}
X & Z & \phi & \theta & \psi & \beta
\end{bmatrix}^\intercal \\
\label{EqToGet-a}
\end{equation}

Next, the Lagrangian model is formulated utilising the deduced constraints.  

\subsection{Lagrange D'Alembert equations} \label{model}

The Lagrange D'Alembert equations for non-holonomic systems are given by 

\begin{equation}
\frac{d}{dt}(\frac{\partial L}{\partial \dot{q}}) - \frac{\partial L}{\partial q} = Q + \lambda_1 a_1 + \lambda_2 a_2
\label{lagrangeEqn}
\end{equation}
where the generalized forces $Q$ = $[0\text{ }0\text{ }0\text{ }0\text{ }T_s\text{ }T_p]^\intercal$. $T_s$ is the torque applied for forward motion, and $T_p$ is the torque applied on the pendulum. $\lambda_1$ and $\lambda_2$ are Lagrange multipliers and $a_1$ and $a_2$ are obtained by simplifying equation (\ref{EqToGet-a}).  


Equations (\ref{xdot_eq}),  (\ref{zdot_eq}) and (\ref{lagrangeEqn}) render the dynamic model of the robot and are simplified to represent the system in control affine form as $\dot{\mathbf{x}} = \mathbf{f}(\mathbf{x}) + \mathbf{G}(\mathbf{x})\mathbf{u}$. Here, ${\mathbf{x}}$ is the state vector given by $\begin{bmatrix}
\phi & \theta & \psi & \beta \textbf{ }\textbf{ } X & Z & \dot{\phi} & \dot{\theta} & \dot{\psi} \textbf{ }\textbf{ } \dot{\beta} & \dot{X} & \dot{Z} 
\end{bmatrix}^\intercal$, \textbf{f} is a smooth vector field, \textbf{G} is a $12\times2$ matrix whose columns are smooth vector fields G$_{i,j}$ and \textbf{u} is the control input vector given by $\begin{bmatrix}
T_s & T_p 
\end{bmatrix}^\intercal$.

The model upon simplification is obtained as: 
\begin{equation}
\begin{bmatrix}
\dot{\phi} \\ \dot{\theta} \\ \dot{\psi} \\ \dot{\beta} \\ \dot{X} \\ \dot{Z} \\ \ddot{\phi} \\ \ddot{\theta} \\ \ddot{\psi} \\ \ddot{\beta} \\ \ddot{X} \\ \ddot{Z}
\end{bmatrix} = 
\begin{bmatrix}
\dot{\phi} \\ \dot{\theta} \\ \dot{\psi} \\ \dot{\beta} \\ \dot{X} \\ \dot{Z} \\ f_7(\mathbf{x}) \\ f_8(\mathbf{x}) \\ f_9(\mathbf{x}) \\ f_{10}(\mathbf{x}) \\ f_{11}(\mathbf{x}) \\ f_{12}(\mathbf{x})
\end{bmatrix} +
\begin{bmatrix}
0 & 0 \\ 0 & 0 \\ 0 & 0 \\ 0 & 0 \\ 0 & 0 \\ 0 & 0 \\ G_{7,1}(\mathbf{x}) & 0 \\ 0 & G_{8,2}(\mathbf{x}) \\  G_{9,1}(\mathbf{x}) & 0 \\ 0 & G_{10,2}(\mathbf{x}) \\ G_{11,1}(\mathbf{x}) & G_{11,2}(\mathbf{x}) \\ G_{12,1}(\mathbf{x}) & G_{12,2}(\mathbf{x})
\end{bmatrix}
\begin{bmatrix}
T_s \\ T_p
\end{bmatrix}
\end{equation}

Note that this system is under-actuated due to just two control knobs $T_s$ and $T_p$. As a result, only a limited number of outputs can be controlled. 

\subsection{System response for a circular motion} \label{og_system_circle_response}

The dynamics of the spherical robot described in section \ref{model} are simulated using the ODE15s solver built into MATLAB. We simulate the steady-state behavior of the robot's circular motion. This configuration is attained when the pendulum's angle and robot's forward speed are held constant, and the remaining state variables are initialized as zero. 

We conduct our system analysis for relatively smaller pendulum angles so that the robot does not lean excessively during steering motion. A steep lean angle during the steering motion would laterally tilt the camera mounted on the robot. Oscillations in this tilt would cause the camera's feedback to be shaky. To circumvent this problem, we focus our analysis on a pendulum's swing of about 0$^{\circ}$ to 15$^{\circ}$ since this is a relatively small inclination. Within this range, simulations are roughly divided into two groups: low pendulum angles of 5$^{\circ}$ and high pendulum angles of 15$^{\circ}$. 

\begin{figure}
    \centering 
\begin{subfigure}{0.2\textwidth}
  \centering
  \includegraphics[width=\linewidth, trim = 1cm 0 1cm 0]{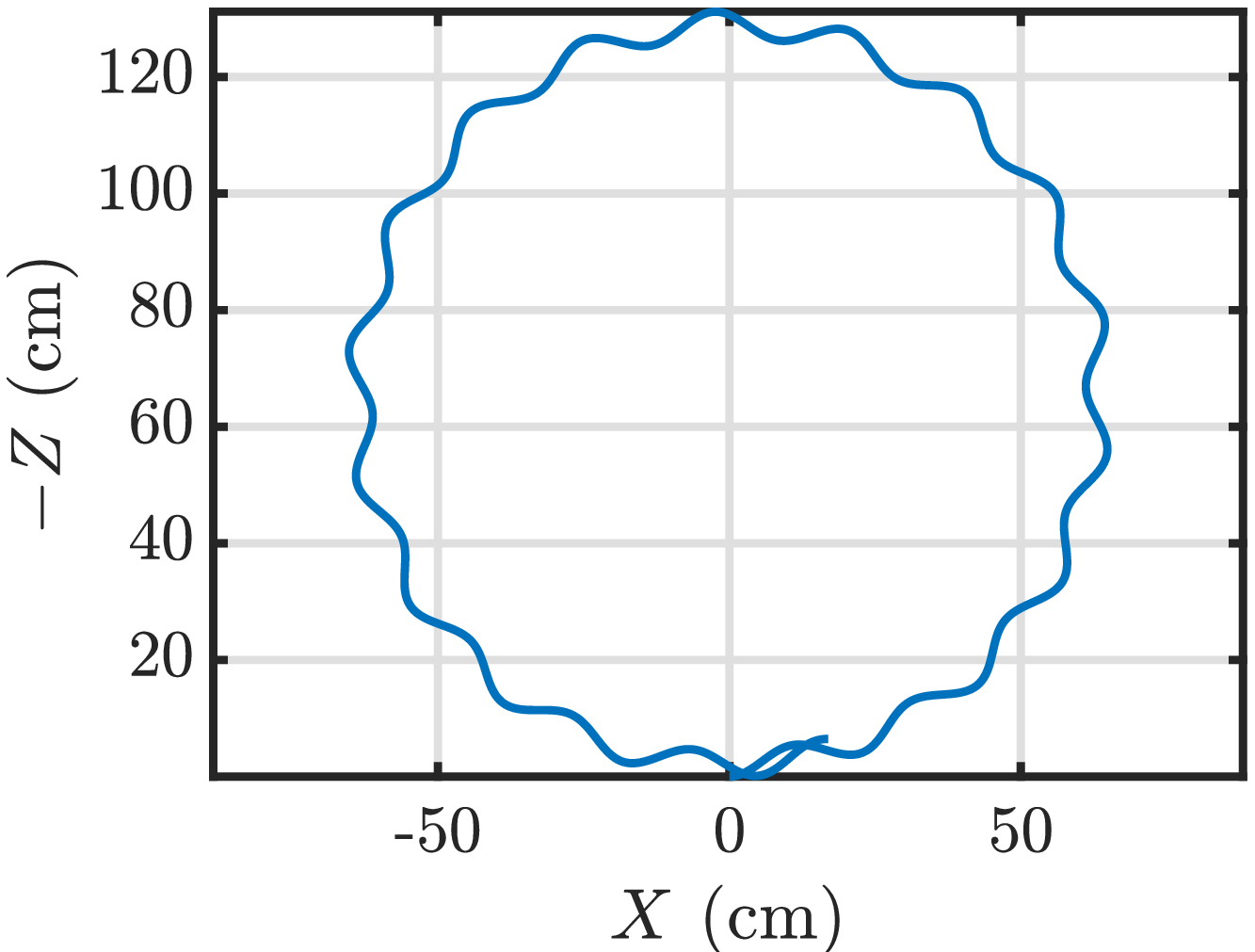}
  \caption{\centering{|$\dot{\psi}$| = 1 rad/s}}
\end{subfigure}\hfil 
\begin{subfigure}{0.2\textwidth}
  \centering
  \includegraphics[width=\linewidth, trim = 1cm 0 1cm 0]{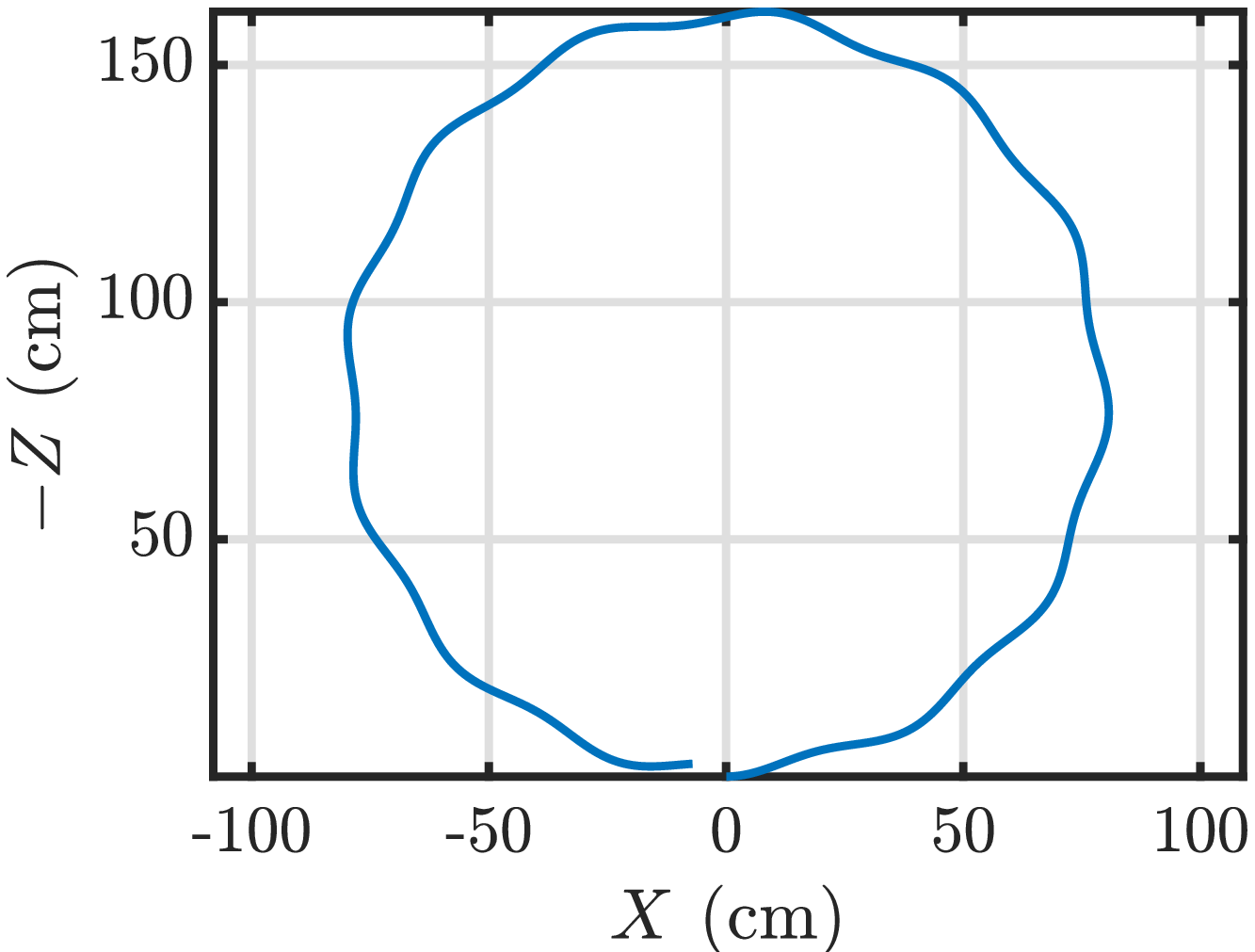}
  \caption{\centering{|$\dot{\psi}$| = 2 rad/s}}
\end{subfigure}\hfil 
\begin{subfigure}{0.2\textwidth}
  \centering
  \includegraphics[width=\linewidth, trim = 1cm 0 1cm 0]{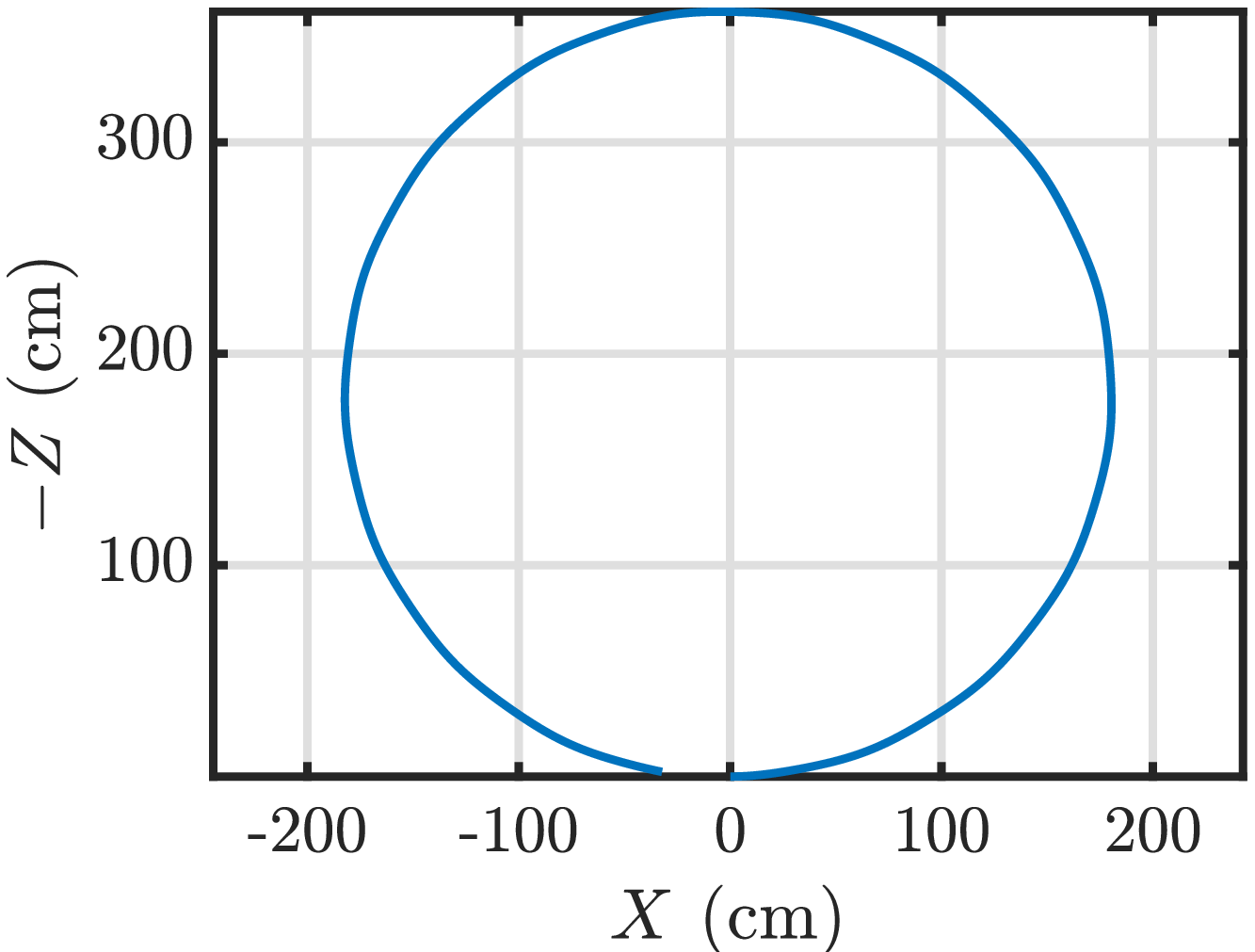}
  \caption{\centering{|$\dot{\psi}$| = 5 rad/s}}
\end{subfigure}\hfil 
\begin{subfigure}{0.2\textwidth}
  \centering
  \includegraphics[width=\linewidth, trim = 1cm 0 1cm 0]{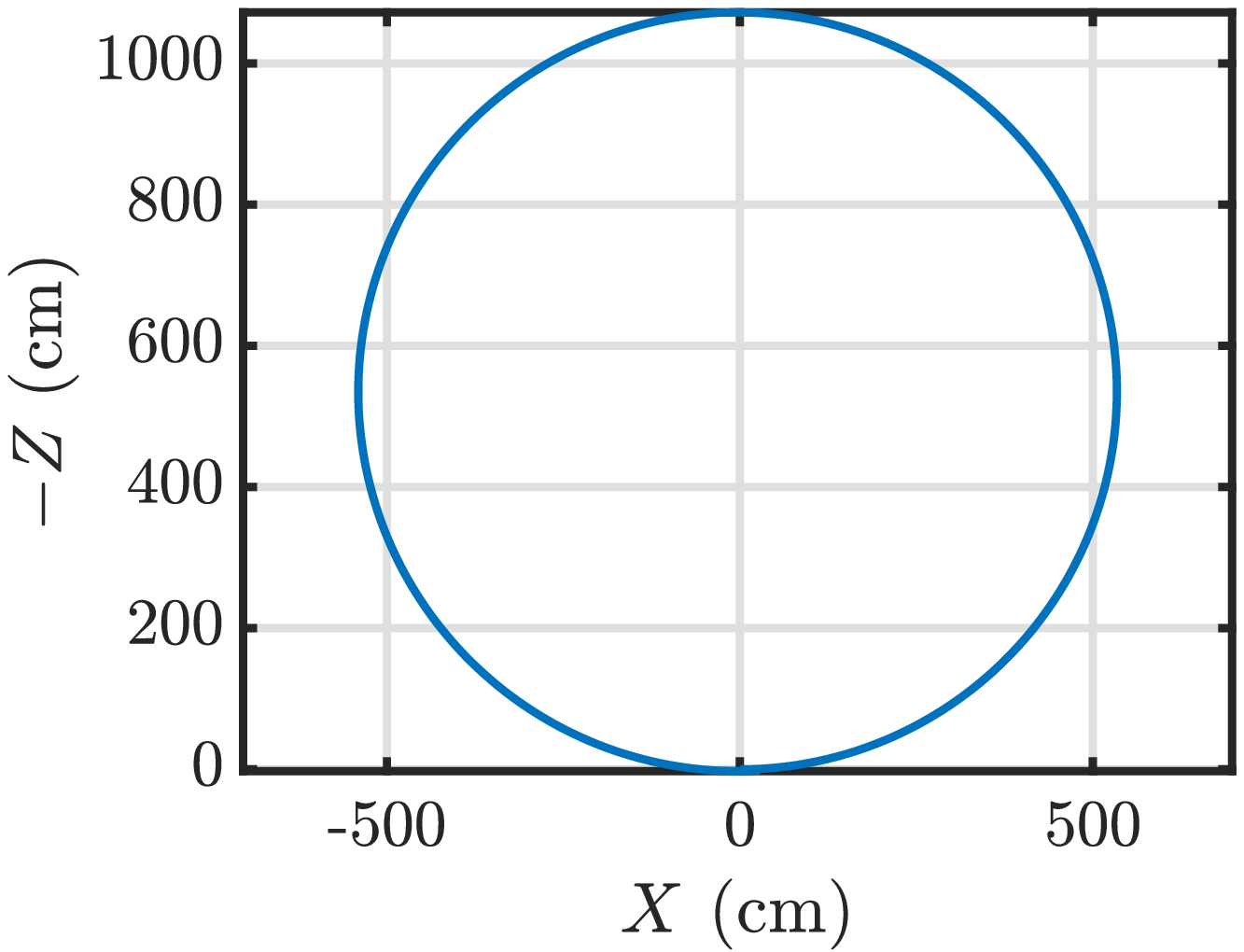}
  \caption{\centering{|$\dot{\psi}$| = 10 rad/s}}
\end{subfigure}\hfil 

\caption{Path followed by robot's COM while moving forward at different speeds |$\dot{\psi}$| with $\beta = 15^{\circ}$}
\label{fig:trajectories}
\end{figure}

We further categorize operation speeds as fast or slow. The system response is depicted in Figure \ref{fig:trajectories}, in which the robot's center of mass follows wobbly circular paths while moving forward at varying speeds and maintaining a pendulum angle ($\beta$) of 15$^{\circ}$. According to this figure, the robot's wobble is minimal at 10 rad/s, and the robot follows a nearly smooth circular trajectory; thus, for our simulations, we consider this a high speed. Similarly, we consider 1 rad/s to be a low speed because the effect of wobbling caused by center of mass perturbation is clearly visible in this case.

\begin{figure}
\begin{minipage}[t]{\linewidth}
    \centering 
\begin{subfigure}{0.2\textwidth}
  \includegraphics[width=\linewidth, trim = 0.8cm 0 0.8cm 0]{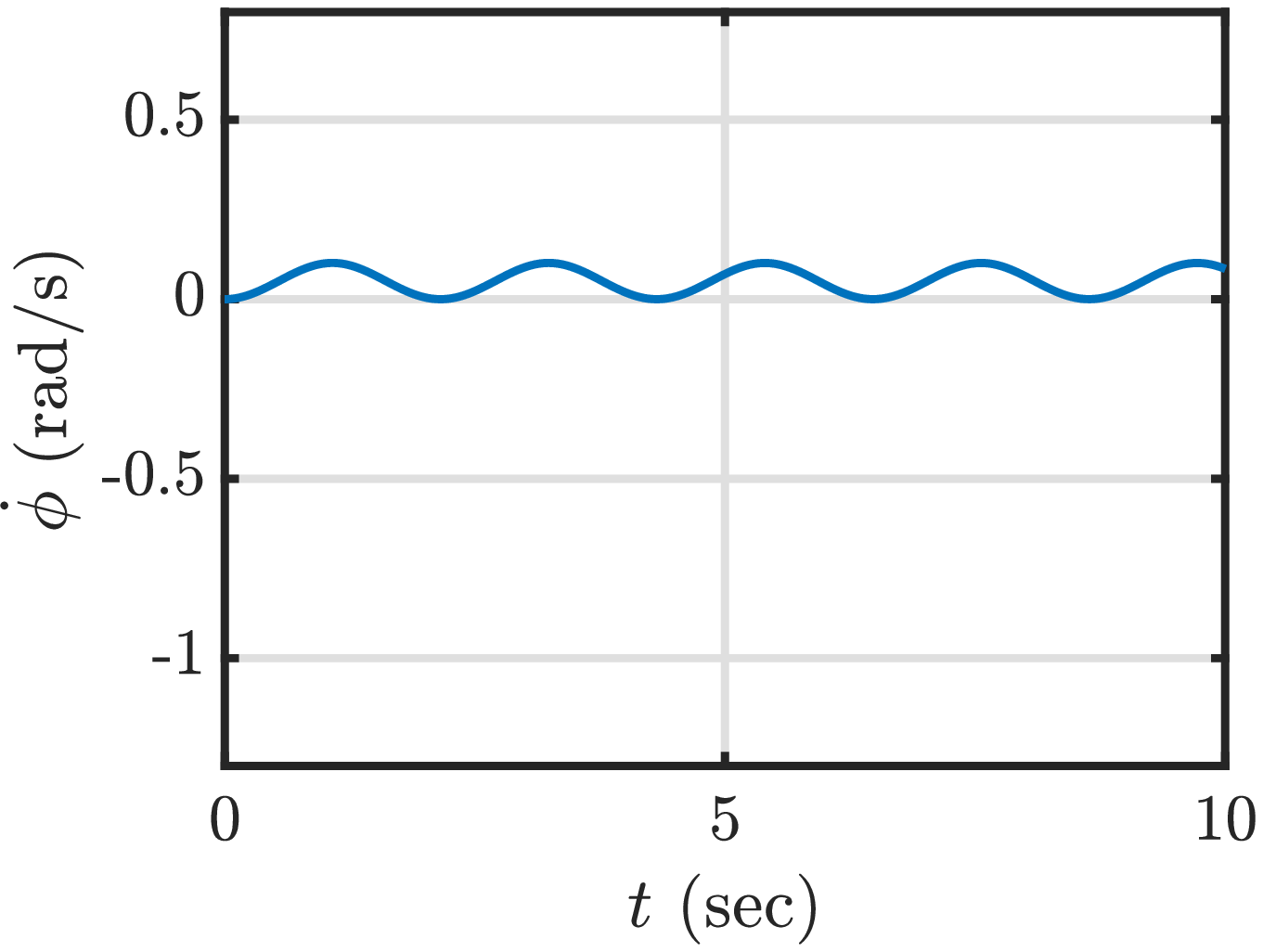}
  \caption{$\dot{\phi}$ vs. Time}
  \label{fig:5deg1radphid}
\end{subfigure}\hfil 
\begin{subfigure}{0.2\textwidth}
  \includegraphics[width=\linewidth, trim = 0.8cm 0 0.8cm 0]{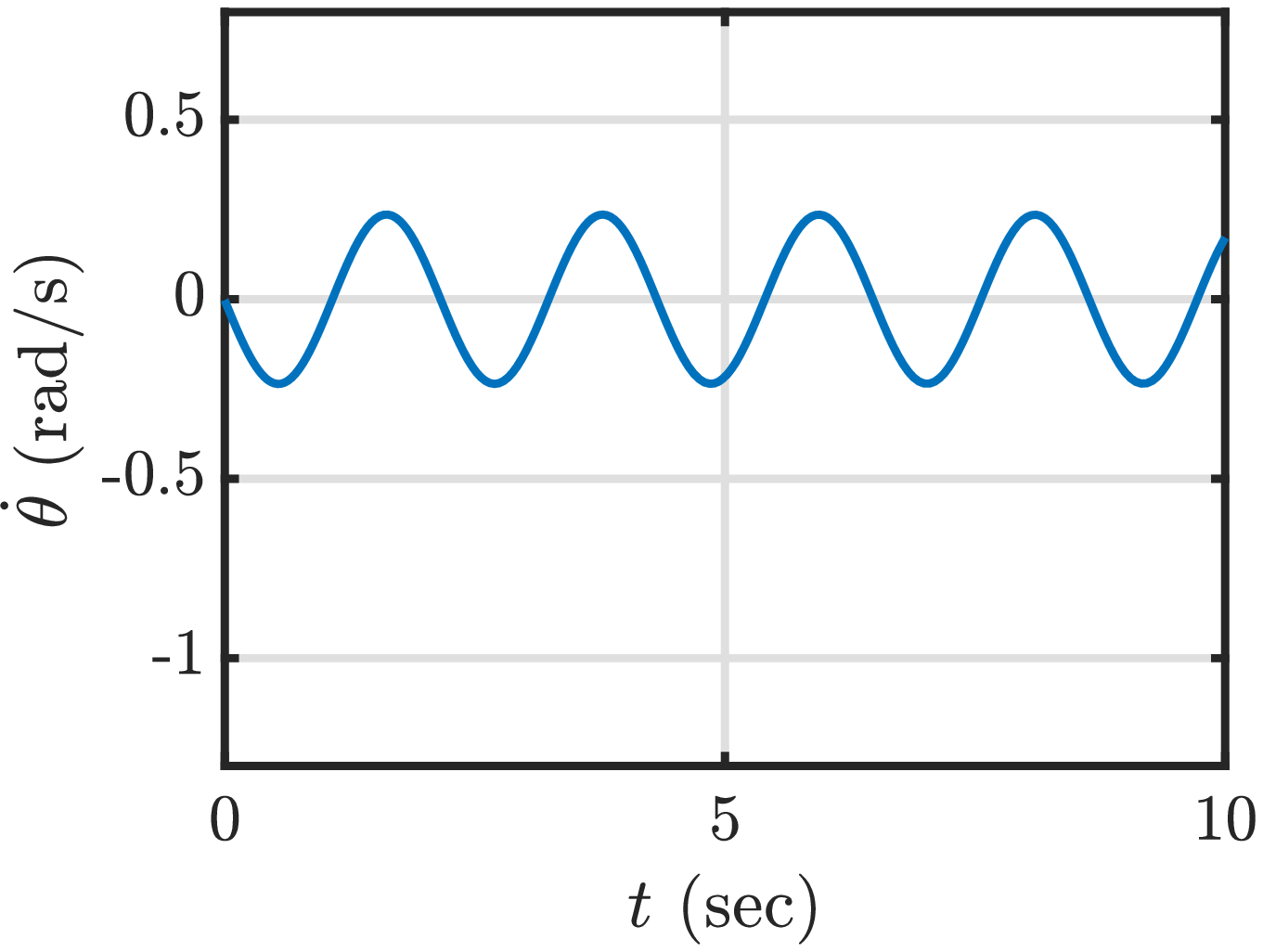}
  \caption{$\dot{\theta}$ vs. Time}
  \label{fig:5deg1radthetad}
\end{subfigure}\hfil 
\begin{subfigure}{0.2\textwidth}
  \includegraphics[width=\linewidth, trim = 0.8cm 0 0.8cm 0]{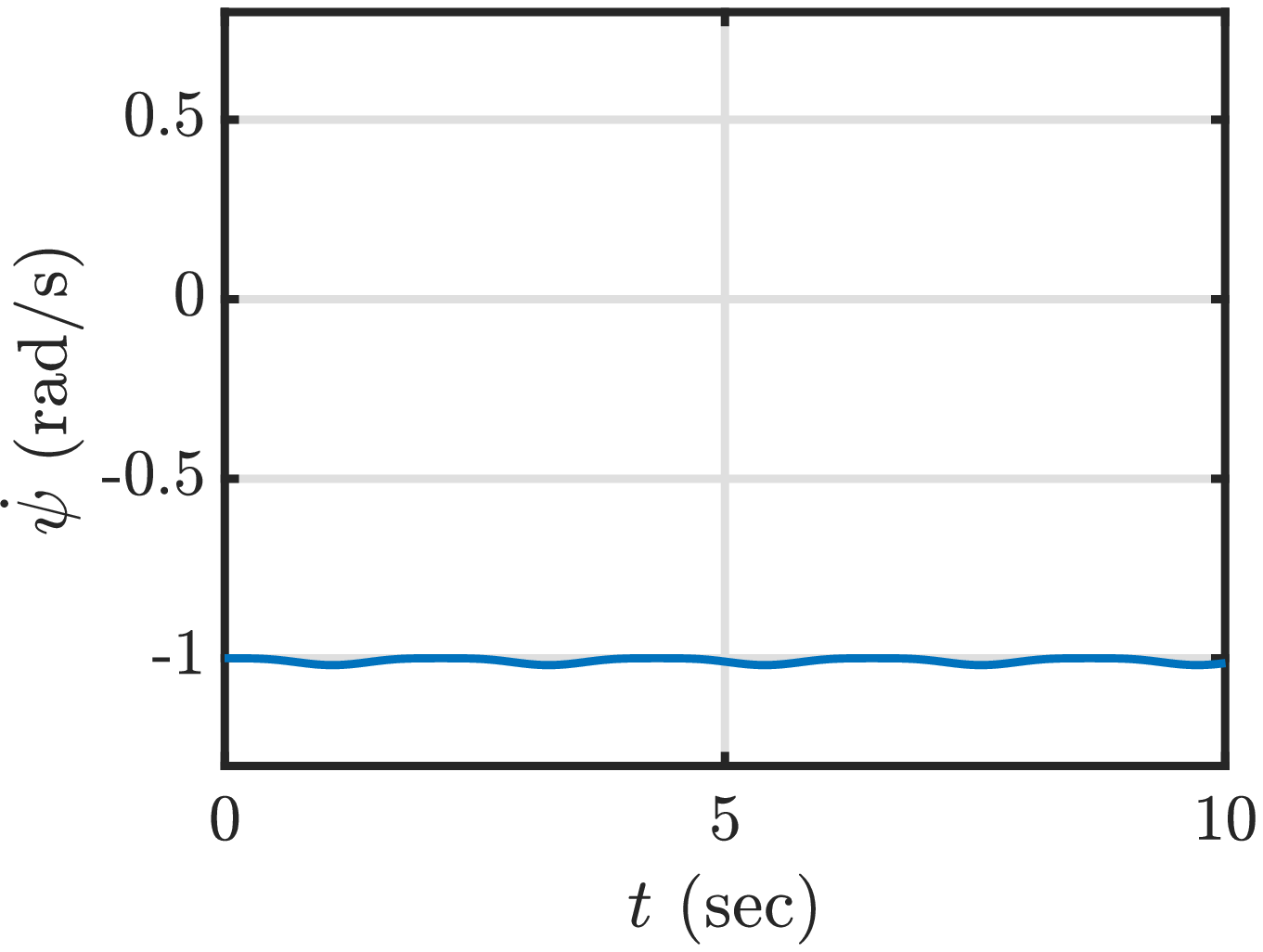}
  \caption{$\dot{\psi}$ vs. Time}
  \label{fig:5deg1radpsid}
\end{subfigure}\hfil 
\begin{subfigure}{0.2\textwidth}
  \includegraphics[width=\linewidth, trim = 0.8cm 0 0.8cm 0]{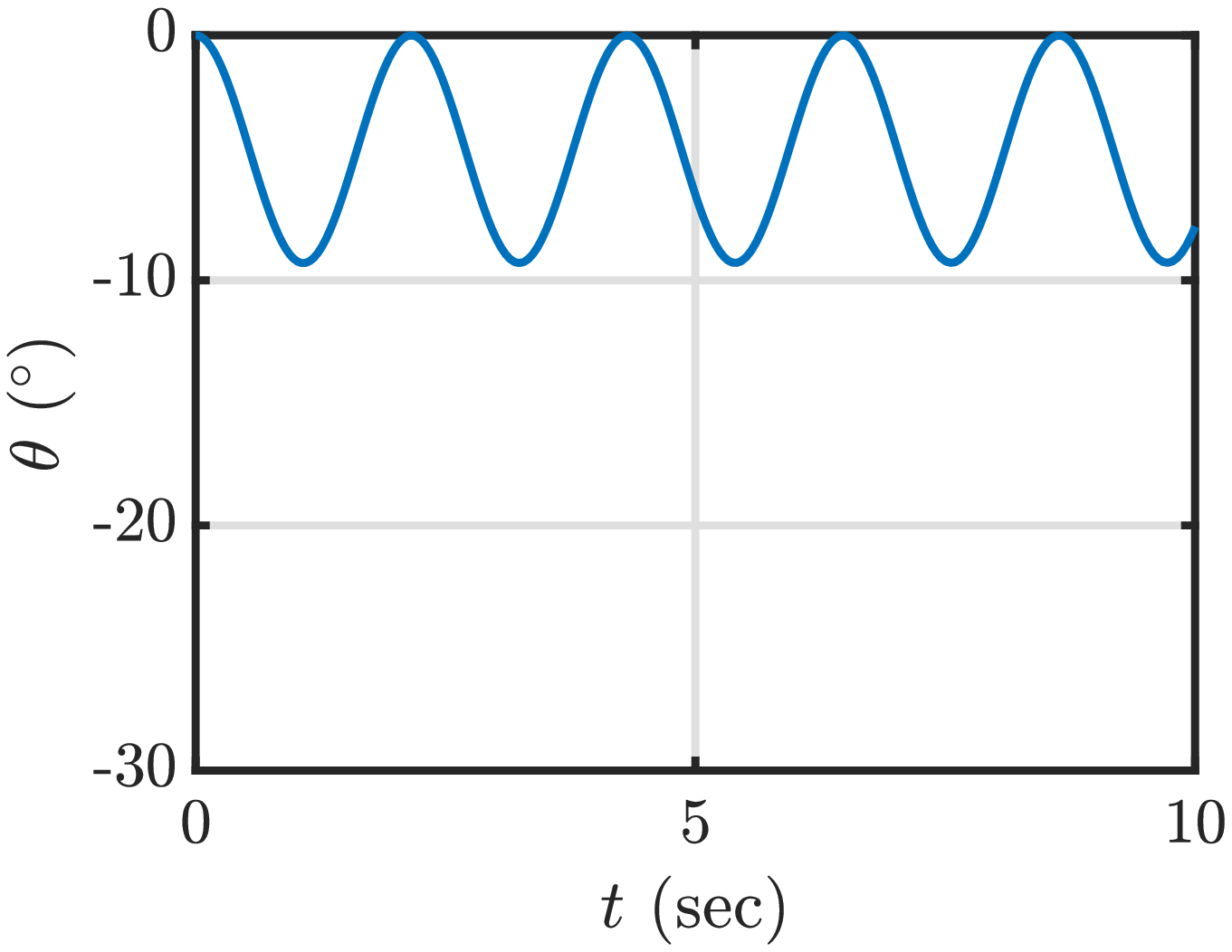}
  \caption{$\theta$ vs. Time}
  \label{fig:5deg1radtheta}
\end{subfigure}\hfil 

\caption{System response at $\beta$ =  5$^{\circ}$ and |$\dot{\psi}$| = 1 rad/s (low speed)}
\label{fig:5deg1rad}
\end{minipage}

\begin{minipage}[t]{\linewidth}
    \centering 
\begin{subfigure}{0.2\textwidth}
  \includegraphics[width=\linewidth, trim = 0.8cm 0 0.8cm 0]{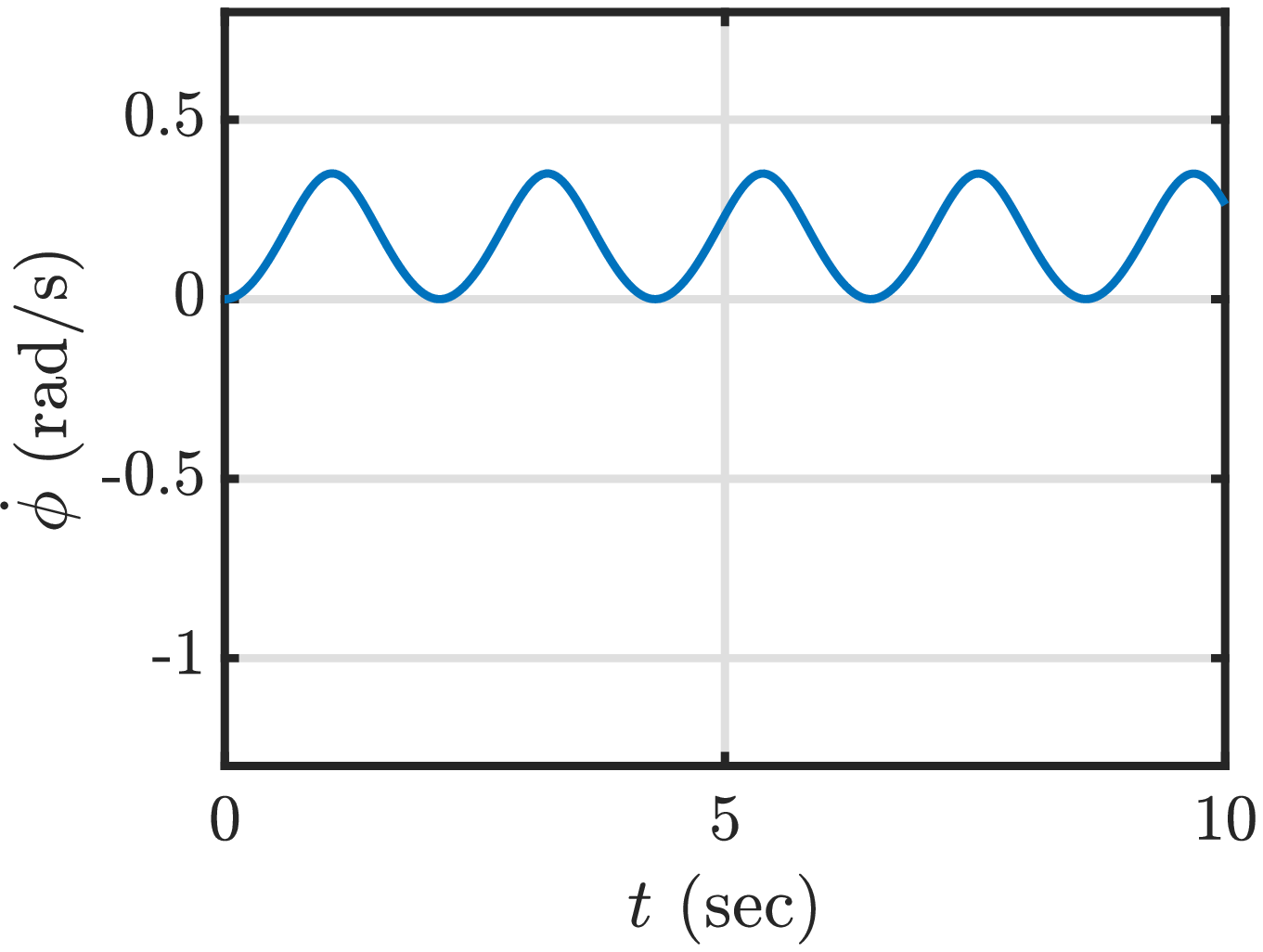}
  \caption{$\dot{\phi}$ vs. Time}
  \label{fig:15deg1radphid}
\end{subfigure}\hfil 
\begin{subfigure}{0.2\textwidth}
  \includegraphics[width=\linewidth, trim = 0.8cm 0 0.8cm 0]{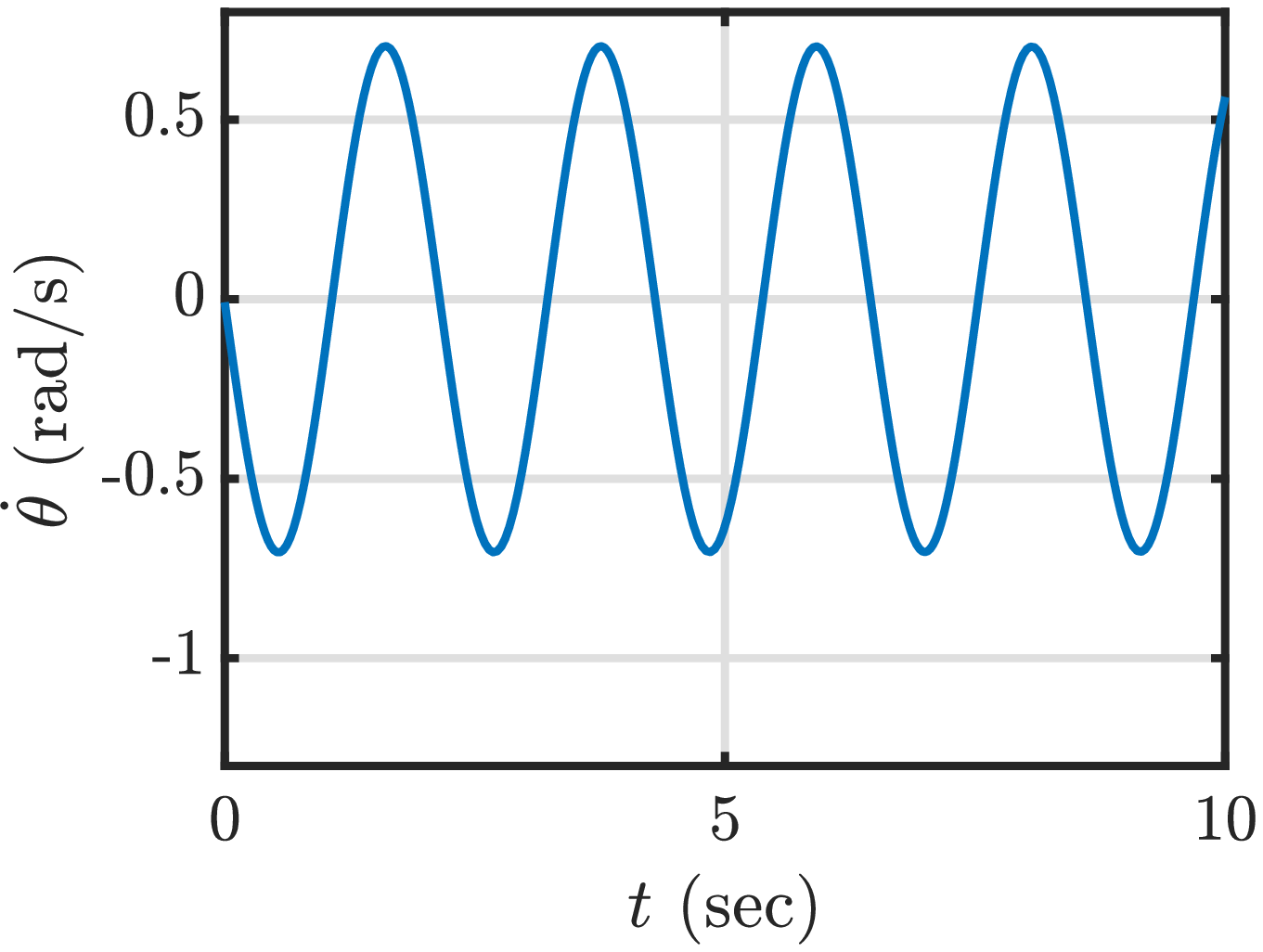}
  \caption{$\dot{\theta}$ vs. Time}
  \label{fig:15deg1radthetad}
\end{subfigure}\hfil 
\begin{subfigure}{0.2\textwidth}
  \includegraphics[width=\linewidth, trim = 0.8cm 0 0.8cm 0]{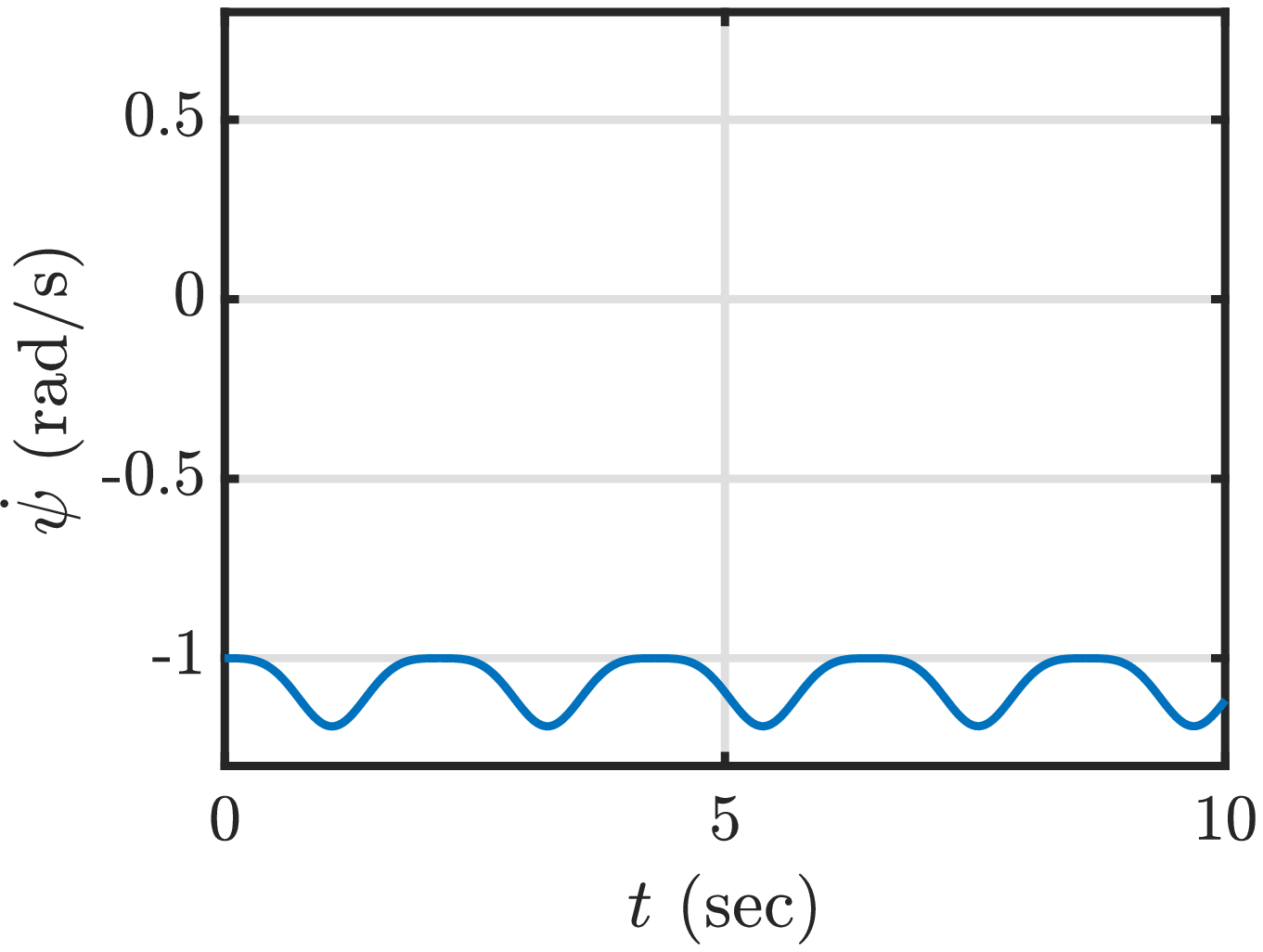}
  \caption{$\dot{\psi}$ vs. Time}
  \label{fig:15deg1radpsid}
\end{subfigure}\hfil 
\begin{subfigure}{0.2\textwidth}
  \includegraphics[width=\linewidth, trim = 0.8cm 0 0.8cm 0]{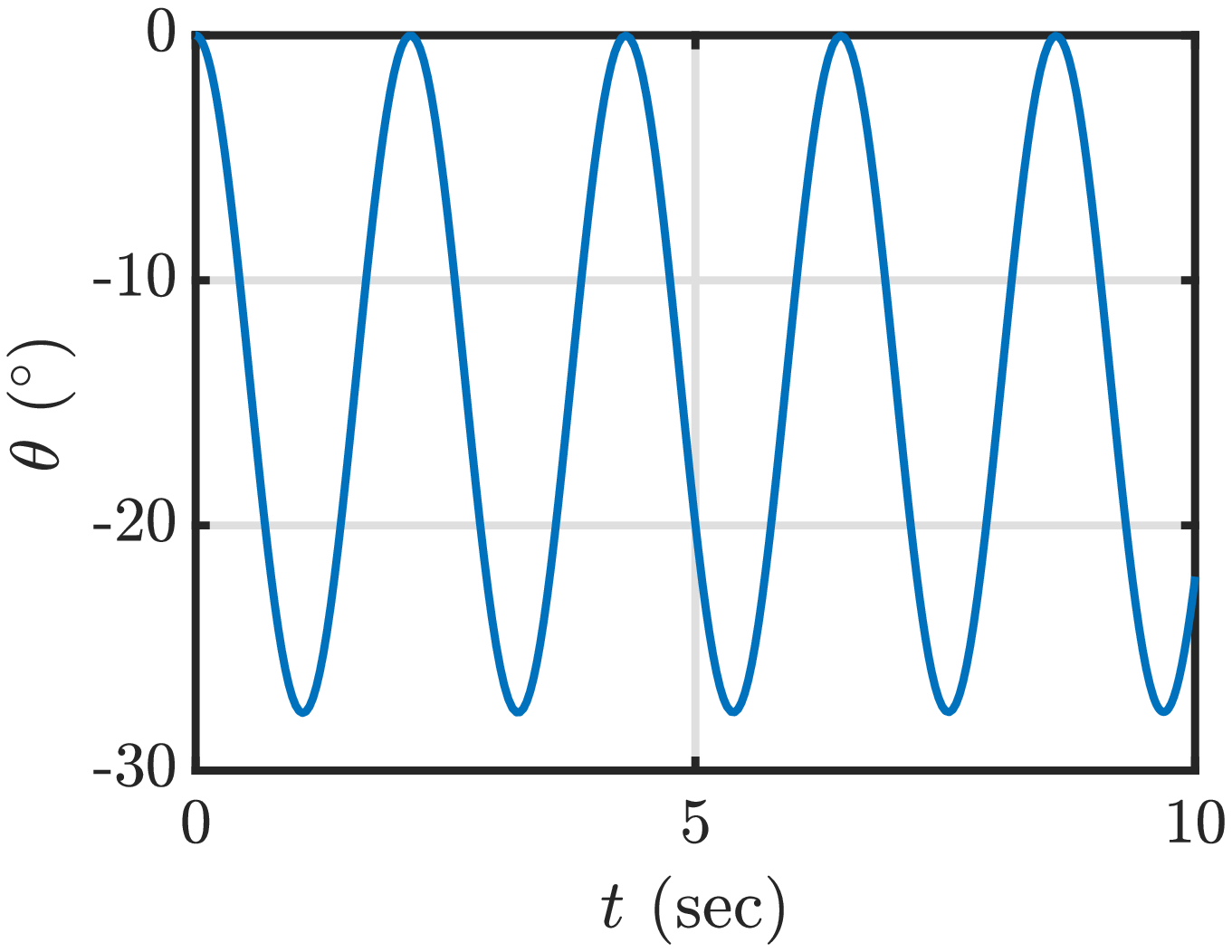}
  \caption{$\theta$ vs. Time}
  \label{fig:15deg1radtheta}
\end{subfigure}\hfil 

\caption{System response at $\beta$ =  15$^{\circ}$ and |$\dot{\psi}$| = 1 rad/s (low speed)}
\label{fig:15deg1rad}
\end{minipage}

\begin{minipage}[t]{\linewidth}
    \centering 
\begin{subfigure}{0.2\textwidth}
  \includegraphics[width=\linewidth, trim = 0.8cm 0 0.8cm 0]{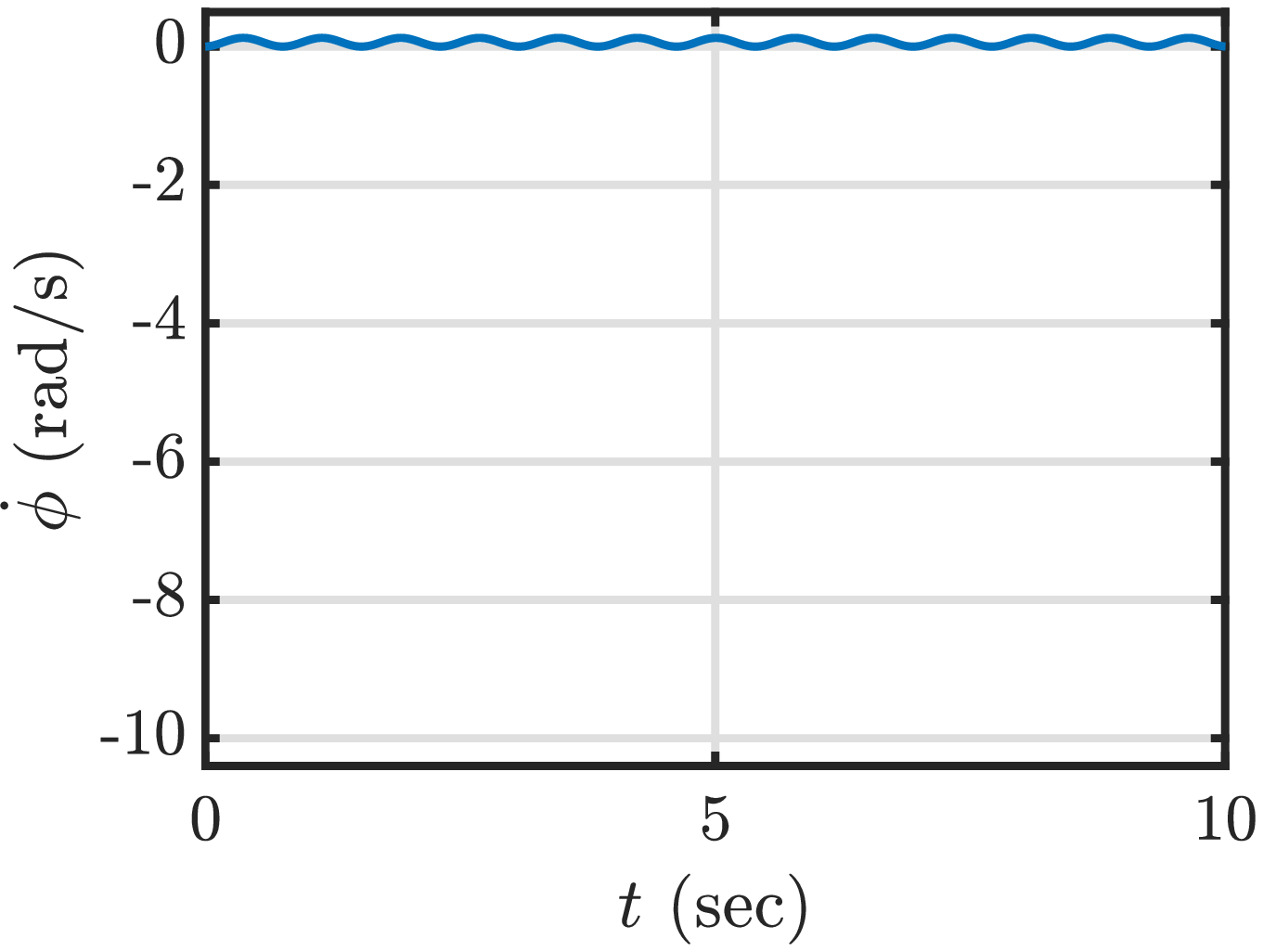}
  \caption{$\dot{\phi}$ vs. Time}
  \label{fig:5deg10radphid}
\end{subfigure}\hfil 
\begin{subfigure}{0.2\textwidth}
  \includegraphics[width=\linewidth, trim = 0.8cm 0 0.8cm 0]{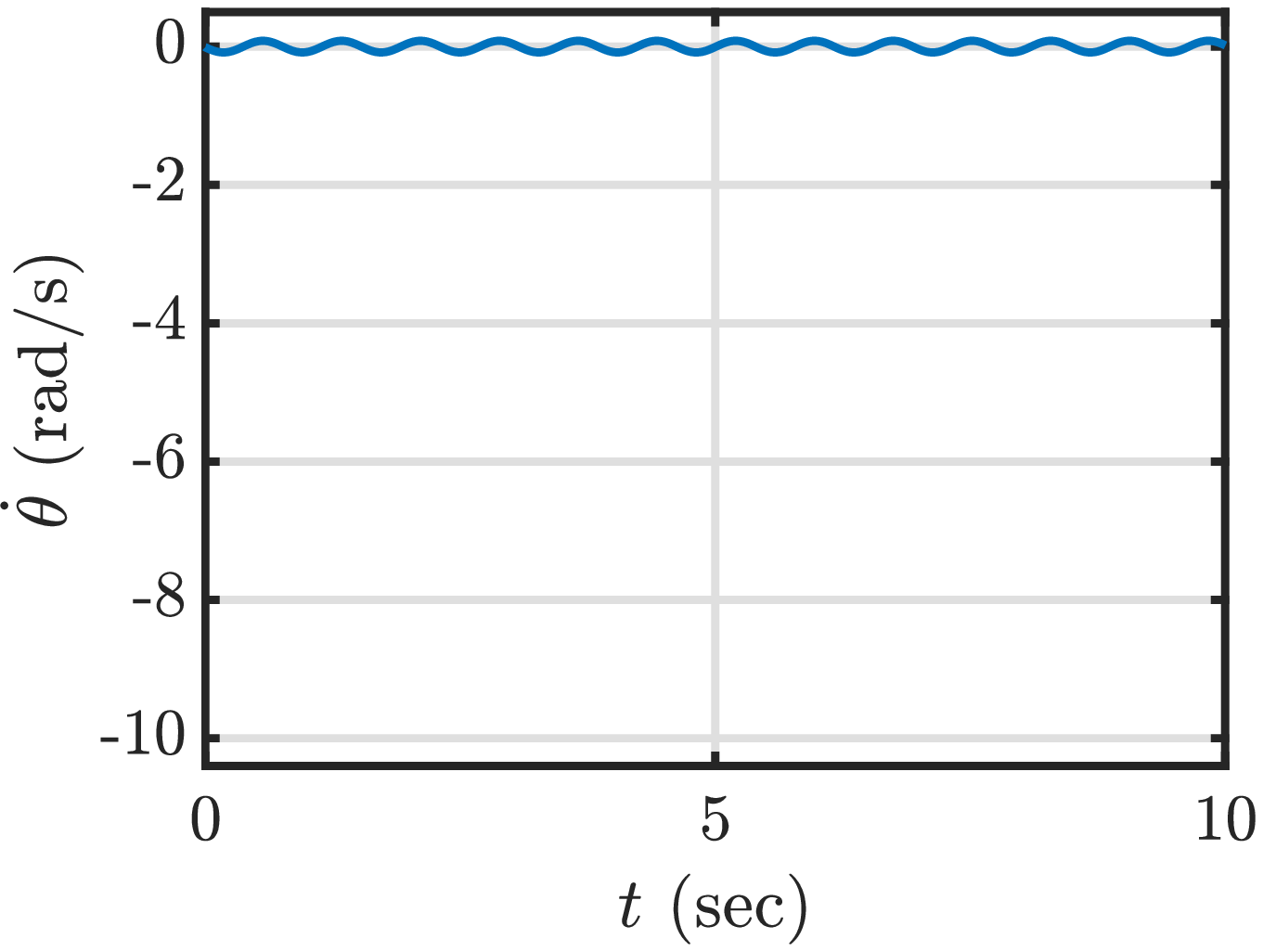}
  \caption{$\dot{\theta}$ vs. Time}
  \label{fig:5deg10radthetad}
\end{subfigure}\hfil 
\begin{subfigure}{0.2\textwidth}
  \includegraphics[width=\linewidth, trim = 0.8cm 0 0.8cm 0]{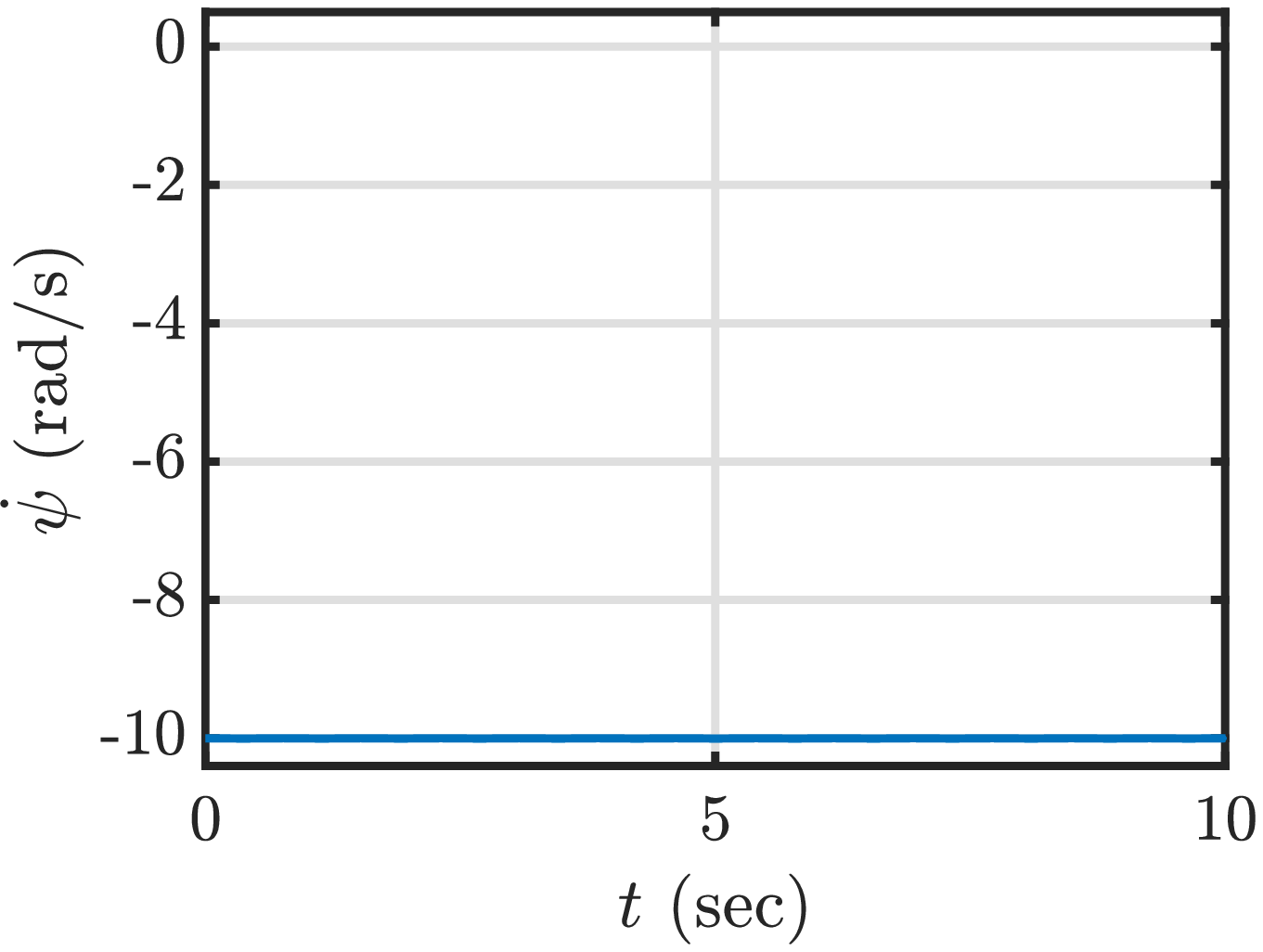}
  \caption{$\dot{\psi}$ vs. Time}
  \label{fig:5deg10radpsid}
\end{subfigure}\hfil 
\begin{subfigure}{0.2\textwidth}
  \includegraphics[width=\linewidth, trim = 0.8cm 0 0.8cm 0]{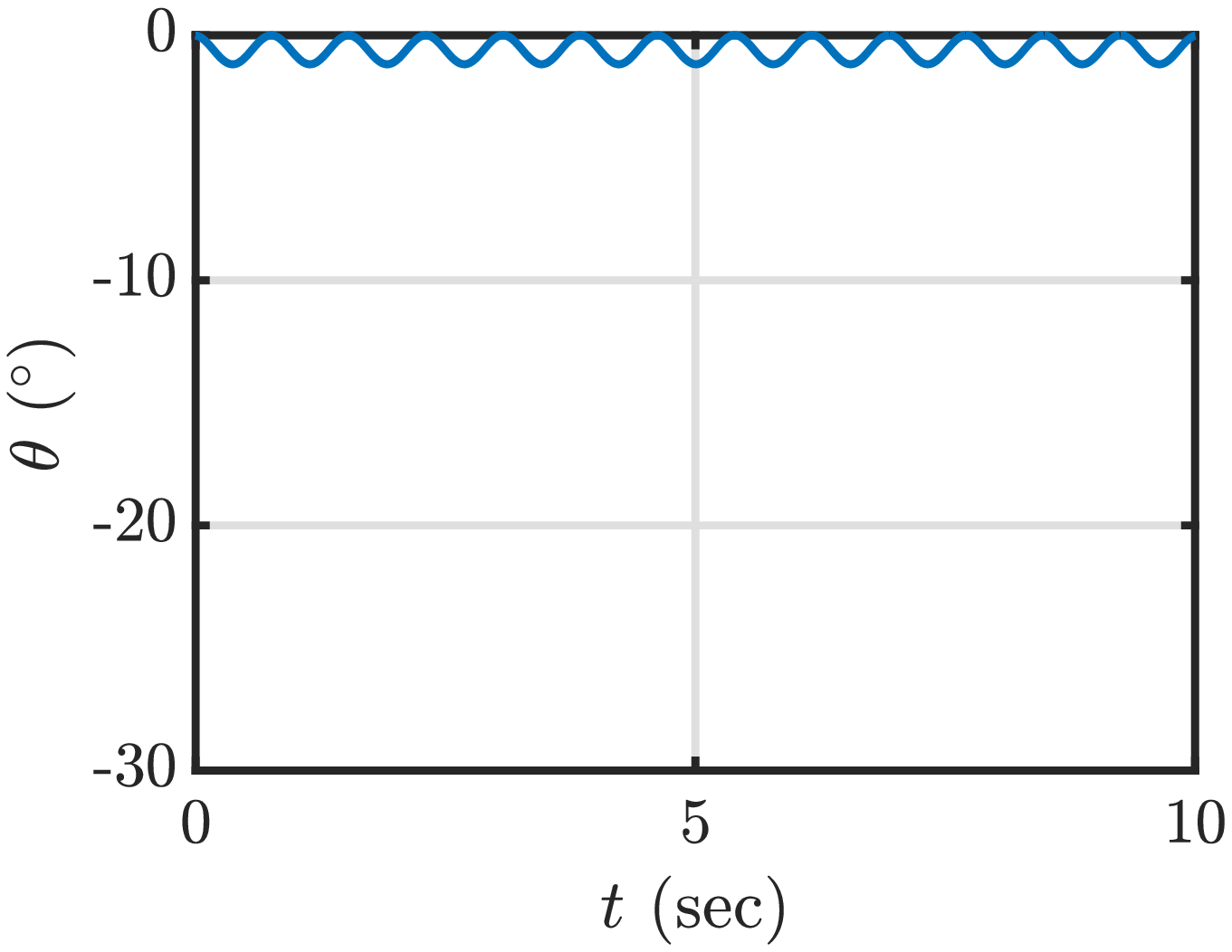}
  \caption{$\theta$ vs. Time}
  \label{fig:5deg10radtheta}
\end{subfigure}\hfil 

\caption{System response at $\beta$ =  5$^{\circ}$ and |$\dot{\psi}$| = 10 rad/s (high speed)}
\label{fig:5deg10rad}
\end{minipage}

\begin{minipage}[t]{\linewidth}
    \centering 
\begin{subfigure}{0.2\textwidth}
  \includegraphics[width=\linewidth, trim = 0.8cm 0 0.8cm 0]{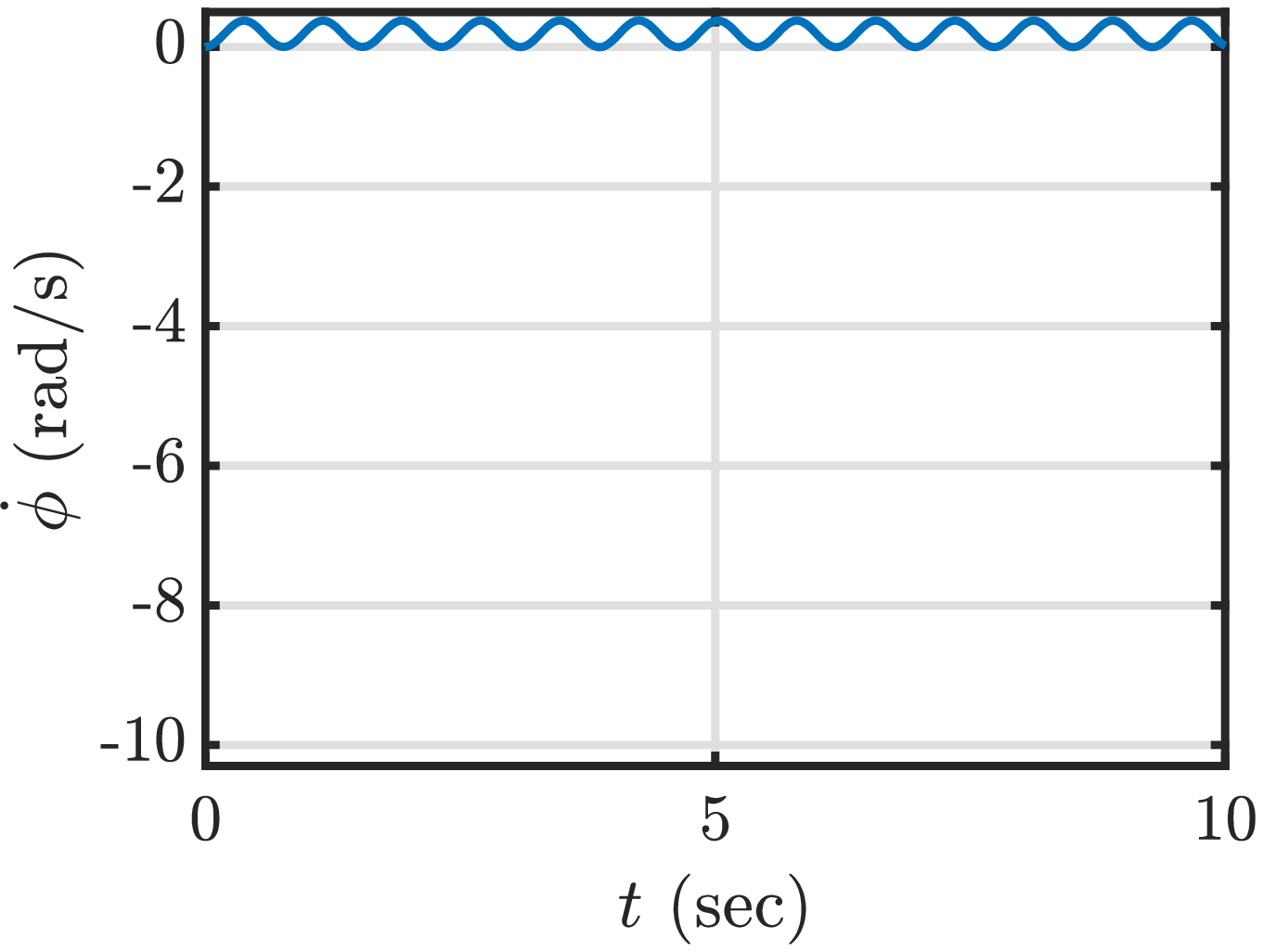}
  \caption{$\dot{\phi}$ vs. Time}
  \label{fig:15deg10radphid}
\end{subfigure}\hfil 
\begin{subfigure}{0.2\textwidth}
  \includegraphics[width=\linewidth, trim = 0.8cm 0 0.8cm 0]{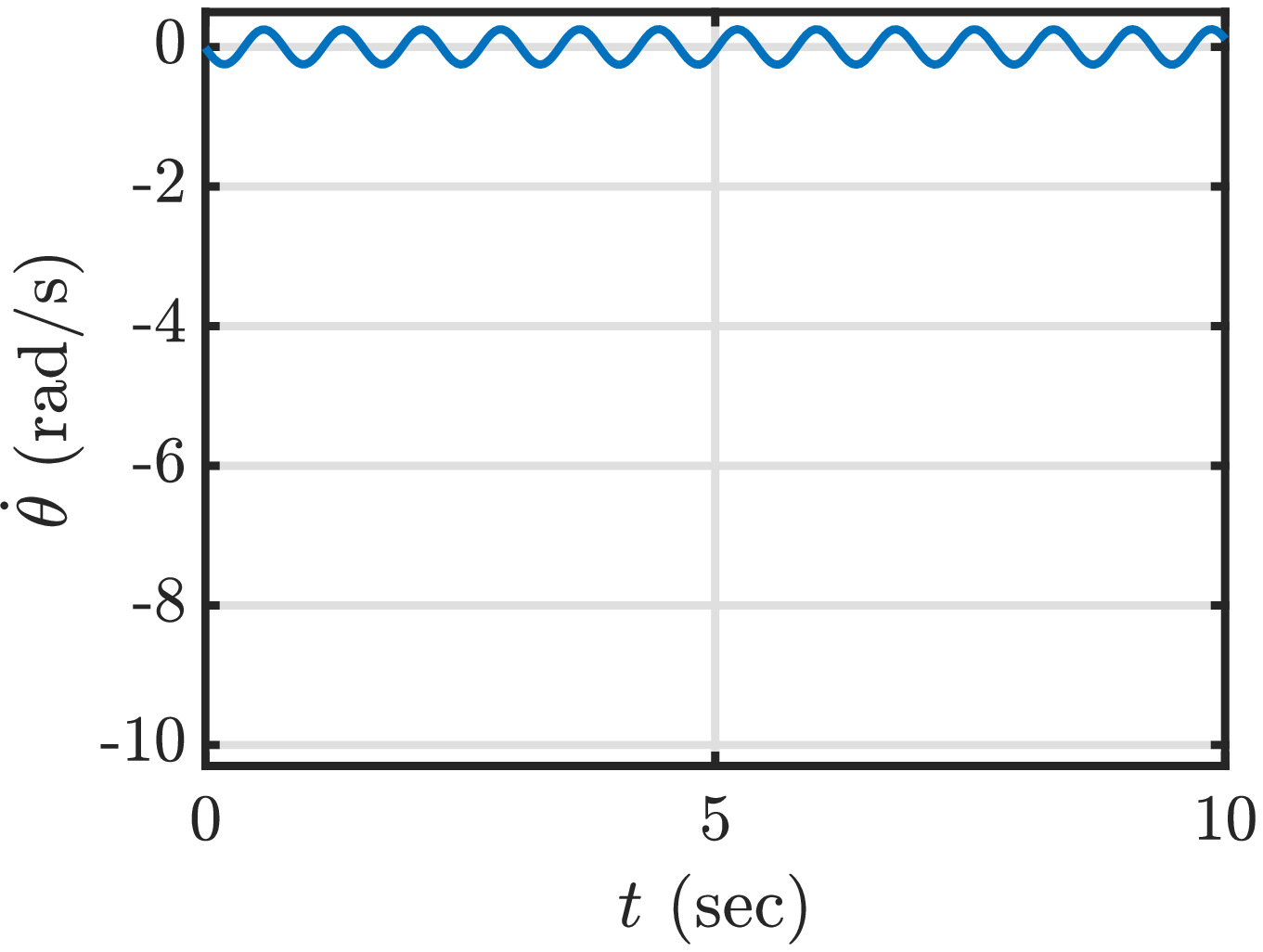}
  \caption{$\dot{\theta}$ vs. Time}
  \label{fig:15deg10radthetad}
\end{subfigure}\hfil 
\begin{subfigure}{0.2\textwidth}
  \includegraphics[width=\linewidth, trim = 0.8cm 0 0.8cm 0]{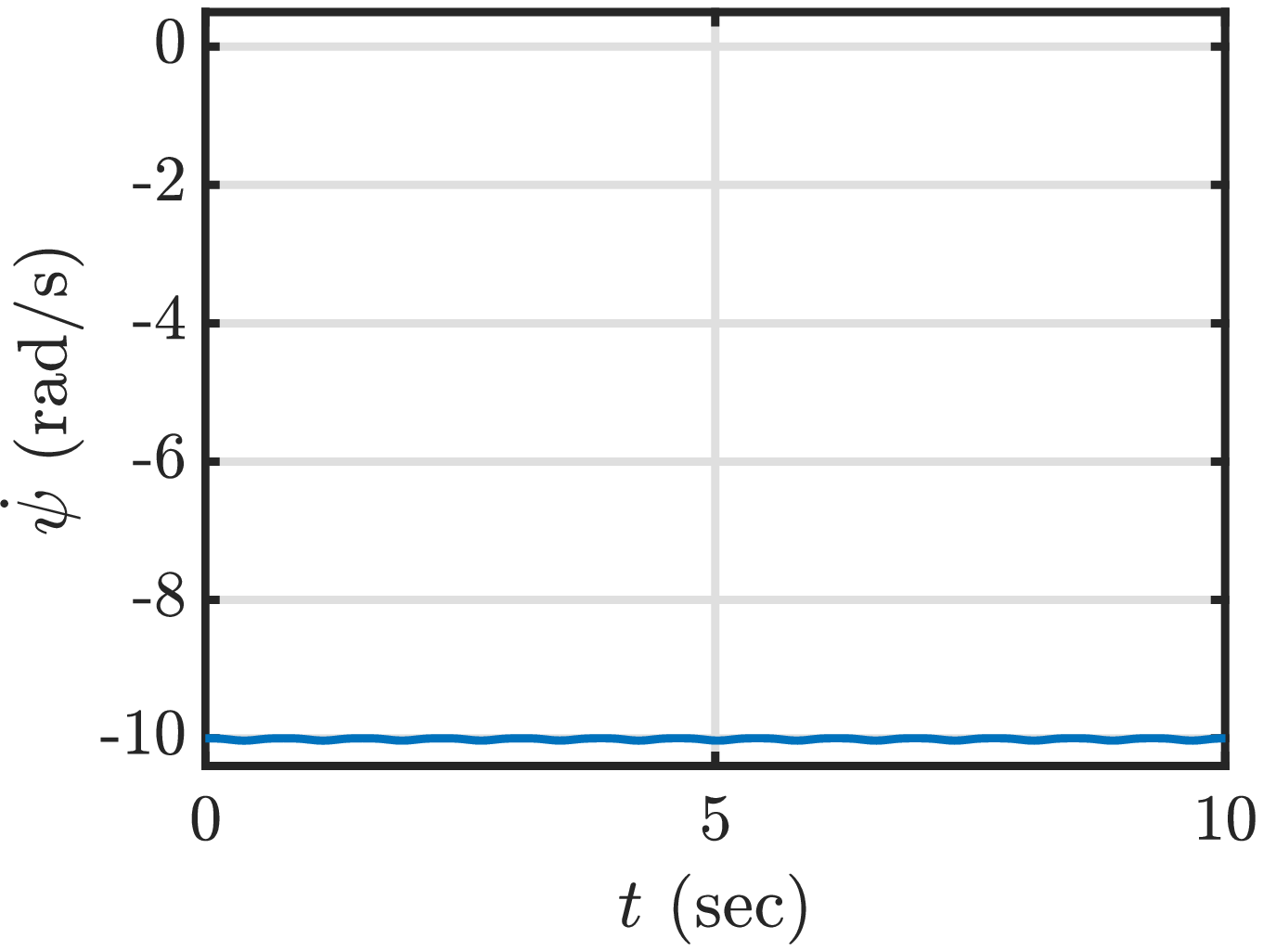}
  \caption{$\dot{\psi}$ vs. Time}
  \label{fig:15deg10radpsid}
\end{subfigure}\hfil 
\begin{subfigure}{0.2\textwidth}
  \includegraphics[width=\linewidth, trim = 0.8cm 0 0.8cm 0]{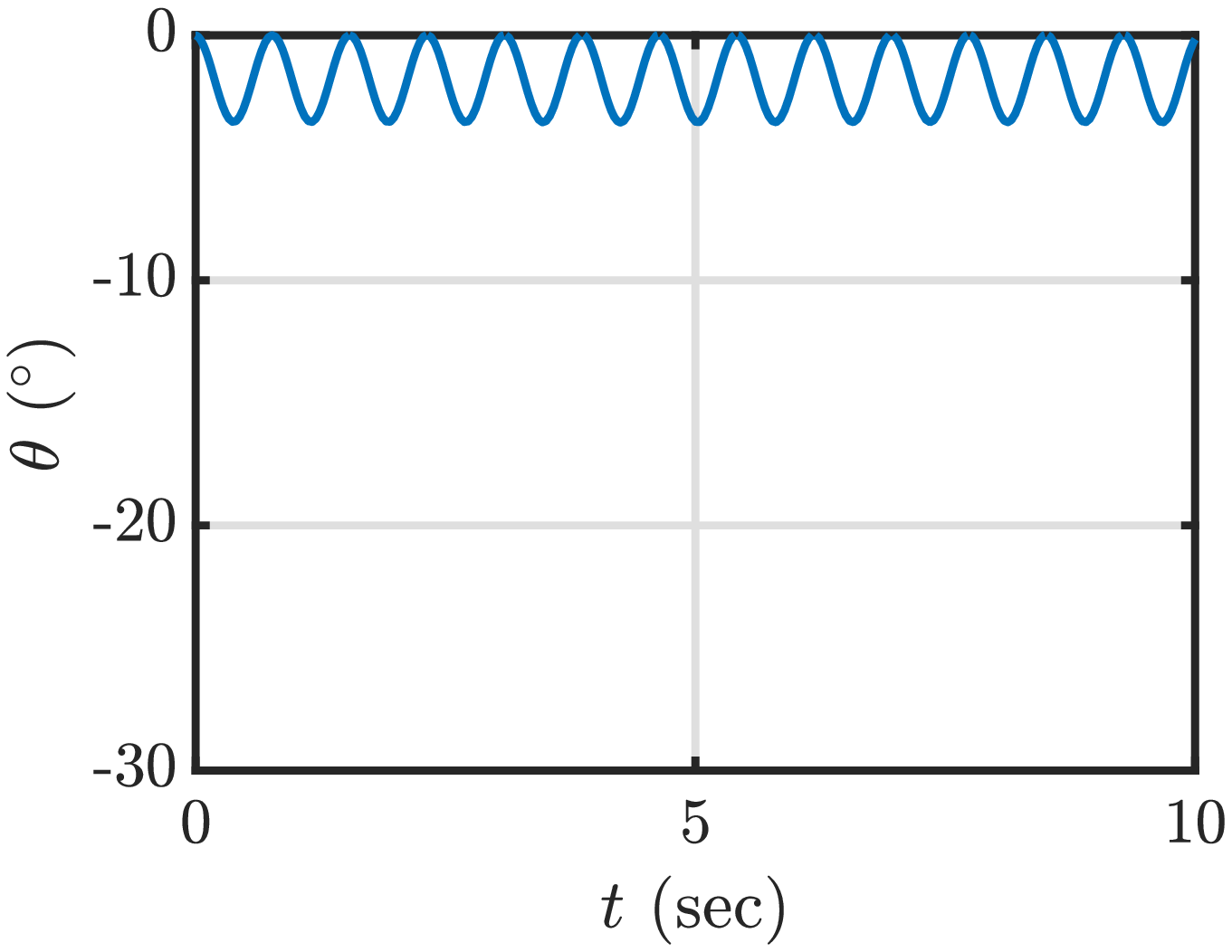}
  \caption{$\theta$ vs. Time}
  \label{fig:15deg10radtheta}
\end{subfigure}\hfil 

\caption{System response at $\beta$ =  15$^{\circ}$ and |$\dot{\psi}$| = 10 rad/s (high speed)}
\label{fig:15deg10rad}
\end{minipage}
\end{figure}

We further investigate the system response for the rate of change of heading angle $\dot{\phi}$, rate of change of lean angle $\dot{\theta}$, forward speed $\dot{\psi}$, and lean angle $\theta$ by simulating the steady state circular motion of the robot with the following four configurations of pendulum angle $\beta$ and forward rolling speed |$\dot{\psi}$|: 
\\
\\
\begin{tabular} { | p {3.8 cm} | p {4.2 cm} | p {4.4 cm} | }
\hline
\multicolumn{3} { | c | }{Configurations}\\
\hline
 & Low pendulum angle ($\beta$ = 5$^{\circ}$) & High pendulum angle ($\beta$ = 15$^{\circ}$)  \\
\hline
Low Speed (|$\dot{\psi}$| = 1 rad/s)  & Figure \ref{fig:5deg1rad}* & Figure \ref{fig:15deg1rad} \\
High speed (|$\dot{\psi}$| = 10 rad/s) & Figure \ref{fig:5deg10rad}* & Figure \ref{fig:15deg10rad}* \\
\hline
\end{tabular}
\\
\\
* Following observations can be made:

\begin{itemize} 
    \item Lean angle $\theta$ is very small in magnitude (refer figure \ref{fig:5deg1radtheta}, \ref{fig:5deg10radtheta}, \ref{fig:15deg10radtheta}) 

    \item Magnitude of $\dot{\phi}$ and  $\dot{\theta}$ are small compared to $\dot{\psi}$ (refer figure \ref{fig:5deg1rad}, \ref{fig:5deg10rad}, \ref{fig:15deg10rad}) 
    
    \item $\dot{\psi}$ remains approximately constant (refer figure \ref{fig:5deg1radpsid}, \ref{fig:5deg10radpsid}, \ref{fig:15deg10radpsid}) 
\end{itemize}

\section{Steady state analysis of wobbly circular motion} \label{analysisOfWobblyCircle}

This section simplifies and linearizes the dynamics of the circular steady-state motion whose configuration was discussed in section \ref{og_system_circle_response}. This leads to the development of expressions for wobble amplitude, wobble frequency, precession rate, and radius of curvature. Then, the system response of the original system is compared to the system response of the simplified system.

\subsection{Model simplification}

The 3D mathematical model of the robot derived in section \ref{modeling} is highly complex and nonlinear. The constituent equations of the model can be simplified by taking the following approximations: 

\begin{itemize}    
    \item Pendulum angle $\beta$ is held constant, i.e. $\dot{\beta}$ $\approx$ 0, $\ddot{\beta}$ $\approx$ 0
    \item Centre of mass of yoke is assumed to be at the hull centre, i.e. $r_y = 0$.
\end{itemize}

Using these approximations, equations (\ref{xdot_eq}),  (\ref{zdot_eq}) and (\ref{lagrangeEqn}) reduce to the following three equations: 

\begin{multline}
    \bigg[ I_h + \frac{I_p}{2} + \frac{3I_y}{2} + \frac{I_y \cos(2\theta)}{2} - \frac{I_p \cos(2\beta + 2\theta)}{2} + \frac{m_p r_p^2}{2}(1 - \cos(2\beta + 2\theta))\bigg] \ddot{\phi} - \bigg[ I_h \sin(\theta) \\ 
    - \frac{m_p r_p r_h}{2}(\sin(\beta) + \sin(\beta + 2\theta))\bigg] \ddot{\psi} -  \bigg[I_h \cos(\theta) - \frac{m_p r_p r_h}{2}(\cos(\beta + 2\theta) - \cos(\beta))\bigg] \dot{\psi} \dot{\theta} \\ - \bigg[ I_y \sin(2\theta) - I_p \sin(2\beta + 2\theta) - m_p r_p^2 \sin(2\beta + 2\theta) + m_p r_p r_h \sin(\beta + \theta)\bigg] \dot{\phi} \dot{\theta} = 0
\label{long_eq_1}
\end{multline}
\begin{multline}
    \bigg[ (I_p + I_h + I_y +m_p r_p^2 + (m_p + m_y + m_h) r_h^2 - 2 m_p r_p r_h \cos(\beta + \theta)\bigg] \ddot{\theta} + \bigg[ m_p r_p r_h \sin(\beta + \theta)\bigg] \dot{\theta}^2\\ + \bigg[ I_h \cos(\theta) + (m_p + m_y + m_h) r_h^2 \cos(\theta) -  \frac{m_p r_p r_h}{2} (\cos(\beta + 2\theta) +  \cos(\beta)) \bigg] \dot{\phi} \dot{\psi} + m_p r_p g \sin(\beta + \theta) \\ + \bigg[\frac{I_y \sin(2\theta)}{2} - \frac{I_p \sin(2\beta + 2\theta)}{2} - \frac{m_p r_p^2 \sin(2\beta + 2\theta)}{2} + m_p r_p r_h \sin(\beta + \theta) \bigg] \dot{\phi}^2  = 0
\label{long_eq_2}    
\end{multline}
\begin{multline}
    \bigg[ I_h + (m_p + m_y + m_h) r_h^2 \cos^2(\theta) \bigg] \ddot{\psi} - \bigg[ \frac{(m_p + m_y + m_h) r_h^2}{2}\sin(2\theta)\bigg] \dot{\psi} \dot{\theta} \\ -  \bigg[ I_h \cos(\theta) + (m_p + m_y + m_h) r_h^2 \cos(\theta) - 2m_p r_p r_h \cos(\beta)\cos^2(\theta) +  2m_p r_p r_h \sin(\beta)\cos(\theta)\sin(\theta) \bigg] \dot{\phi} \dot{\theta} \\ - \bigg[ I_h \sin(\theta) - m_p r_p r_h \sin(\beta)\cos^2(\theta) - m_p r_p r_h \cos(\beta)\cos(\theta)\sin(\theta) \bigg] \ddot{\phi} = 0
\label{long_eq_3}
\end{multline}


Equations (\ref{long_eq_1}), (\ref{long_eq_2}) and (\ref{long_eq_3}) constitute the mathematical model of the system at all pendulum angles and all forward speeds. These equations can be further simplified by taking the following approximations:

\begin{itemize}
    \item  Pendulum angle $\beta$ is small in magnitude ($\sin\beta$ $\approx$ $\beta$ and $\cos\beta$ $\approx$ 1).
    \item  Lean angle $\theta$ is small in magnitude ($\sin\theta$ $\approx$ $\theta$ and $\cos\theta$ $\approx$ 1).
    \item Forward speed $\dot{\psi}$ is constant.
    \item $\dot{\psi}$ is larger than $\dot{\phi}$ and $\dot{\theta}$. Hence $\dot{\phi}\dot{\theta}$, $\dot{\phi}^2$, $\dot{\theta}^2$ can be neglected in comparison to $\dot{\psi}$.
      
\end{itemize}

These approximations are used to further simplify equations (\ref{long_eq_1}), (\ref{long_eq_2}), and (\ref{long_eq_3}) to give: 

\begin{equation}
    \ddot{\phi} = \frac{I_h}{I_h + 2I_y}\dot{\psi}\dot{\theta}
    \label{smalltheta_phidd}
\end{equation}
\begin{multline}
    \bigg[ I_p + I_h + I_y +m_p r_p^2 + (m_p + m_y + m_h) r_h^2 - 2 m_p r_p r_h \bigg] \ddot{\theta} \\ + \bigg[ I_h + (m_p + m_y + m_h) r_h^2 -  m_p r_p r_h \bigg] \dot{\phi} \dot{\psi} + m_p r_p g (\beta + \theta) = 0
    \label{smalltheta_thetadd}
\end{multline}
\begin{equation}
    \ddot{\psi} = 0
    \label{smalltheta_psidd}
\end{equation}

\subsection{Formulation of circular motion characteristics}

A spherical robot's circular motion characteristics include precession rate, wobble amplitude, wobble frequency, and radius of curvature. The expressions for these quantities can now be obtained using the simplified model.

\begin{description}
    \item[Precession rate] 
Using constant $\dot{\psi}$, equation (\ref{smalltheta_phidd}) can be integrated to obtain $\dot{\phi}$.

\begin{equation}
    \dot{\phi} = \frac{I_h}{I_h + 2I_y}\dot{\psi}\theta + c
    \label{smalltheta_phid}
\end{equation}
where $c$ is a constant of integration. We assume that the robot starts with initial $\theta$ and $\dot{\phi} = 0$, which gives $c=0$. 

\item[Wobbling] 
From equations (\ref{smalltheta_thetadd}) and (\ref{smalltheta_phid}), we get

\begin{multline}
    \bigg[ I_p + I_h + I_y +m_p r_p^2 + (m_p + m_y + m_h) r_h^2 - 2 m_p r_p r_h \bigg] \ddot{\theta} \\ + \bigg[\frac{I_h(I_h + (m_p + m_y + m_h) r_h^2 -  m_p r_p r_h )}{I_h + 2I_y}\dot{\psi}^2 + m_p r_p g \bigg]\theta = - m_p r_p g \beta 
    \label{smallthetathetadd2}
\end{multline}

Equation (\ref{smallthetathetadd2}) is a standard second-order equation, the solution of which is given by: 

\begin{equation}
    \theta = A (1 - \cos(\omega t))
    \label{smallthetatheta}
\end{equation}

where $A$ is the amplitude of oscillations and $\omega$ is the frequency of oscillations given by: 
\begin{equation}
    A = \frac{- m_p r_p g \beta}{\frac{I_h(I_h + (m_p + m_y + m_h) r_h^2 -  m_p r_p r_h )}{I_h + 2I_y}\dot{\psi}^2 + m_p r_p g}
    \label{theta_amplitude}
\end{equation} 

\begin{equation}
    \omega = \sqrt{\frac{m_p r_p g  + \frac{I_h (I_h + (m_p + m_y + m_h) r_h^2  -  m_p r_p r_h )}{I_h + 2I_y} \dot{\psi}^2}{I_p + I_h + I_y +m_p r_p^2 + (m_p + m_y + m_h) r_h^2 - 2 m_p r_p r_h }}
    \label{thetafreq}
\end{equation}

\item[Radius of curvature]
For a sphere moving in a circular trajectory with pure rolling, the radius of curvature $\rho$ can be approximated as

\begin{equation}
    Time = \frac{2\pi}{\dot{\phi}_{mean}} = \frac{2\pi \rho}{\dot{\psi}r_h} \implies
    \rho = \frac{\dot{\psi} r_h}{\dot{\phi}_{mean}}
    \label{time_eqn}
\end{equation}

\end{description}

\begin{equation}
    \rho = \frac{ - r_h \bigg([I_h+(m_p+m_y+m_h)r_h^2-m_p r_p r_h ] I_h \dot{\psi}^2 + [ I_h+2I_y ] m_p r_p g \bigg)}{ m_p r_p g I_h \beta} 
    \label{radiusOfCurv}
\end{equation}

The dynamics of the circular steady-state motion can be further simplified based on forward speed: Low forward speeds and High forward speeds.

\subsubsection{Low forward Speeds} \label{slow_assumptions}

The ratio of $r_h \dot{\psi}^2$ to $g$ is relatively small at low forward speeds. Consequently, we can neglect the terms containing $r_h \dot{\psi}^2$. This approximation further simplifies the equations (\ref{theta_amplitude}), (\ref{thetafreq}), and (\ref{radiusOfCurv}) to yield the following expressions: 

\begin{description}
    
    \item[Wobbling] The expressions for wobble amplitude and frequency become:
    \begin{equation}
        A = -\beta
        \label{lowspeed_amplitude}
    \end{equation}
    
    \begin{equation}
        \omega = \sqrt{\frac{m_p r_p g}{I_p + I_h + I_y +m_p r_p^2 + (m_p + m_y + m_h) r_h^2 - 2 m_p r_p r_h}}
        \label{lowspeed_frequency}
    \end{equation}    

    \item[Radius of curvature]

    \begin{equation}
    \rho = \frac{ - r_h ( I_h+2I_y)}{ I_h \beta} 
    \label{lowspeed_radiusOfCurv}
    \end{equation}
    
\end{description}

\subsubsection{High forward Speed}


At high forward speeds, the ratio of $g$ to $r_h \dot{\psi}^2$ is relatively small. Consequently, we can neglect the terms containing $g$. This approximation further simplifies the equations (\ref{theta_amplitude}), (\ref{thetafreq}), and (\ref{radiusOfCurv}) to yield the following expressions:  

\begin{description}
    \item[Wobbling] The expressions for wobble amplitude and frequency become:
    \begin{equation}
        A = - \frac{(I_h + 2I_y)(m_p r_p g) \beta}{I_h(I_h + (m_p + m_y + m_h) r_h^2 -  m_p r_p r_h ) \dot{\psi}^2}
        \label{highspeed_theta_amplitude}
    \end{equation} 
    
    \begin{equation}
        \omega = \dot{\psi} \sqrt{\frac{ I_h (I_h + (m_p + m_y + m_h) r_h^2  -  m_p r_p r_h )}{(I_h + 2I_y)(I_p + I_h + I_y +m_p r_p^2 + (m_p + m_y + m_h) r_h^2 - 2 m_p r_p r_h)}}
        \label{highspeed_thetafreq}
    \end{equation} 

    \item[Radius of curvature]

    \begin{equation}
        \rho =  - \bigg( \frac{I_h+(m_p+m_y+m_h)r_h^2-m_p r_p r_h }{ m_p r_p g \beta} \bigg) r_h \dot{\psi}^2
        \label{highspeed_radiusOfCurv}
    \end{equation}
 
\end{description}

\subsection{Analysis of circular motion characteristics} \label{model_simplification_results}

In this section, we analyze the expressions for wobble amplitude and frequency, precession rate, and radius of curvature to comprehend their nature, dependence on parameters such as $\beta$ and $\dot{\psi}$, and compare them to the system response derived from the original model.

\begin{figure}
    \centering 
\begin{subfigure}{0.2\textwidth}
  \centering
  \includegraphics[width=\linewidth, trim = 1cm 0 1cm 0]{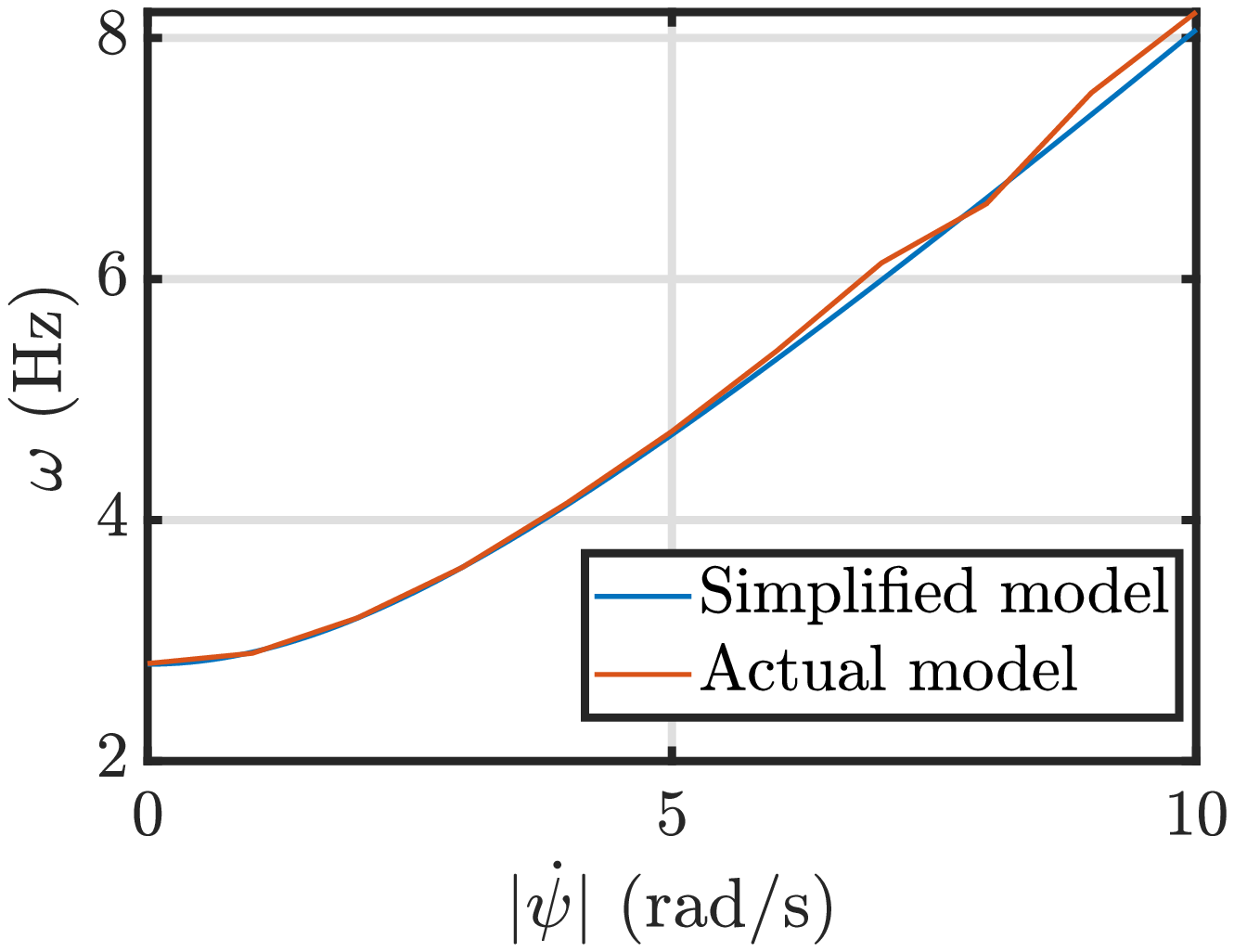}
  \caption{\centering{Wobble frequency}}
  \label{fig:Wobfreqvsforspeed}
\end{subfigure}\hfil 
\begin{subfigure}{0.2\textwidth}
  \centering
  \includegraphics[width=\linewidth, trim = 1cm 0 1cm 0]{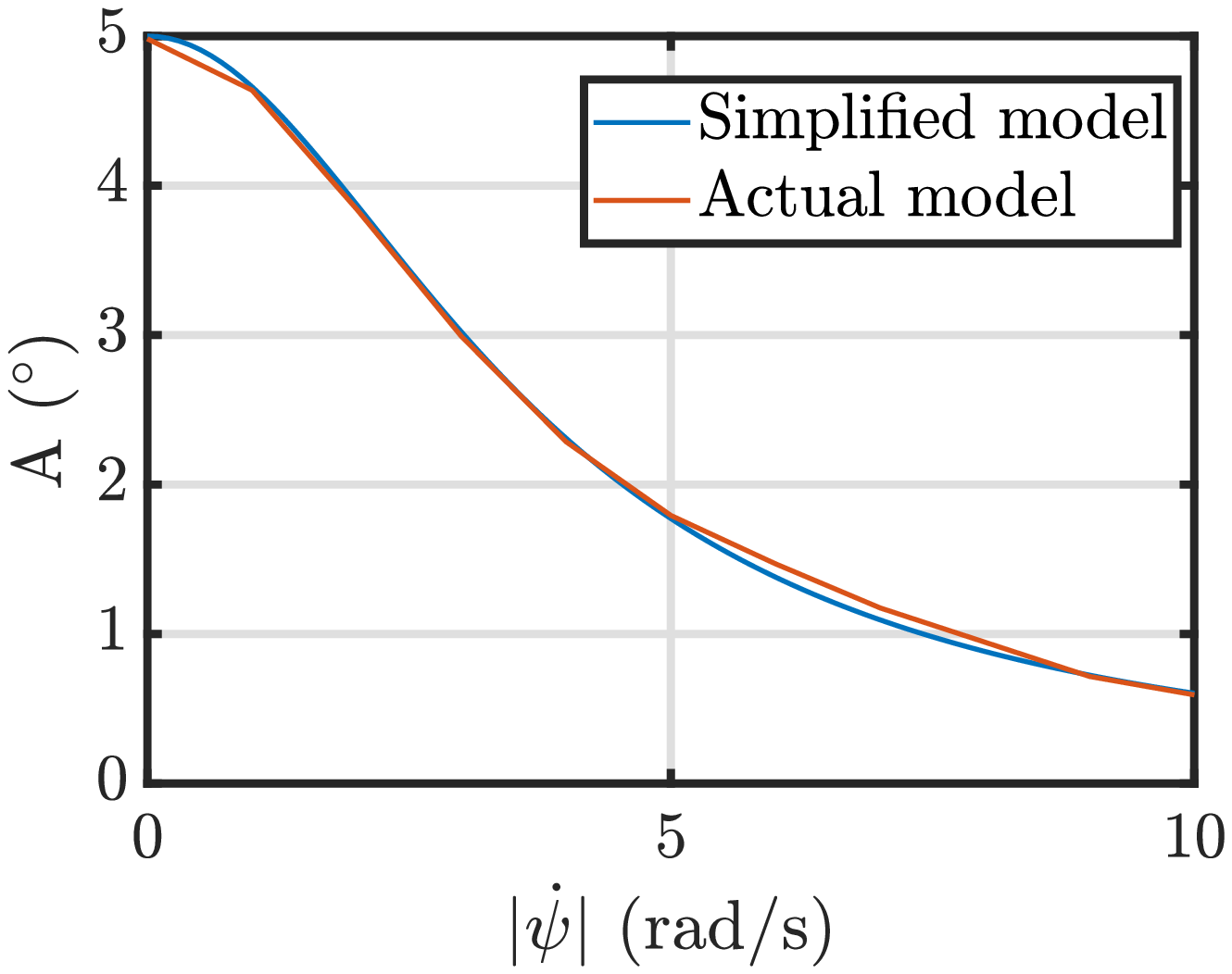}
  \caption{\centering{Wobble amplitude}}
  \label{fig:Wobamplvsforspeed}
\end{subfigure}\hfil 
%
%
\begin{subfigure}{0.2\textwidth}
  \centering
  \includegraphics[width=\linewidth, trim = 1cm 0 1cm 0]{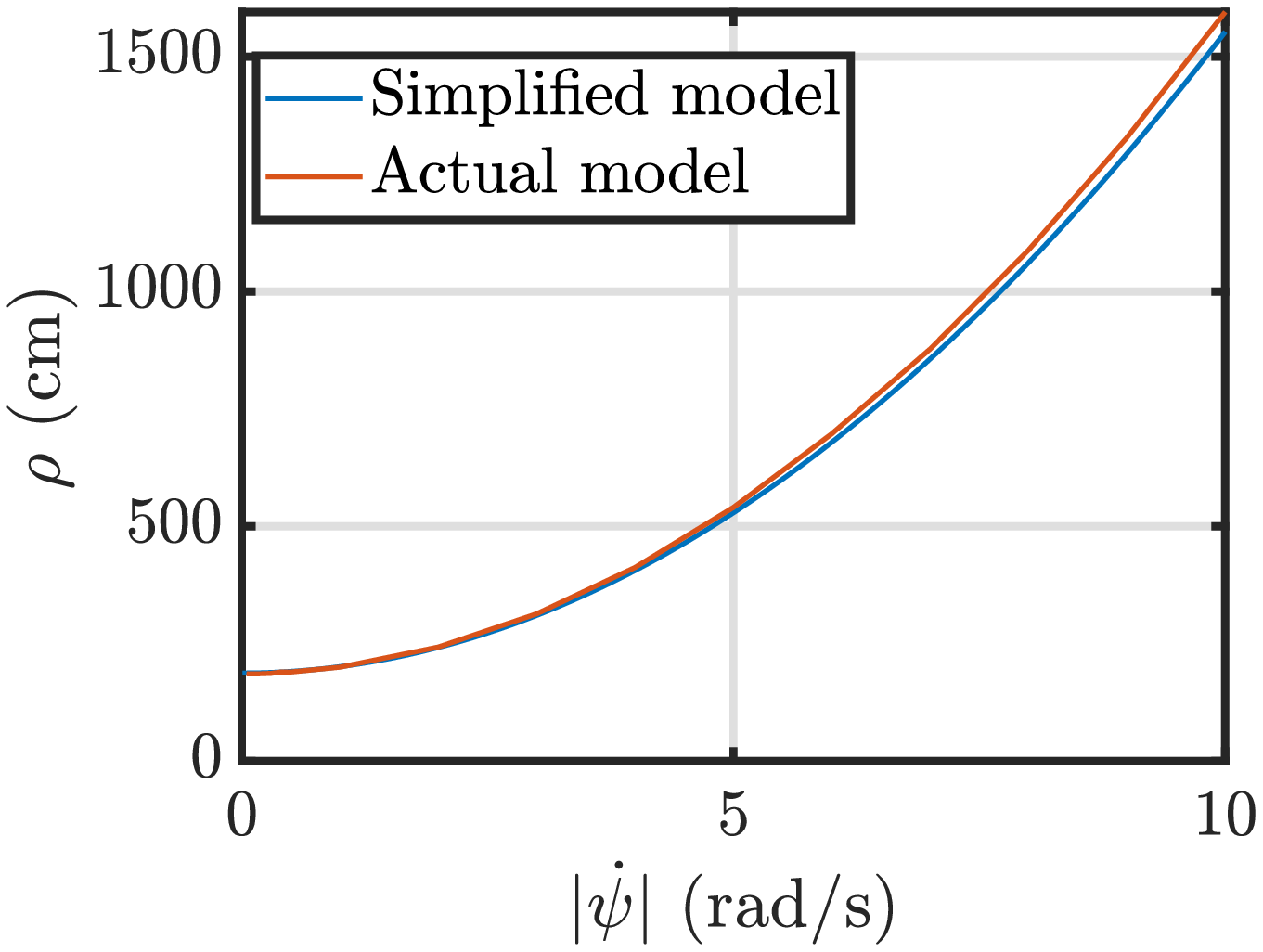}
  \caption{\centering{Radius of curvature}}
  \label{fig:RadiusvsForspeed}
\end{subfigure}\hfil 
\begin{subfigure}{0.2\textwidth}
  \centering
  \includegraphics[width=\linewidth, trim = 1cm 0 1cm 0]{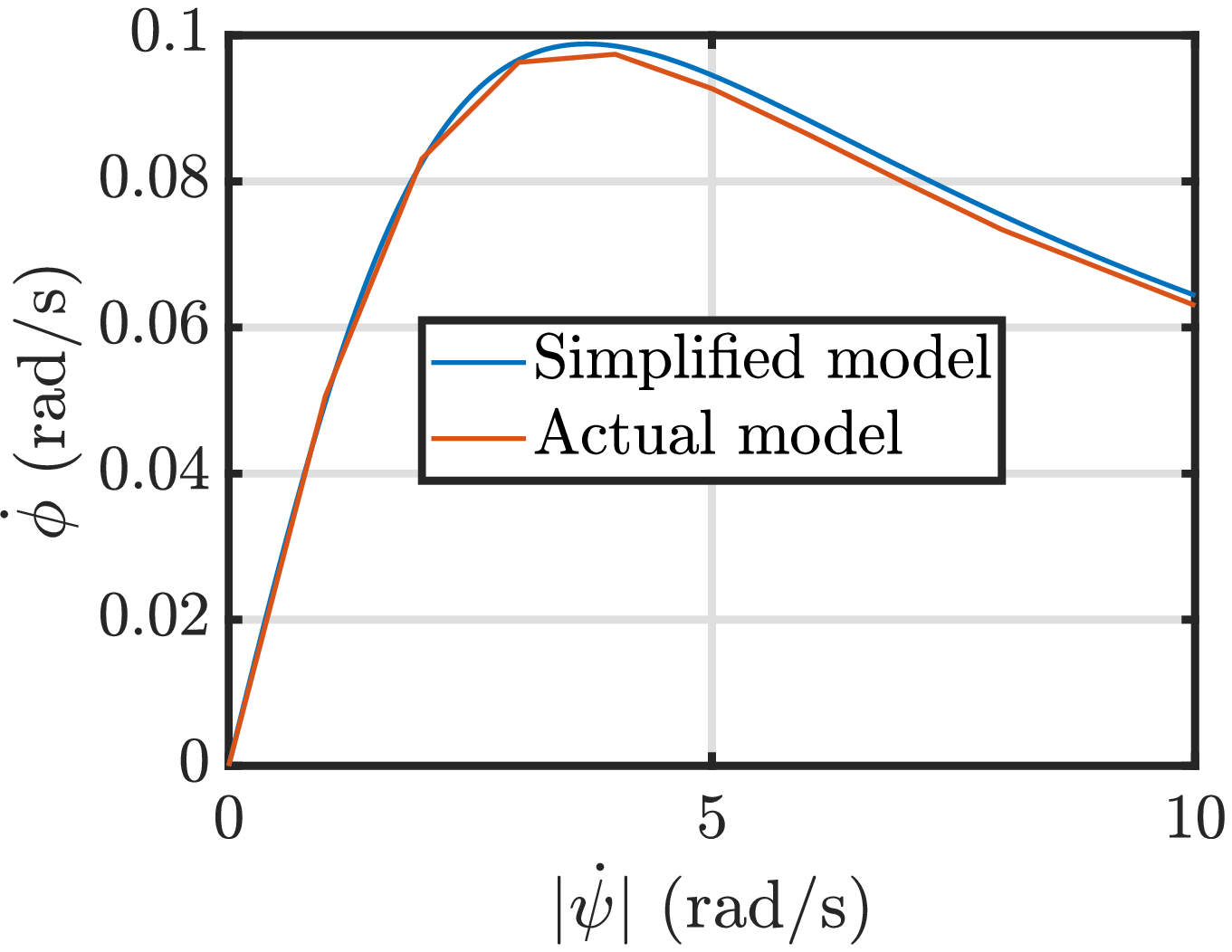}
  \caption{\centering{Precession rate}}
  \label{fig:Precvsforspeed}
\end{subfigure}\hfil 

\caption{Circular motion characteristics vs forward speed $\dot{\psi}$ at a constant pendulum angle ($\beta$ = $5^{\circ}$)}
\label{fig:wobbling_trends_speed}
\end{figure}

\begin{figure}
    \centering 
\begin{subfigure}{0.2\textwidth}
  \centering
  \includegraphics[width=\linewidth, trim = 1cm 0 1cm 0]{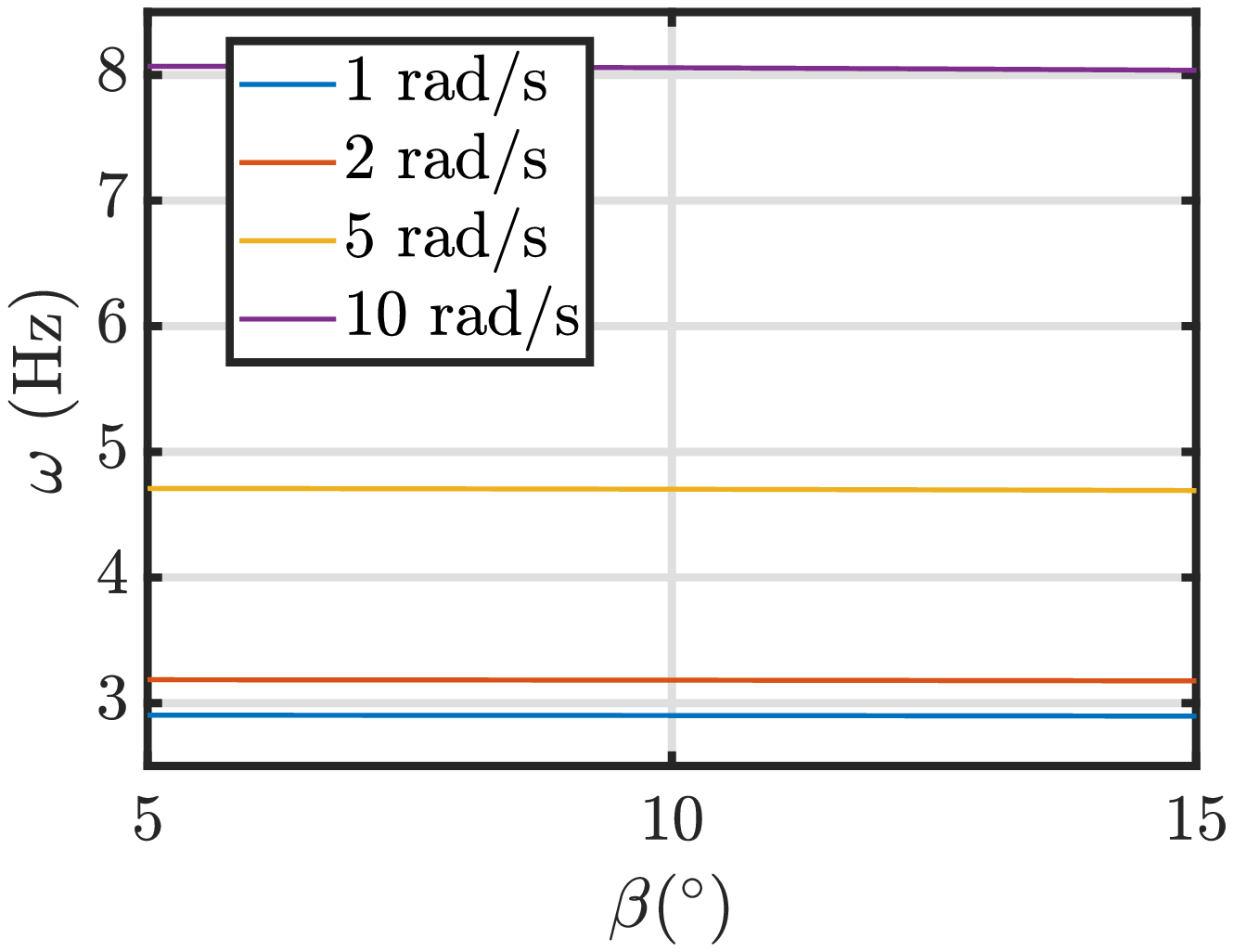}
  \caption{\centering{Wobble frequency}}
  \label{fig:Wobfreqvspendangle}
\end{subfigure}\hfil 
\begin{subfigure}{0.2\textwidth}
  \centering
  \includegraphics[width=\linewidth, trim = 1cm 0 1cm 0]{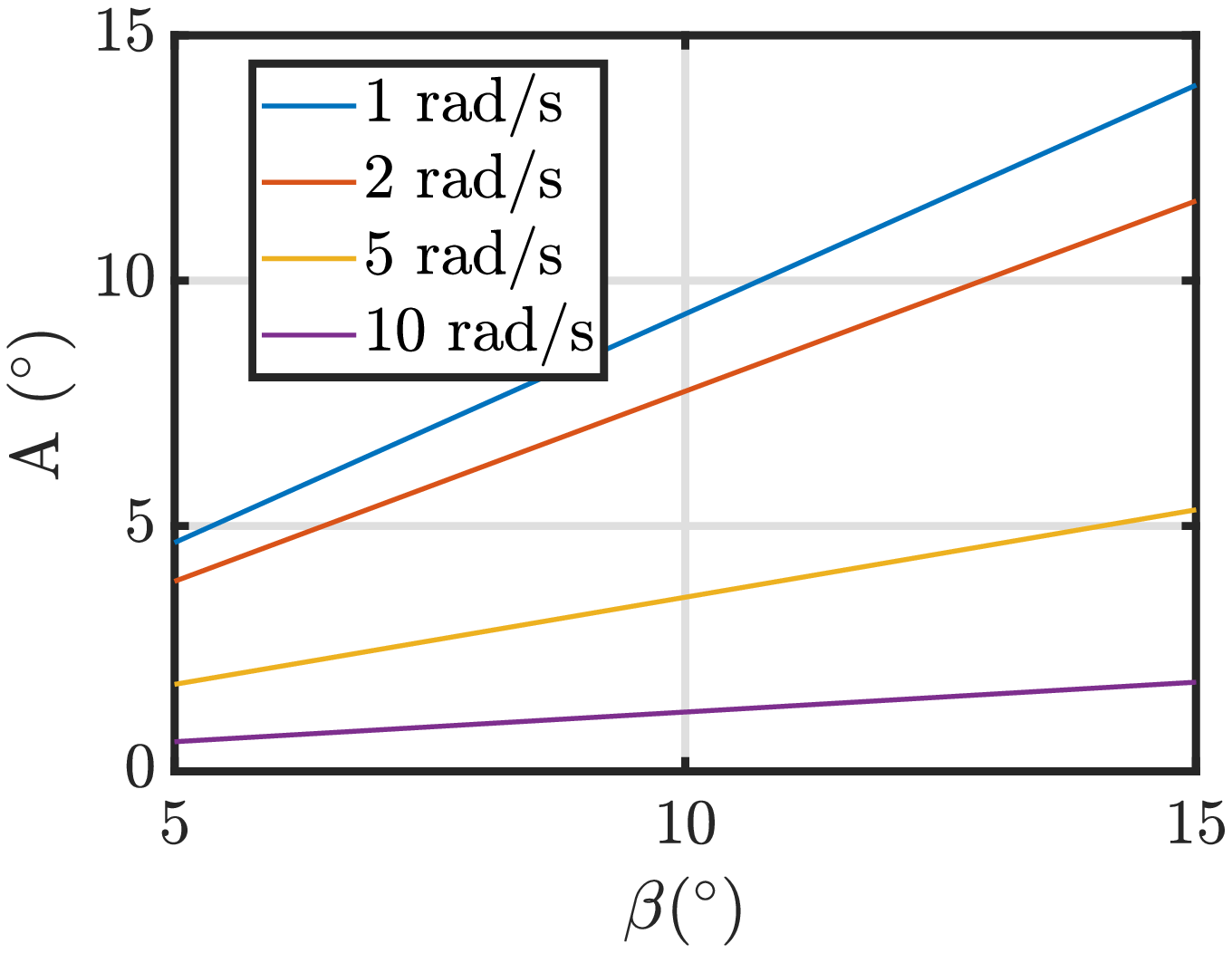}
  \caption{\centering{Wobble amplitude}}
  \label{fig:Wobamplvspendangle}
\end{subfigure}\hfil 
%
%
\begin{subfigure}{0.2\textwidth}
  \centering
  \includegraphics[width=\linewidth, trim = 1cm 0 1cm 0]{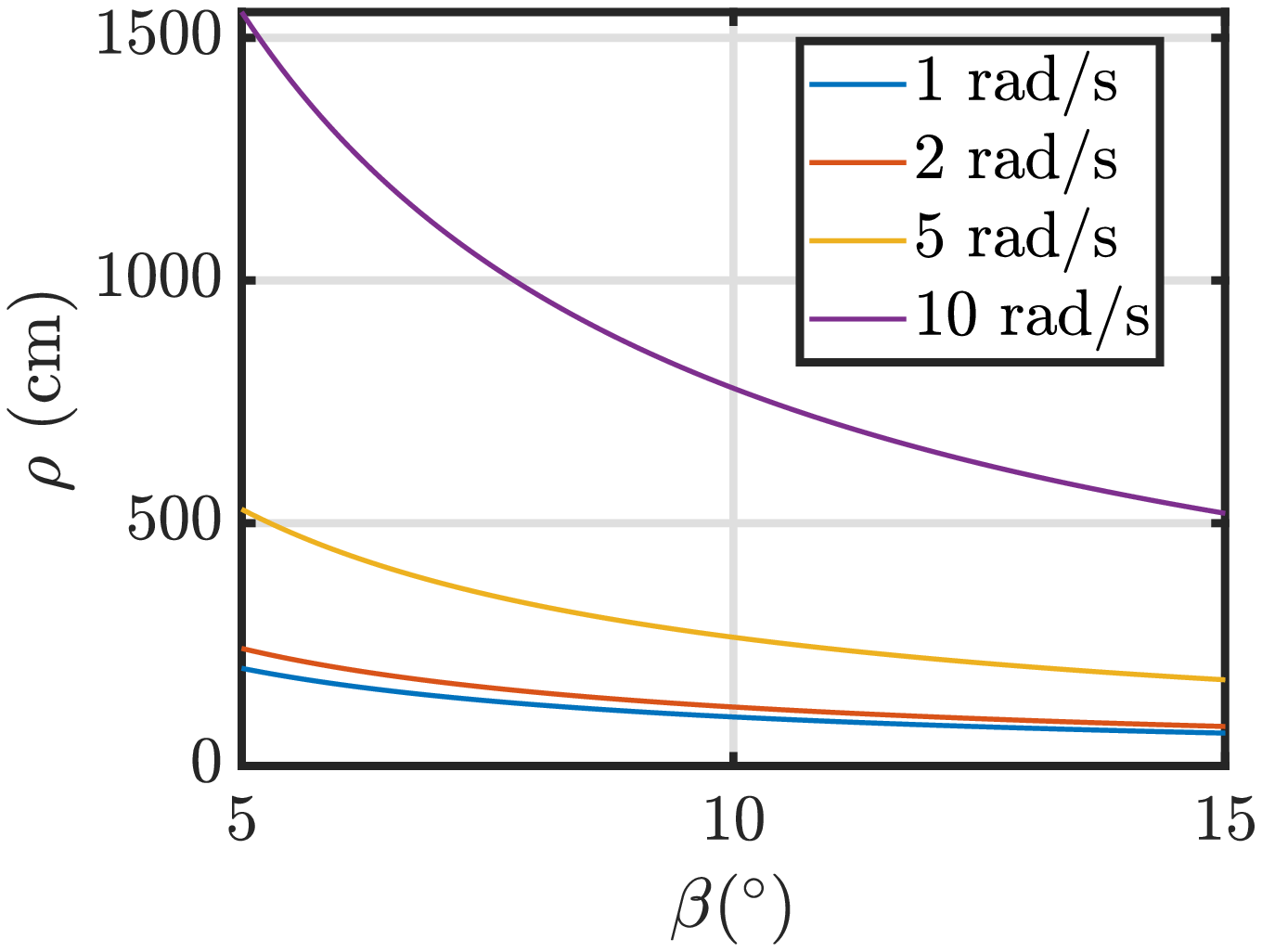}
  \caption{\centering{Radius of curvature}}
  \label{fig:Radiusvspendangle}
\end{subfigure}\hfil 
\begin{subfigure}{0.2\textwidth}
  \centering
  \includegraphics[width=\linewidth, trim = 1cm 0 1cm 0]{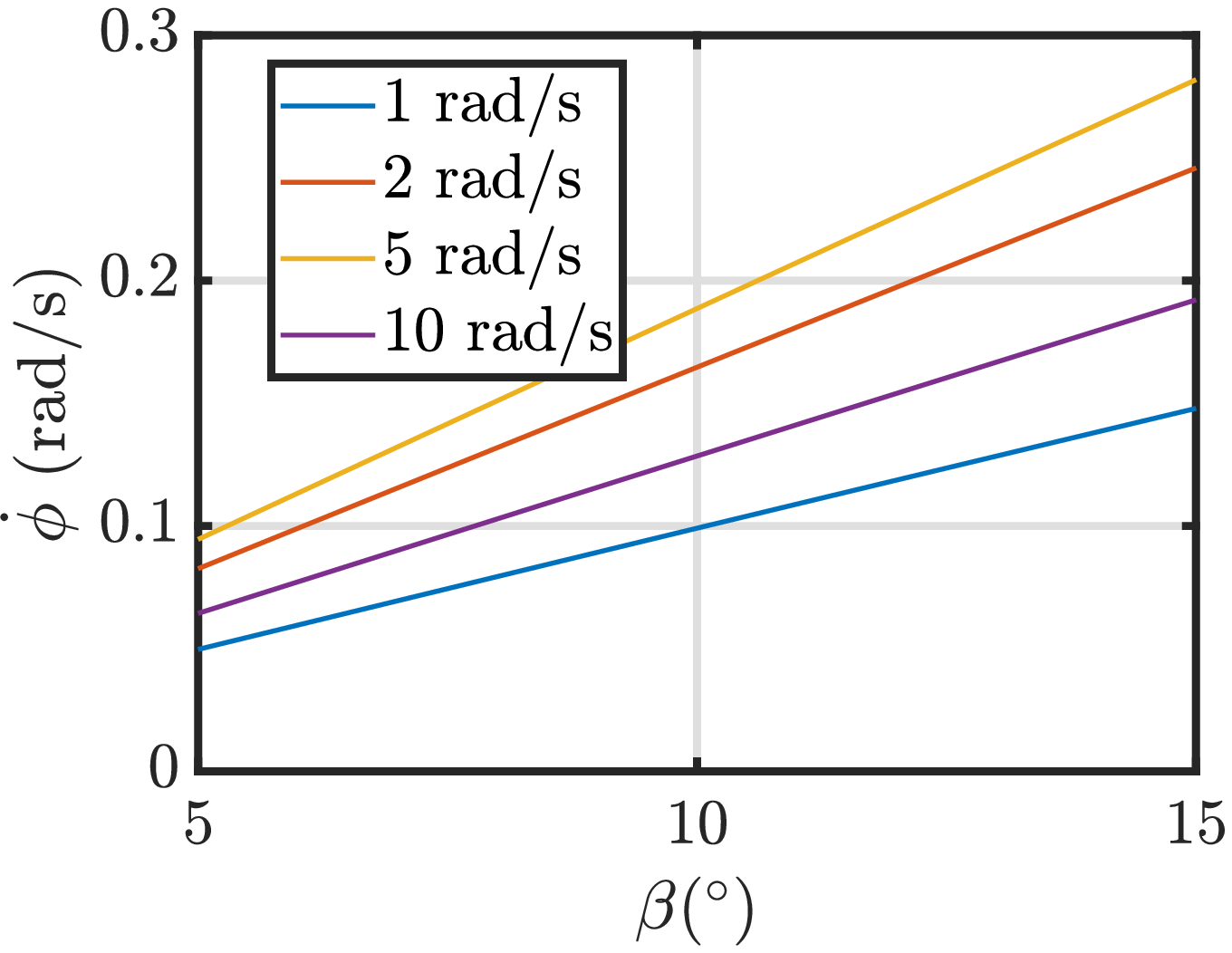}
  \caption{\centering{Precession rate}}
  \label{fig:Precvspendangle}
\end{subfigure}\hfil 

\caption{Circular motion characteristics vs pendulum angle $\beta$ at different forward speeds $\dot{\psi}$}
\label{fig:wobbling_trends_beta}
\end{figure}

\begin{description}
    
    \item[Wobbling] The spherical robot's wobbling or lateral oscillations are characterized by the lean angle $\theta$. Based on the expression (\ref{smallthetatheta}), $\theta$ is sinusoidal with oscillations centered away from the origin. The mean value of $\theta$ equals the amplitude of oscillations. This suggests that the magnitude of $\theta$ oscillations around the mean position increases as the mean value of $\theta$ moves further from the origin.

    The relationship between wobble frequency and forward speed at a constant pendulum angle of $5^{\circ}$ is depicted in figure \ref{fig:Wobfreqvsforspeed}. The wobble frequency is found to be nearly constant at low forward speeds, a behavior confirmed by the equation (\ref{lowspeed_frequency}). From equation (\ref{highspeed_thetafreq}) and figure \ref{fig:Wobfreqvsforspeed}, it can be seen that at high forward speeds, the wobble frequency increases approximately linearly with forward speed. The figure also illustrates the similarity between the system response for wobble frequency generated by the original model and the simplified model equation (\ref{thetafreq}) at different speeds. 
    
    The relationship between wobble frequency and pendulum angle for a range of constant forward speeds is shown graphically in figure \ref{fig:Wobfreqvspendangle}, which is based on equation (\ref{thetafreq}). It can be inferred from the figure that for any constant forward speed, the wobble frequency is independent of the pendulum angle. We can also observe that the wobble frequency is greater at higher speeds for any given pendulum angle. 
    
    The relationship between wobble amplitude and forward speed for a constant pendulum angle of $5^{\circ}$ is illustrated in figure \ref{fig:Wobamplvsforspeed}. This figure and equation (\ref{theta_amplitude}) demonstrate that the amplitude of oscillations in $\theta$ varies inversely with $\dot{\psi}^2$. Figure \ref{fig:Wobamplvsforspeed} also demonstrates that lateral oscillations are significantly reduced at high speeds, a behavior confirmed by the equation (\ref{highspeed_theta_amplitude}). The figure also illustrates the similarity between the system response for the amplitude of lateral oscillations generated by the original model and the simplified model equation (\ref{theta_amplitude}) at different speeds.

    The relationship between wobble amplitude and pendulum angle for a range of constant forward speeds is shown graphically in figure \ref{fig:Wobamplvspendangle}, which is based on equation (\ref{theta_amplitude}). For constant forward speed, the wobble amplitude is directly proportional to pendulum angle $\beta$. We can also observe that wobble amplitude decreases with increasing speed for any given pendulum angle. 

\end{description}

\begin{description}
    \item[Radius of curvature]

    According to the equation (\ref{radiusOfCurv}), the radius of curvature depends on the angle of the pendulum and forward speed. If we control these two variables, we can indirectly affect the radius of curvature of the robot. This can facilitate the robot's movement along a curved path.
    
     Figure \ref{fig:RadiusvsForspeed} illustrates the relationship between the radius of curvature and forward speed for a constant pendulum angle of $5^{\circ}$. At low forward speeds, the radius of curvature is observed to be nearly constant, a behavior confirmed by the equation (\ref{lowspeed_radiusOfCurv}). The expression (\ref{highspeed_radiusOfCurv}) for the radius of curvature at high speed is comparable to the radius of curvature expressions reported in the literature \cite{rad_curv1,rad_curv2} for all speed ranges. Equation (\ref{highspeed_radiusOfCurv}) demonstrates that the radius of curvature at high forward speeds is directly proportional to the square of the forward speed. Consequently, higher speeds result in a larger turning radius. Figure \ref{fig:RadiusvsForspeed} also compares the radius of curvature values generated by the original model with those generated by the simplified model (refer equation (\ref{radiusOfCurv})). 

    The relationship between the radius of curvature and pendulum angle for a range of constant forward speeds is shown graphically in figure \ref{fig:Radiusvspendangle}, which is based on Equation (\ref{radiusOfCurv}). The radius of curvature is inversely proportional to the pendulum angle for constant forward speed. Therefore, the robot makes sharper turns as the pendulum tilts further. For any given pendulum angle, we can see that the radius of curvature increases with higher speed. We can also observe that the radius of curvature varies significantly as the pendulum's angle changes at high speeds. 

\end{description}

\begin{description}
    \item[Precession rate] The robot's precession rate indicates how quickly it completes one complete revolution as it moves in a circle. According to the equation (\ref{smalltheta_phid}), the magnitude of the precession rate is directly proportional to the lean angle. Consequently, the precession rate oscillates when the robot's motion involves lateral oscillations. This section analyzes the precession rate's mean value to comprehend its behavior in relation to forward velocity and pendulum angle. 

    The relationship between the mean precession rate and forward speed for a constant pendulum angle of $5^{\circ}$ is illustrated in figure \ref{fig:Precvsforspeed}. Figure  \ref{fig:Precvsforspeed} and equation (\ref{smalltheta_phid}) illustrate how the mean precession rate increases at low forward speeds but is inversely proportional to $\dot{\psi}$ at high forward speeds. Equation (\ref{time_eqn}) represents the relationship between the precession rate, the radius of curvature, and forward speeds. As the radius of curvature remains nearly constant at low speeds (refer figure \ref{fig:RadiusvsForspeed}), we observe a linear increase in the value of the mean precession rate as the speed increases. However, as the radius of curvature increases at a nonlinear rate at high speeds, the mean precession rate decreases as the speed increases linearly. Figure \ref{fig:Precvsforspeed} also illustrates the similarity between the system response for the mean precession rate generated by the original model and the simplified model equation (\ref{smalltheta_phid}) at different speeds.  

    The relationship between the mean precession rate and pendulum angle for a range of constant forward speeds is shown graphically in figure \ref{fig:Precvspendangle}, which is based on equation (\ref{smalltheta_phid}). The figure illustrates that the mean precession rate is directly proportional to the pendulum angle. Contrarily, the mean precession rate does not increase or decrease monotonically with speed for a given pendulum angle.
    
\end{description}

 \section{Controller design} \label{controller}

The spherical robot can be commanded to execute specific maneuvers such as turning by controlling the value of pendulum angle $\beta$, or forward speed, by controlling $\dot{\psi}$, in a teleoperation setup~\cite{aca_paper}. Controlling the robot's speed and pendulum angle, as discussed in section \ref{model_simplification_results}, results in indirect control over the radius of curvature as per equation (\ref{radiusOfCurv}). It is evident from section \ref{og_system_circle_response} that the robot does not move in an exact circular arc but rather wobbles. Therefore, it is necessary to stabilize this wobbly behavior, which is characterized by the oscillations of the lean angle $\theta$. In this section, we present a controller design for semi-autonomous robot operation to achieve the following control objectives: 

\begin{enumerate} 
    \item Heading control: Moving the robot along the desired curve or line by changing the turning radius
    \item Wobble control: Limiting lateral oscillations to obtain sharp feedback from the robot's camera 
\end{enumerate}

For the robot's lateral oscillations to stabilize, the lean angle $\theta$ must remain fixed ($\dot{\theta}_{des}$ = 0). To achieve heading control, the desired value of the pendulum's angle $\beta_{des}$ would be determined based on the required radius of curvature for a given operational speed $\dot{\psi}$. The robot's $\dot{\psi}_{des}$ will be set according to the requirements of the teleoperator. For example, if careful surveillance is required, the teleoperator may need to move the robot slowly along a path. Rapidly reaching a destination may necessitate traveling at a faster rate. It should be noted that $\beta_{des}$ should be small due to space constraints and the need to maintain the robot's upright position for enhanced camera coverage. The following sections describe the controller's strategy for achieving these objectives and the results it achieved.

\subsection{Approach}

With only two inputs, $T_s$ and $T_p$, there are only two controllable quantities. Therefore, we can choose between the following outputs, which are related to forward motion (speed) and steering (pendulum angle/lean angle rate):

\begin{equation}
\mathbf{y} = 
\begin{bmatrix}
y_1 \\ y_2
\end{bmatrix} = 
\begin{bmatrix}
\dot{\psi} \\ \beta
\end{bmatrix} \text{    or    } \mathbf{y} = 
\begin{bmatrix}
y_1 \\ y_3
\end{bmatrix} = 
\begin{bmatrix}
\dot{\psi} \\\dot{\theta}
\end{bmatrix}
\end{equation}

Similar to input-output feedback linearization approach, we repeatedly differentiate the potential output functions $y_1$, $y_2$ and $y_3$ until the input \textbf{u} appears to obtain: 

\begin{equation}
\begin{bmatrix}
\dot{y}_1 \\ \ddot{y}_2 \\ \dot{y}_3
\end{bmatrix} = 
\begin{bmatrix}
\ddot{\psi} \\ \ddot{\beta} \\ \ddot{\theta}
\end{bmatrix} = 
\begin{bmatrix}
f_{9}(\mathbf{x}) \\ f_{10}(\mathbf{x}) \\ f_{8}(\mathbf{x}) 
\end{bmatrix} + 
\begin{bmatrix}
G_{9,1}(\mathbf{x}) & 0 \\ 0 & G_{10,2}(\mathbf{x}) \\ 0 & G_{8,2}(\mathbf{x})
\end{bmatrix}
\begin{bmatrix}
T_s \\ T_p
\end{bmatrix}
\label{three_outputs}
\end{equation}

It can be observed from equation (\ref{three_outputs}) that $\ddot{\psi}$ depends only on $T_s$. Whereas $\ddot{\beta}$ and $\ddot{\theta}$ depend only on $T_p$. 

As input $T_s$ is exclusive to a particular output $\ddot{\psi}$, we design the speed control torque ($T_s$) separately using a proportional controller with high gain $K_{p,\dot{\psi}}$ as shown:

\begin{equation}
T_s = K_{p,\dot{\psi}} (\dot{\psi}_{des} - \dot{\psi})    
\end{equation}

To control pendulum angle and wobble, the controller input $T_p$ is designed as a linear combination of the required torques for $\beta$ and $\dot{\theta}$ control. Let $T_{p,\beta}$ be the pendulum control torque required to maintain a $\beta_{des}$. Let $T_{p,\dot{\theta}}$ be the wobble control torque required to maintain $\dot{\theta}_{des}=0$. Then a net torque $T_p$ can be found as a weighted sum of wobble control torque ($T_{p,\dot{\theta}}$) and pendulum control torque ($T_{p,\beta}$) as shown:

\begin{equation}
T_p = \gamma T_{p,\dot{\theta}} + \delta T_{p,\beta}
\label{comboTp}
\end{equation}

\begin{description} [before={\renewcommand\makelabel[1]{\normalfont ##1}}]

    \item[Wobble control torque ($T_{p,\dot{\theta}}$) design:]  We implement the underlying concepts of input-output feedback linearization via static feedback to cancel nonlinear terms appearing in $\dot{y}_3$ and then apply a proportional controller term $v_{\theta}$. 

\begin{equation}
T_{p,\dot{\theta}} = \Bigg[\frac{1}{G_{8,2}(\mathbf{x})}( - f_{8}(\mathbf{x}) + v_{\theta})\Bigg]
\text{ where } v_{\theta} = K_{p,\dot{\theta}} (\dot{\theta}_{des} - \dot{\theta})
\end{equation}

    \item[Pendulum control torque ($T_{p,\beta}$) design:] We implement a combination of feedback term $v_{\beta}$ and feedforward term to counteract gravitational torque. $v_{\beta}$ consists of a proportional-derivative (PD) controller with high gains $K_{p,\beta}$ and $K_{d,\beta}$ as shown: 

\begin{equation}
T_{p,\beta} = [m_p g r_p \sin(\beta+\theta) + v_{\beta}]
\text{ where } v_{\beta} = K_{p,\beta} (\beta_{des} - \beta) + K_{d,\beta} (\dot{\beta}_{des} - \dot{\beta})
\end{equation}

\end{description}

\begin{figure}
\centering
\includegraphics[width=\linewidth]{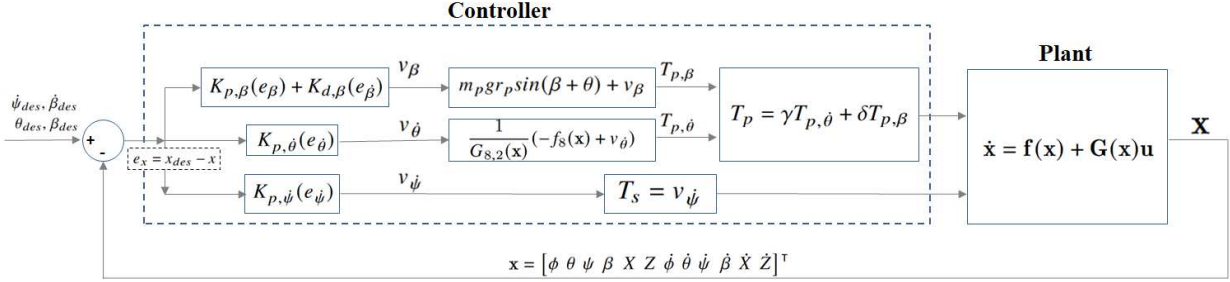}
\caption{Block diagram for controller design}
\label{fig:blockDiagram}
\end{figure}

The proposed controller architecture is depicted in a block diagram shown in  figure \ref{fig:blockDiagram}. We empirically tune the speed control gain ($K_{p,\dot{\psi}}$) and the pendulum control gains ($K_{p,\beta}$ and $K_{d,\beta}$) to very large values to emulate our robot's actuation system, which has a lower level controller that controls the angular position of the pendulum and the speed of the motor driving the robot forward almost instantaneously. We also empirically tune the wobble control gain ($K_{p,\dot{\theta}}$). However, the wobble control gain ($K_{p,\dot{\theta}}$) has a much smaller value than the speed control gain ($K_{p,\dot{\psi}}$) and the pendulum control gains ($K_{p,\beta}$ and $K_{d,\beta}$). This is because the proportional controller term ($v_{\theta}$) in wobble control torque ($T_{p,\dot{\theta}}$) is not responsible for canceling out the nonlinear terms associated with $\theta$'s dynamics. In the case of wobble control torque ($T_{p,\dot{\theta}}$), input-output linearization cancels out the nonlinear terms, whereas this is not the case with pendulum control torque ($T_{p,\beta}$) or speed control torque ($T_s$).


To tune the $\gamma$ and $\delta$ values, the $T_p$ controller is implemented for three distinct scenarios, and the system response is analyzed. First, we examine the performance at the extremes by disabling pendulum control torque ($T_{p,\beta}$) or wobble control torque ($T_{p,\dot{\theta}}$). This is accomplished by setting ($\gamma$=0, $\delta$=1) and ($\gamma$=1, $\delta$=0) respectively. Next, we test the performance of $T_p$ controller with various values of ($\gamma$,$\delta$), with each value ranging from 0 to 1, to obtain various linear combinations of pendulum control torque ($T_{p,\beta}$) and wobble control torque ($T_{p,\dot{\theta}}$).

We analyze the performance of these three types of scenarios to determine which configuration of ($\gamma$,$\delta$) achieves the goal of moving the robot in a circle at a constant speed and a desired radius of curvature while simultaneously eliminating wobbling. Before implementing the proposed controller with the scenarios discussed previously, we move the robot in a wobbly circle at a steady state for the first five seconds, similar to section \ref{analysisOfWobblyCircle}. The value of the radius of curvature for this motion can be determined using the equation (\ref{radiusOfCurv}) corresponding to the initial configuration of a pendulum angle of 15$^{\circ}$ and a forward speed |$\dot{\psi}$| of 1. The control response of the three discussed scenarios is summarised below, along with figures comparing the system responses generated by each case:

\begin{figure}
    \centering 
\begin{subfigure}{0.3\textwidth}
  \includegraphics[width=\linewidth, trim = 1cm 0 1cm 0]{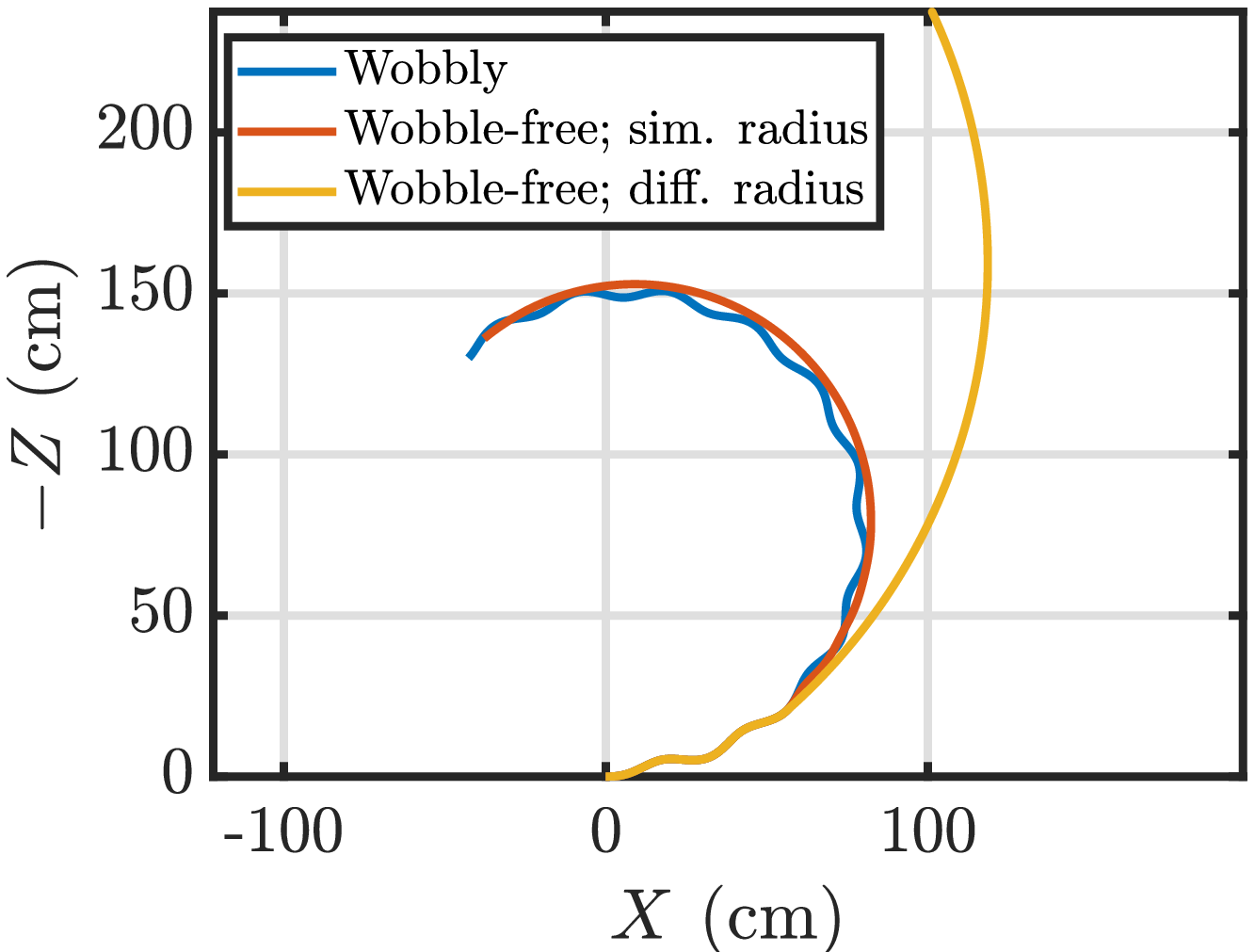}
  \caption{\centering{Robot's path}}
  \label{fig:MultiControlCompTraj_15deg}
\end{subfigure}\hfil 
\begin{subfigure}{0.3\textwidth}
  \includegraphics[width=\linewidth, trim = 1cm 0 1cm 0]{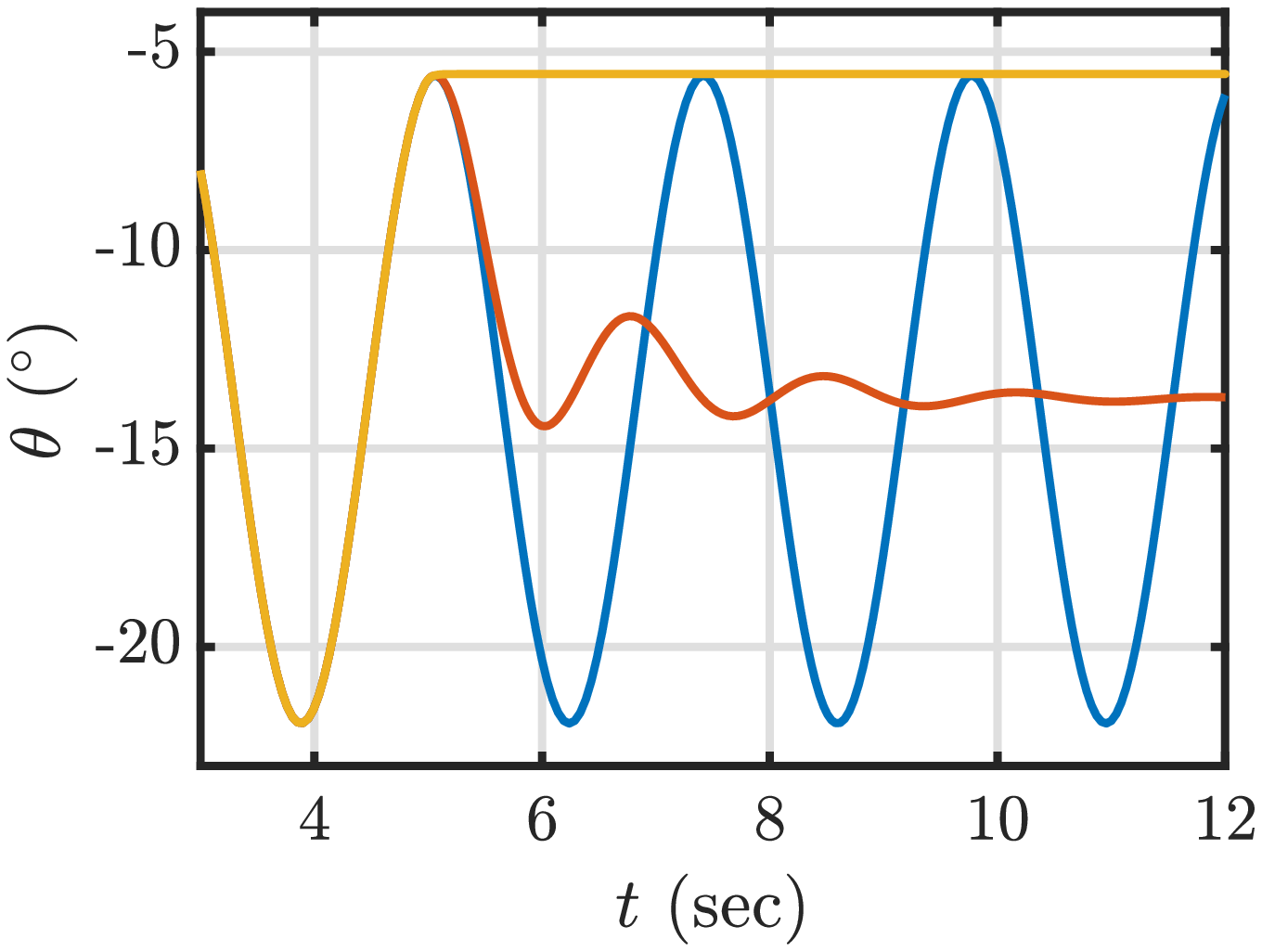}
  \caption{\centering{Wobbling vs. time}}
  \label{fig:MultiControlCompTheta_15deg}
\end{subfigure}

\medskip
\begin{subfigure}{0.3\textwidth}
  \includegraphics[width=\linewidth, trim = 1cm 0 1cm 0]{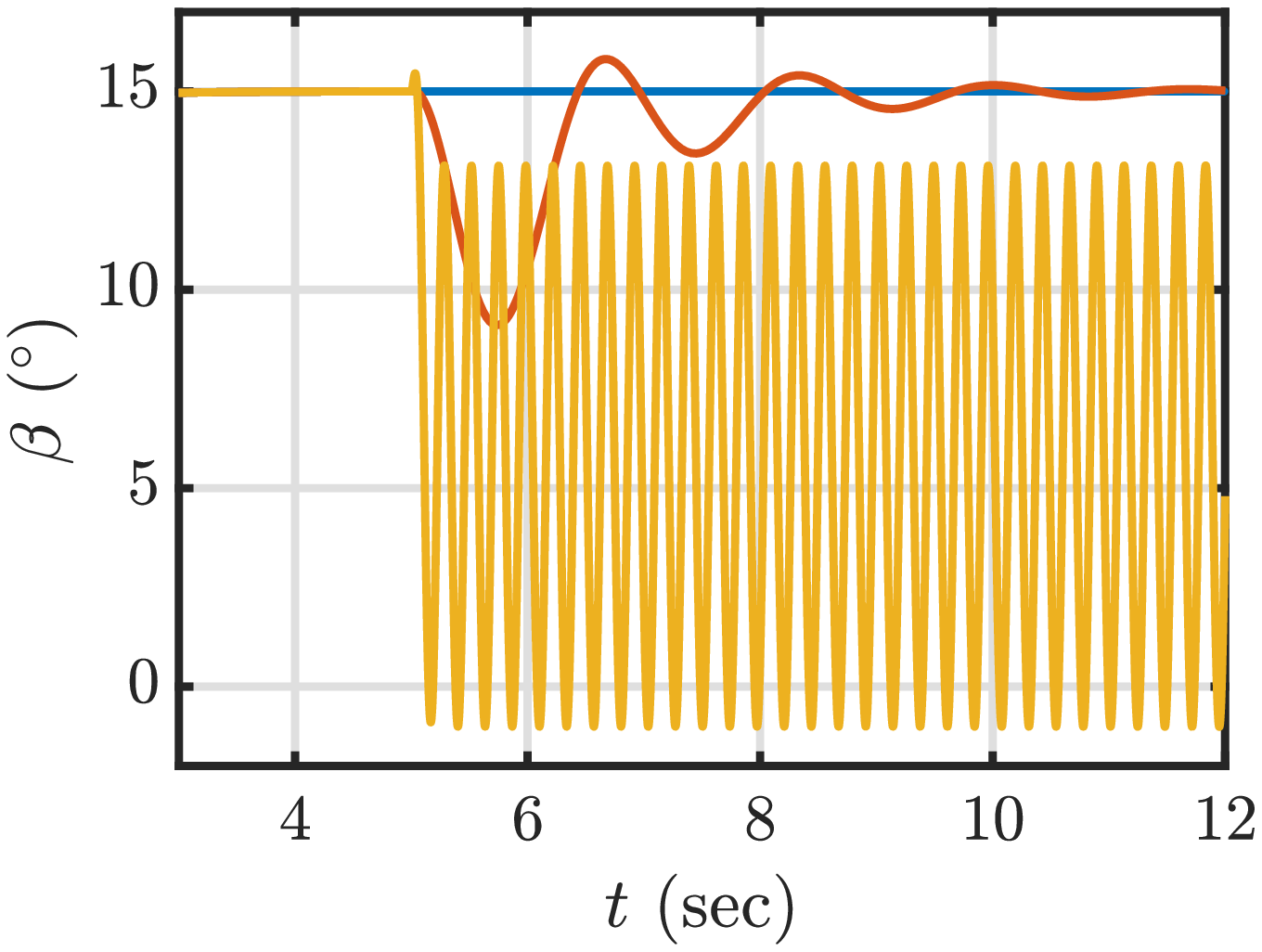}
  \caption{\centering{Pendulum Angle vs. time}}
  \label{fig:MultiControlCompBeta_15deg}
\end{subfigure}\hfil 
\begin{subfigure}{0.3\textwidth}
  \includegraphics[width=\linewidth, trim = 1cm 0 1cm 0]{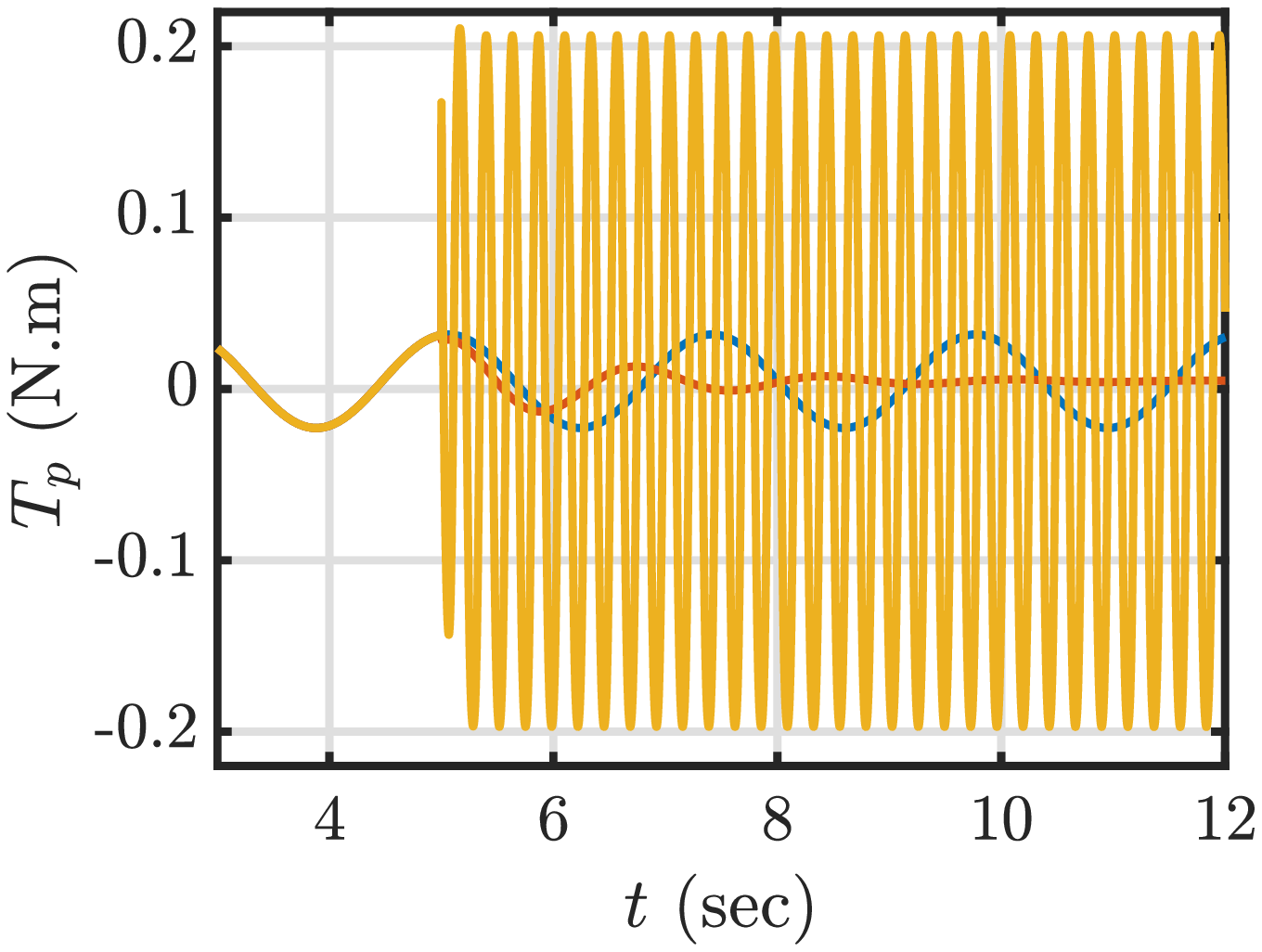}
  \caption{\centering{Pendulum torque vs. time}}
\end{subfigure}

\caption{System response by choosing different linear combinations of $\gamma T_{p,\dot{\theta}} + \delta T_{p,\beta}$ during a circular motion with $\beta_{des}$ = 15$^{\circ}$} 
\label{fig:multiCircle}
\end{figure}

\begin{description}
    \item[Wobbly] Wobbly behavior with desired radius of curvature: ($\gamma$ = 0, $\delta$ = 1).
    
    In this case, only the pendulum control torque ($T_{p,\beta}$) is at work to control the $\beta$. Though the robot moves in a desired radius of curvature, it wobbles along the way. Hence, the lean angle $\theta$ exhibits oscillations as expected due to wobbling as shown in figure \ref{fig:MultiControlCompTheta_15deg}. Figure \ref{fig:MultiControlCompBeta_15deg} depicts that the pendulum angle $\beta$ remains stabilized at 15$^{\circ}$ as per the command. 
    \item[Wobble-free; diff. radius] Wobble-free behavior with a different radius of curvature: ($\gamma$ = 1, $\delta$ = 0).
    
    Here, we toggle the controller to switch to the wobble control torque ($T_{p,\dot{\theta}}$), which is responsible for reducing the wobbling. This configuration successfully stabilizes the lean angle $\theta$ as depicted in figure \ref{fig:MultiControlCompTheta_15deg}. However, the pendulum angle $\beta$ starts exhibiting very high-frequency oscillations that can be observed in figure \ref{fig:MultiControlCompBeta_15deg}. Figure \ref{fig:MultiControlCompTraj_15deg} shows that the robot loses track of the path it was supposed to move and starts moving on a curve with a higher radius of curvature because the mean value of pendulum angle $\beta$ decreases.
    \item[Wobble-free; sim. radius] Wobble-free behavior with a similar radius of curvature: ($\gamma$ = 0.9, $\delta$ = 0.1)
    
    This case captures the best of both worlds by using both the components pendulum control torque ($T_{p,\beta}$) and wobble control torque ($T_{p,\dot{\theta}}$) serving different functions. The robot moves in a path with a desired curvature radius as shown in \ref{fig:MultiControlCompTraj_15deg}. The lean angle $\theta$ and $\beta$ stabilize with time as shown in figures \ref{fig:MultiControlCompTheta_15deg} and \ref{fig:MultiControlCompBeta_15deg} respectively. 
\end{description}

\subsection{Results}

\begin{figure}
    \centering 
\begin{subfigure}{0.3\textwidth}
  \includegraphics[width=\linewidth, trim = 1cm 0 1cm 0]{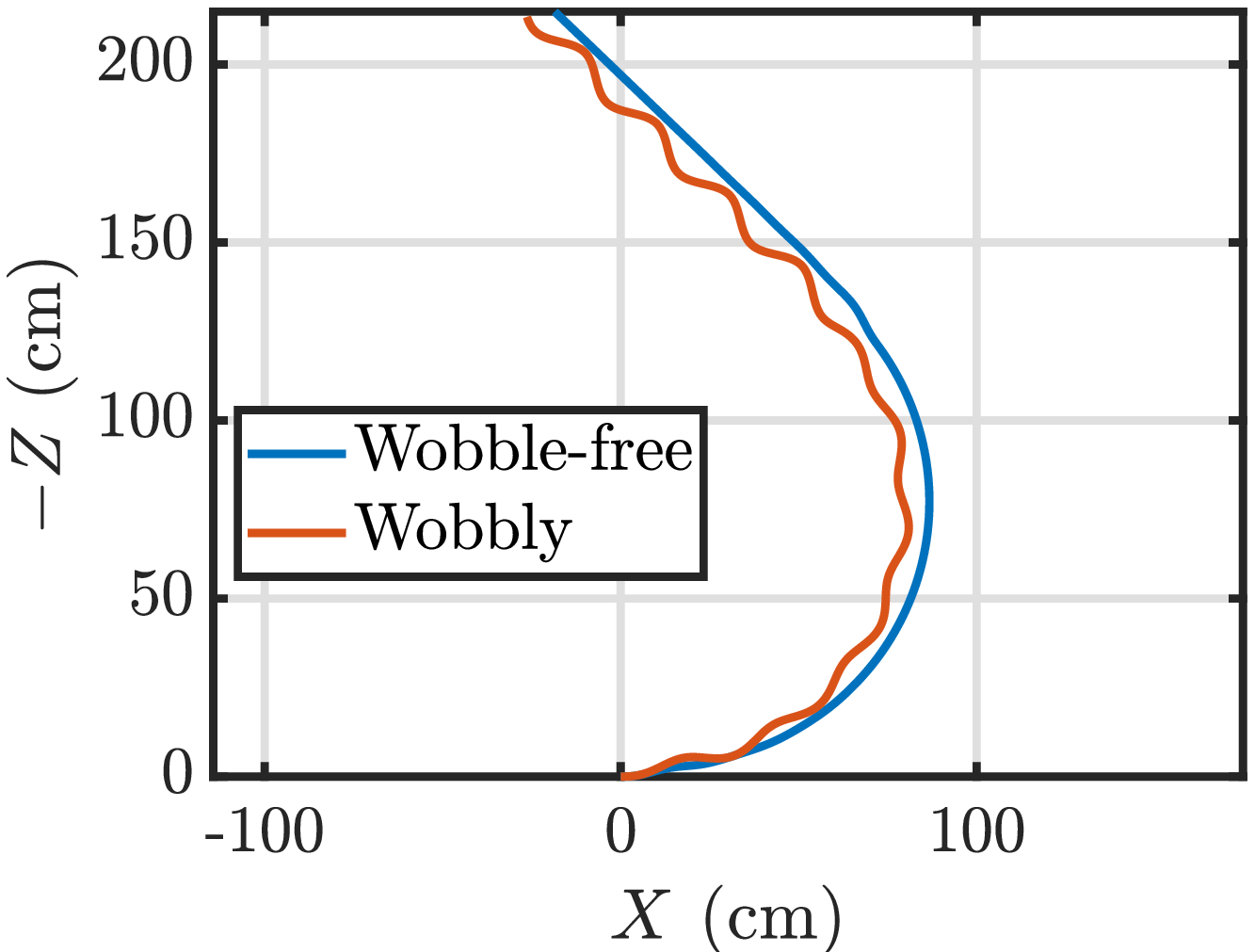}
  \caption{\centering{Robot's path}}
  \label{fig:controlCompTraj_15deg}
\end{subfigure}\hfil 
\begin{subfigure}{0.3\textwidth}
  \includegraphics[width=\linewidth, trim = 1cm 0 1cm 0]{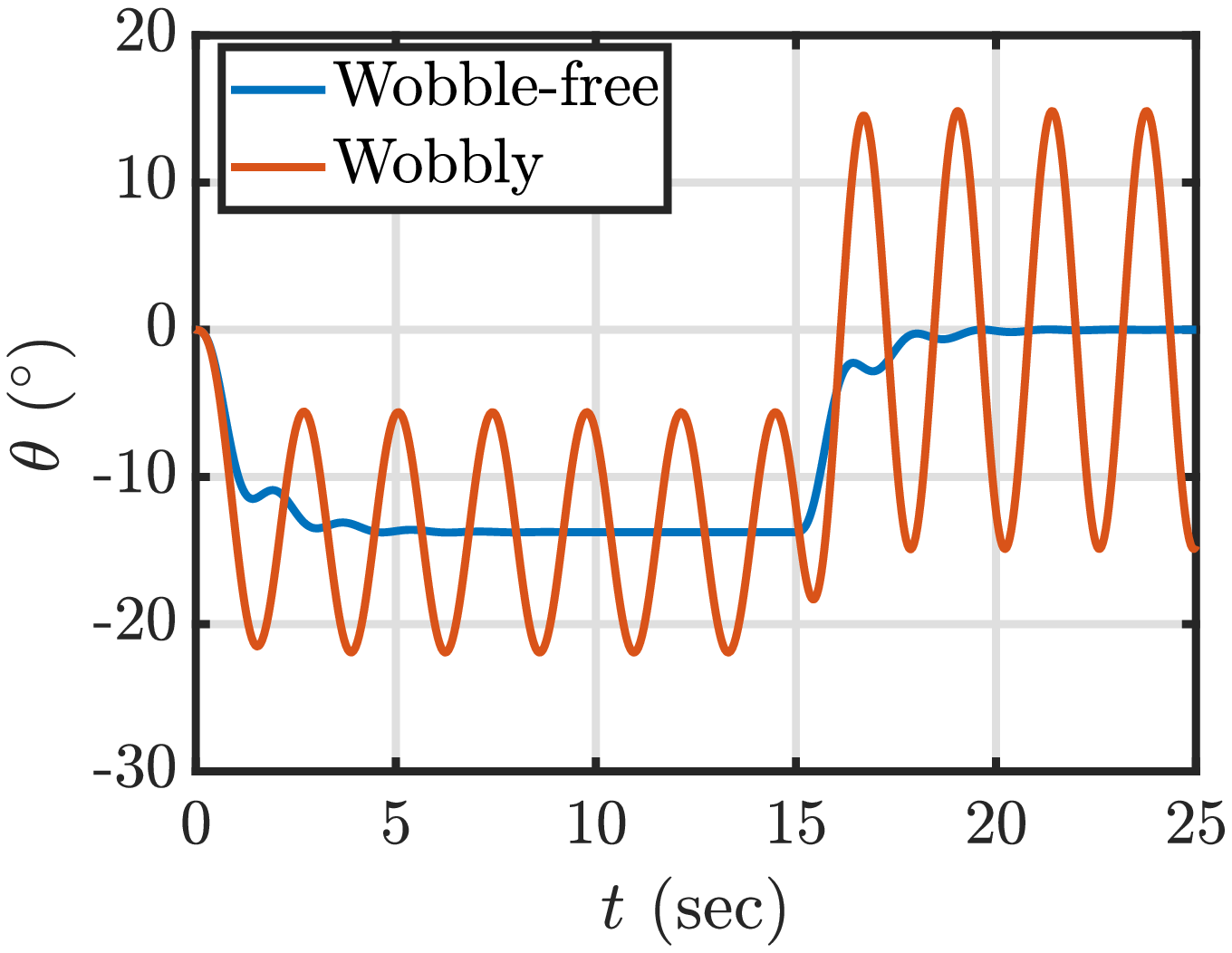}
  \caption{\centering{Wobbling vs. time}}
  \label{fig:controlCompTheta_15deg}
\end{subfigure}\hfil 

\medskip
\begin{subfigure}{0.3\textwidth}
  \includegraphics[width=\linewidth, trim = 1cm 0 1cm 0]{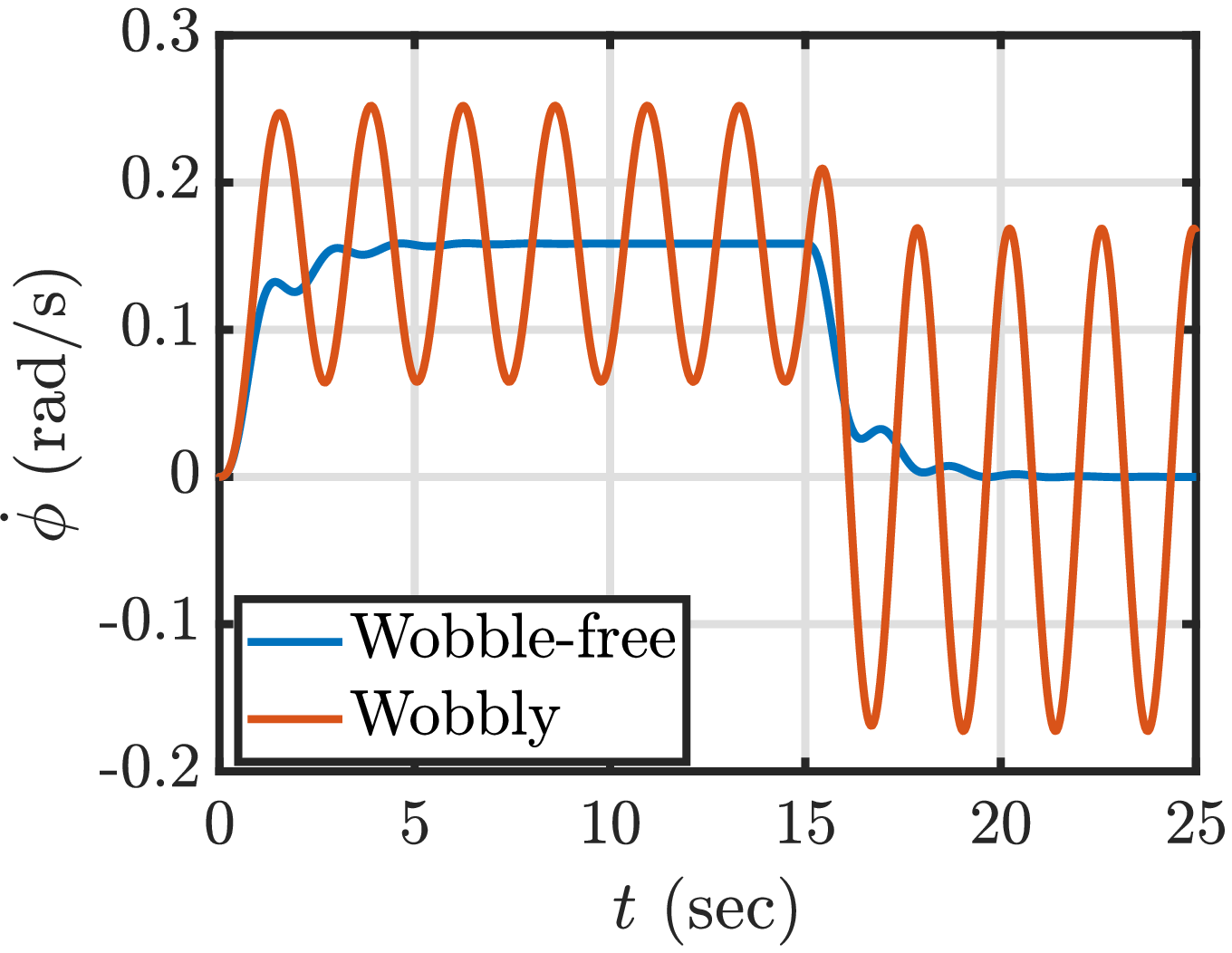}
  \caption{\centering{Precession rate vs. time}}
  \label{fig:controlCompPhid_15deg}
\end{subfigure}\hfil 
%
\begin{subfigure}{0.3\textwidth}
  \includegraphics[width=\linewidth, trim = 1cm 0 1cm 0]{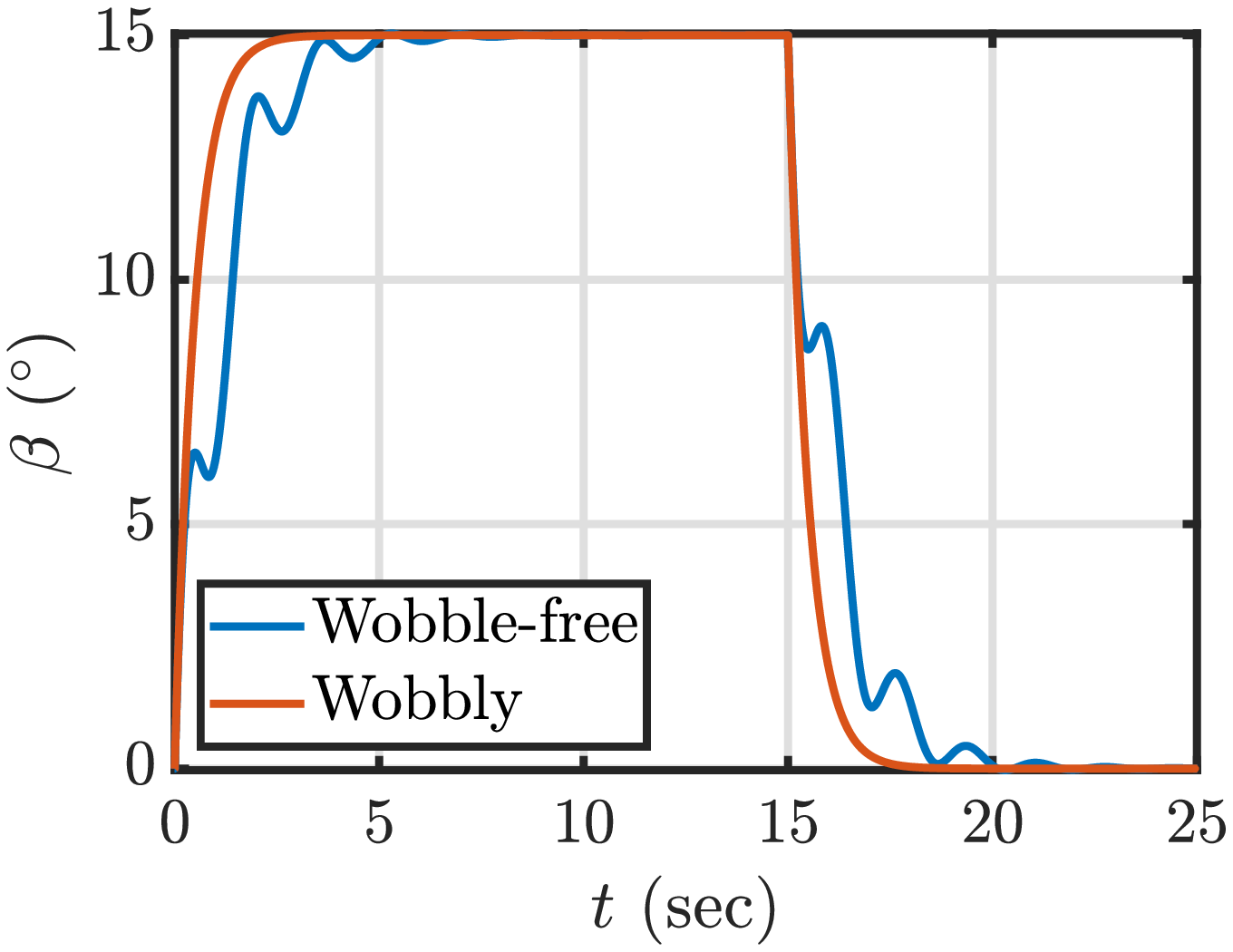}
  \caption{\centering{{\tabular[t]{@{}l@{}}Pendulum Angle \\ vs. time\endtabular}}}
  \label{fig:controlCompBeta_15deg} 
\end{subfigure}\hfil 

\caption{Comparison between wobbly and wobble-free motion when $\beta_{des}$ = 15$^{\circ}$ during turning motion} 
\label{fig:TurningMotion}
\end{figure}

We use the $T_p$'s linear combination ($\gamma$ = 0.9, $\delta$ = 0.1) to execute a wobble-free turning maneuver as shown in figure \ref{fig:TurningMotion}. We compare the wobble-free system response to a wobbly turning maneuver in which only the pendulum control torque ($T_{p,\beta}$) is used for pendulum angle control \footnote{The video at \url{https://youtu.be/430yfKLEphw} demonstrates the 3D multibody dynamic simulation of the pendulum-actuated spherical robot generated in Simulink using the VRML (Virtual Reality Modeling Language) functionality. We illustrate the controller design results in the video by contrasting the robot's wobbly and wobble-free turning maneuvers modeled in this study.}.

Figure \ref{fig:controlCompTheta_15deg} demonstrates that the wobbling has decreased significantly due to a negligible change in the lean angle $\theta$ when the linear combination-based controller ($T_p$) is used. A non-zero stabilization of $\theta$ during the turning phase causes the robot to tilt toward the turn's center. Then, $\theta$ stabilizes at 0$^{\circ}$, indicating that the robot regains its upright position following the completion of its turning motion.

As depicted in figure \ref{fig:controlCompPhid_15deg}, the linear combination-based controller ($T_p$) can also stabilize the precession rate. During the period in which the robot executes a turning motion, the figures display a non-zero stable precession rate. As soon as the robot moves in a straight line, the precession rate becomes zero.

The evolution of pendulum angle $\beta$ with time depicted in figure \ref{fig:controlCompBeta_15deg} is also realizable in practice. It is neither oscillatory nor increasing or decreasing monotonically. It is fascinating that $\beta$ exhibits a smoother response during the robot's wobbly motion. However, when the robot executes a wobble-free motion due to the linear combination-based controller ($T_p$), the resulting $\beta$ response exhibits oscillations in its behavior during the transient phase, which eventually stabilizes as the robot reaches a steady state.

\begin{figure}
    \centering 
\begin{subfigure}{0.26\textwidth}
  \includegraphics[width=\linewidth, trim = 1cm 0 1cm 0]{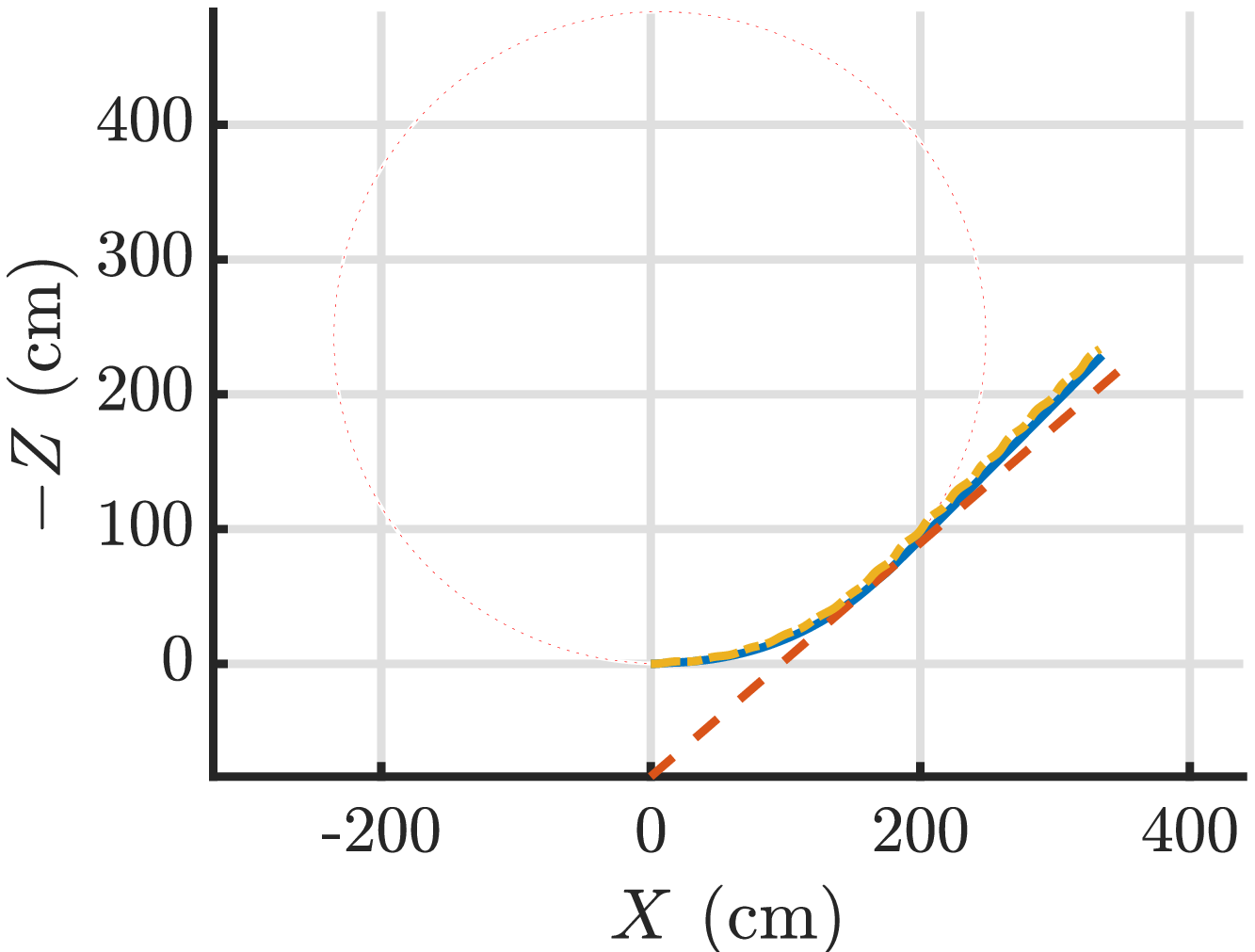}
  \caption{\centering{$\beta_{des}$ = 5$^{\circ}$}}
  \label{fig:turnAnalysis_5deg}
\end{subfigure}\hfil 
\begin{subfigure}{0.26\textwidth}
  \includegraphics[width=\linewidth, trim = 1cm 0 1cm 0]{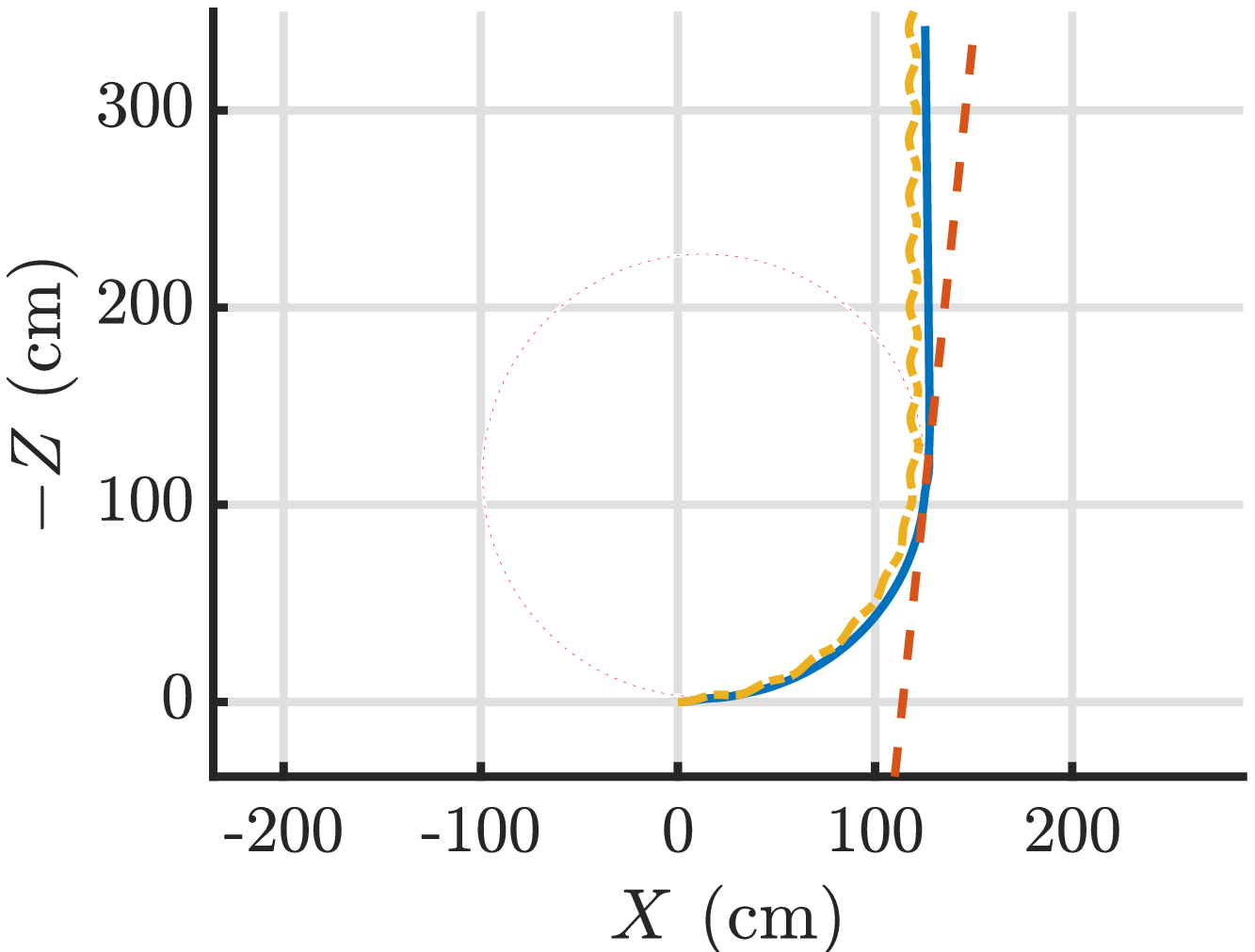}
  \caption{\centering{$\beta_{des}$ = 10$^{\circ}$}}
  \label{fig:turnAnalysis_10deg}
\end{subfigure}\hfil 
\begin{subfigure}{0.26\textwidth}
  \includegraphics[width=\linewidth, trim = 1cm 0 1cm 0]{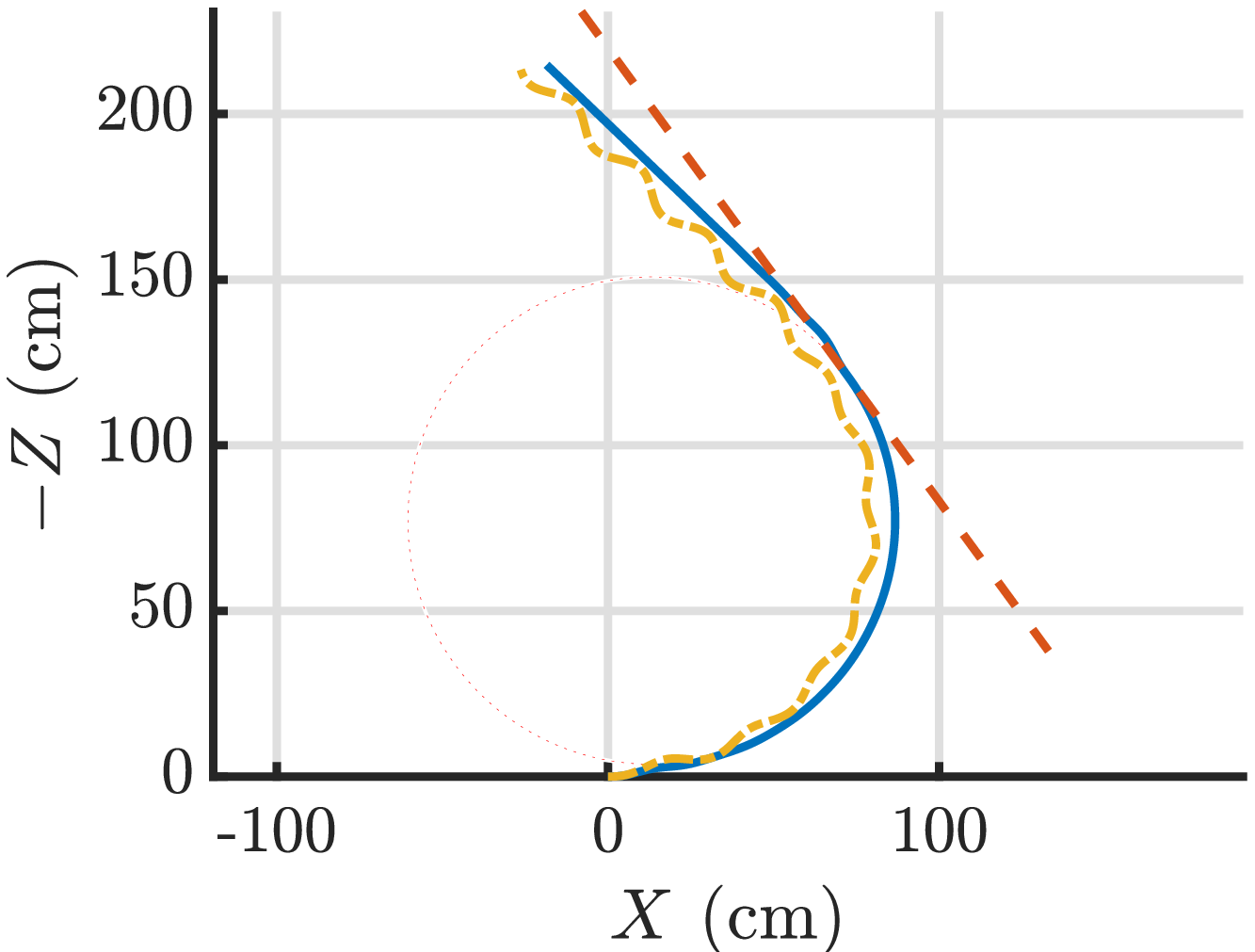}
  \caption{\centering{$\beta_{des}$ = 15$^{\circ}$}}
  \label{fig:turnAnalysis_15deg}
\end{subfigure}\hfil 

\medskip
\begin{subfigure}{0.26\textwidth}
  \includegraphics[width=\linewidth, trim = 1cm 0 1cm 0]{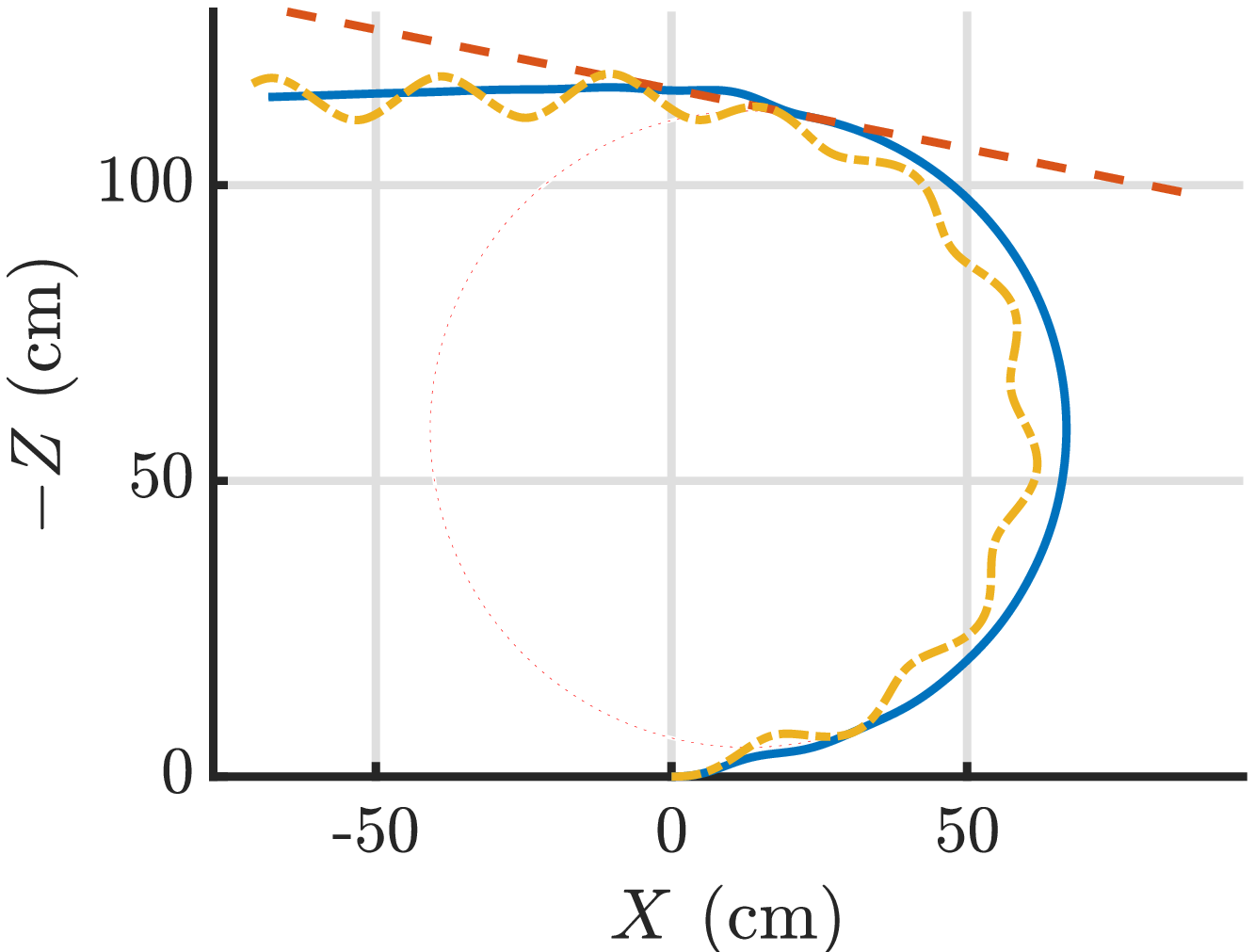}
  \caption{\centering{$\beta_{des}$ = 20$^{\circ}$}}
  \label{fig:turnAnalysis_20deg}
\end{subfigure}\hfil 
\begin{subfigure}{0.26\textwidth}
  \includegraphics[width=\linewidth, trim = 1cm 0 1cm 0]{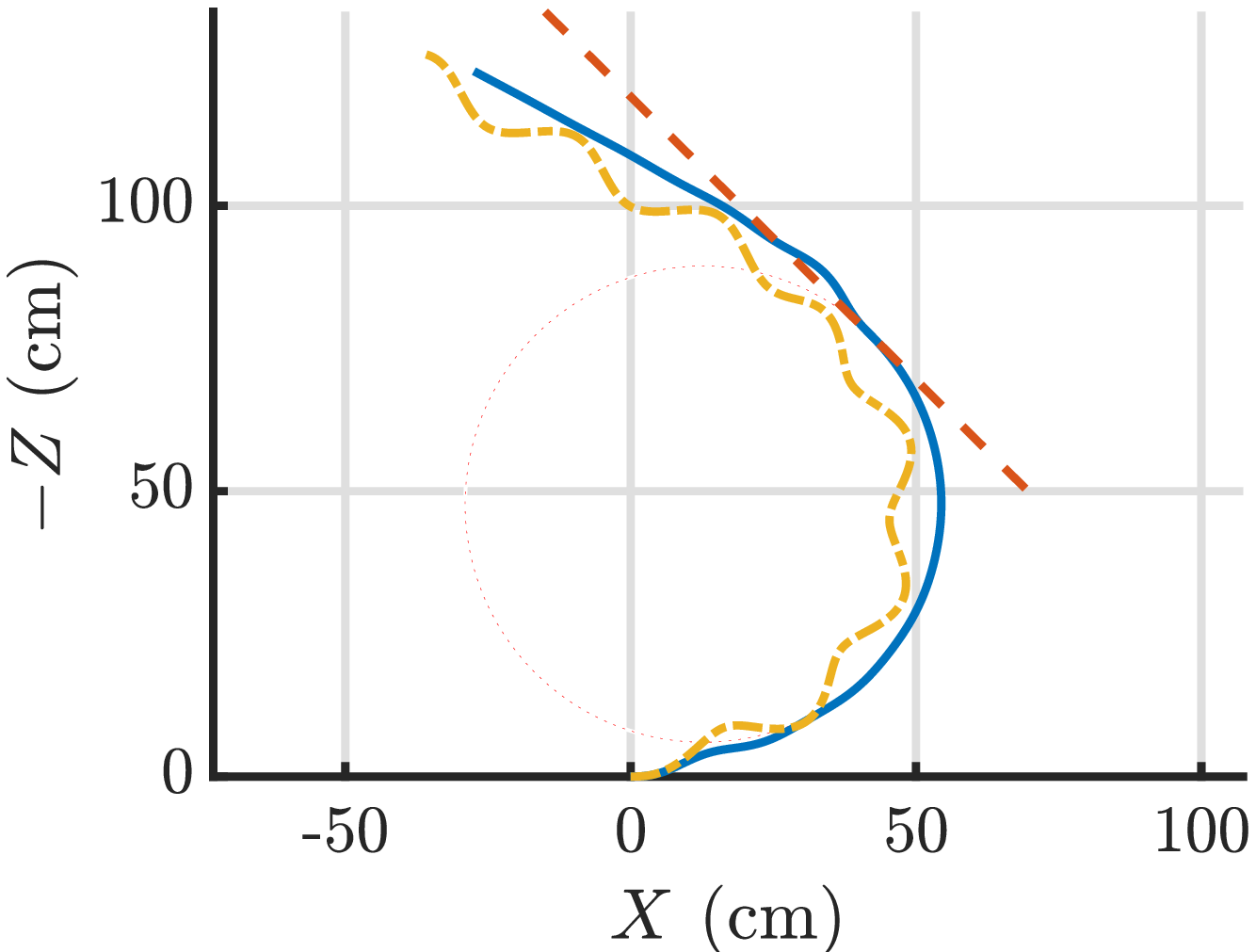}
  \caption{\centering{$\beta_{des}$ = 25$^{\circ}$}}
  \label{fig:turnAnalysis_25deg}
\end{subfigure}\hfil 
\begin{subfigure}{0.26\textwidth}
  \includegraphics[width=\linewidth, trim = 1cm 0 1cm 0]{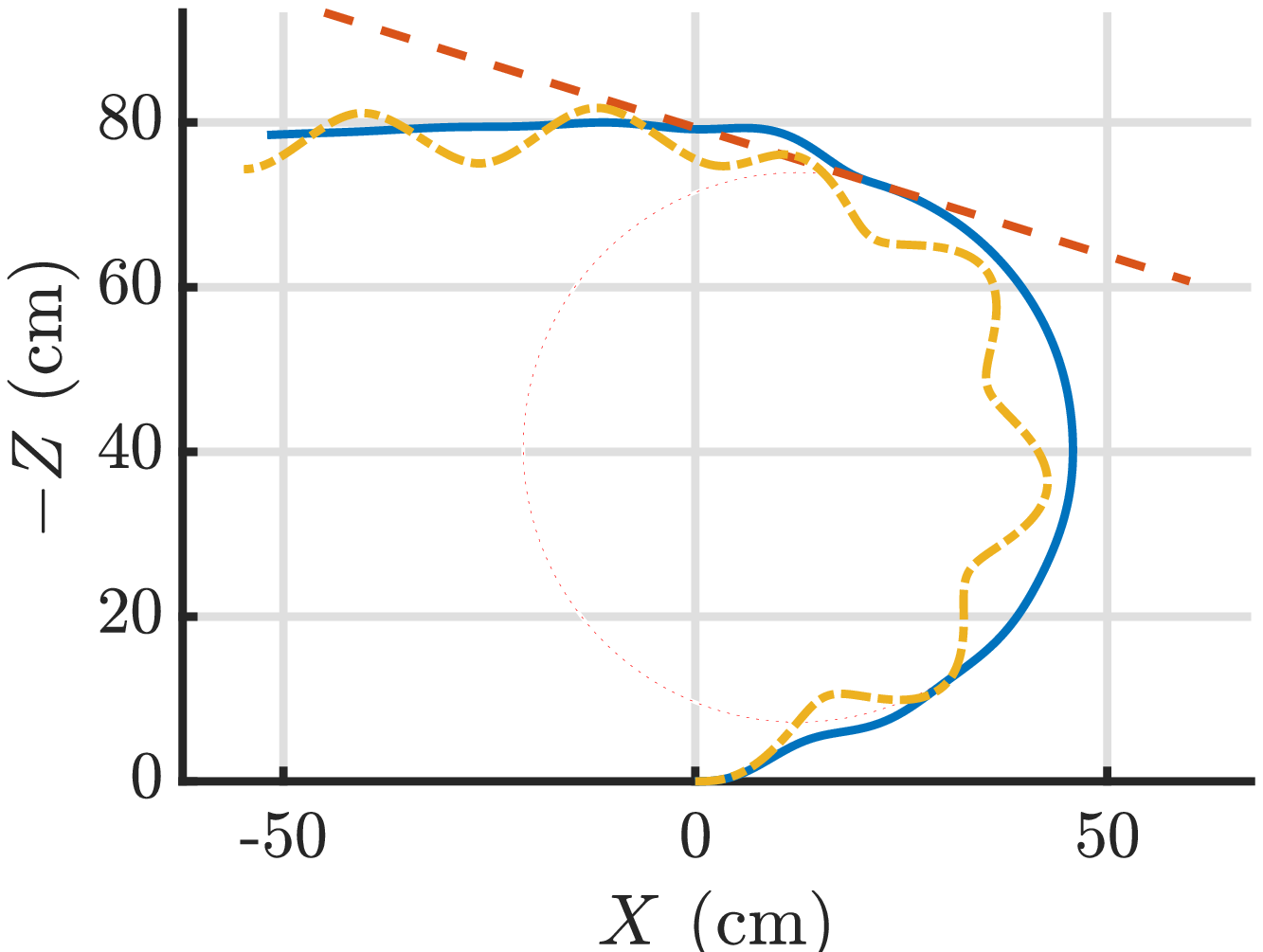}
  \caption{\centering{$\beta_{des}$ = 30$^{\circ}$}}
  \label{fig:turnAnalysis_30deg}
\end{subfigure}\hfil 

\caption{Comparison between the path taken by the robot during wobbly and wobble-free turning motion while moving with different value of pendulum angle during the circular arc} 
\label{fig:wobbleFreeTurningResponseComparision}
\end{figure}

Figure \ref{fig:wobbleFreeTurningResponseComparision} illustrates a comparison between wobbly and wobble-free turning maneuvers with the proposed controller for varying pendulum angle $\beta$ values at forward speed $\dot{\psi}$ of 1 rad/s. The solid line style represents motion without wobble, whereas the dash-dotted line style represents motion with wobble. During the circular arc section of the path, we can observe that the radius of curvature for these two types of response varies slightly. After the turning maneuver, both trajectories converge to move in the same direction.

\begin{figure}
    \centering 
\begin{subfigure}{0.3\textwidth}
  \includegraphics[width=\linewidth, trim = 1cm 0 1cm 0]{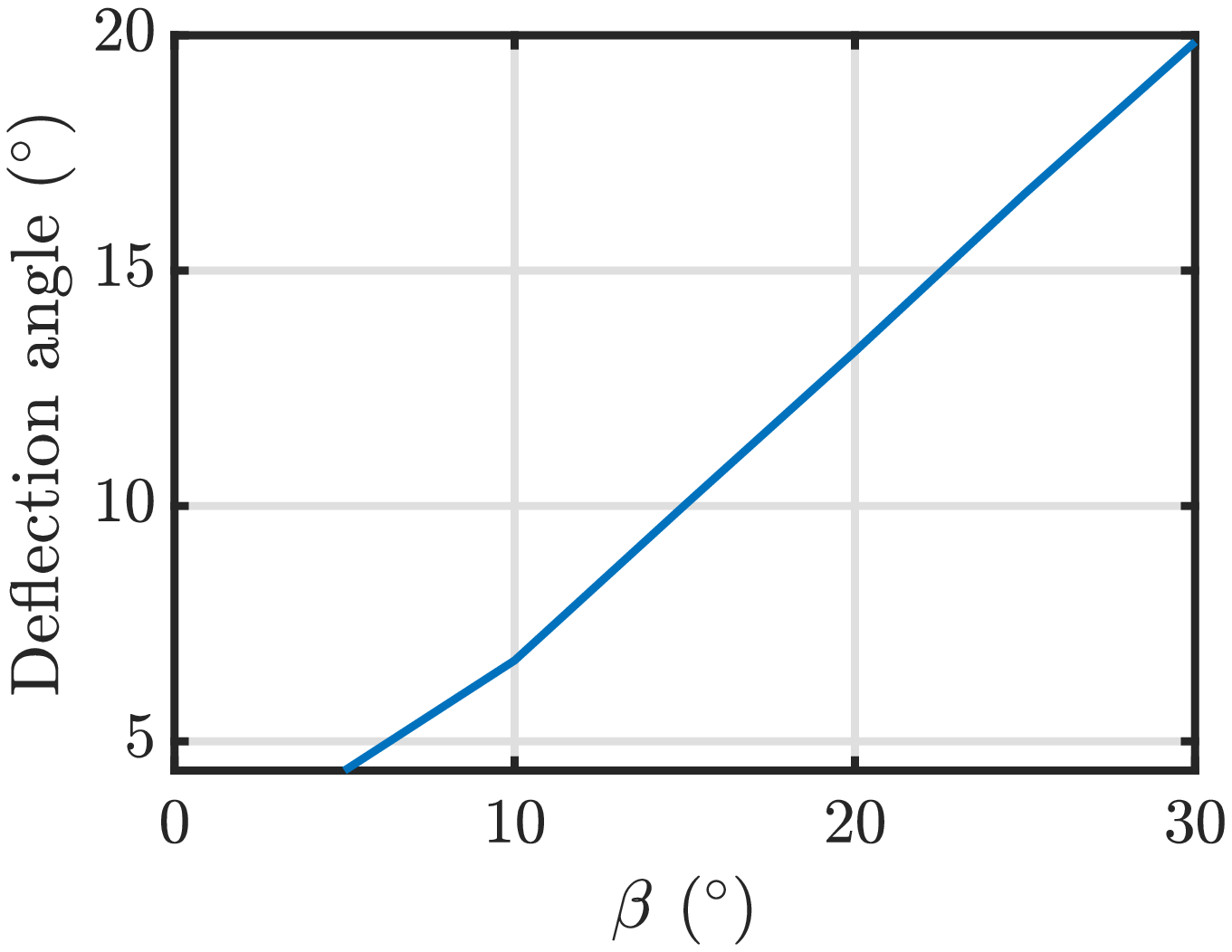}
  \caption{\centering{Deviation from the tangent at the end of circular arc in turning motion}}
  \label{fig:heading_err_vs_beta}
\end{subfigure}\hfil 
\begin{subfigure}{0.3\textwidth}
  \includegraphics[width=\linewidth, trim = 1cm 0 1cm 0]{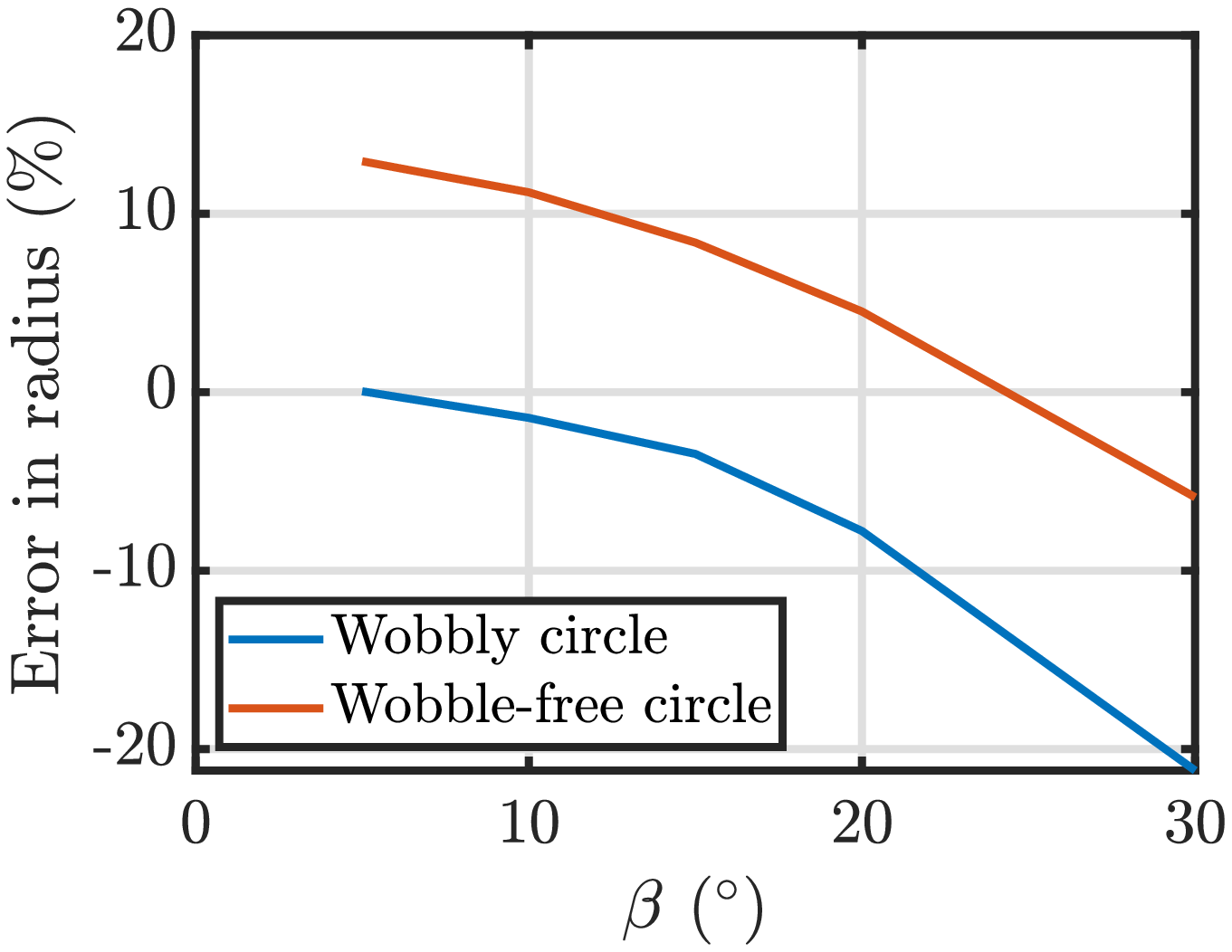}
  \caption{\centering{Error in radius of curvature observed in turning motion compared to values obtained from equation (\ref{radiusOfCurv})}}
  \label{fig:radius_err_vs_beta}
\end{subfigure}

\caption{Errors as a function of $\beta$} 
\label{fig:errorTrends}
\end{figure}

The dotted line style represents the circle that the wobble-free turning maneuver is a part of while tracing the circular arc. The dashed line represents the tangent to this circle at the point where $\beta_{des}$ is toggled to 0$^{\circ}$. Due to the settling time associated with this toggle command, the tangent at the toggle point is not parallel to the direction of the final straight line for both wobbly and wobble-free straight lines. This deflection is shown as a function of pendulum angle $\beta$ in figure \ref{fig:heading_err_vs_beta}. Observably, the deflection increases as the angle of the pendulum increases.

At a given speed of operation, the radius of curvature during a turning maneuver can be approximated using the equation (\ref{radiusOfCurv}) for various values of pendulum angle $\beta$. The error in the radius of curvature is depicted in \ref{fig:radius_err_vs_beta} as the percentage difference between the observed values and the desired values predicted by equation (\ref{radiusOfCurv}). For both wobbly and wobble-free motion, the percentage error in the radius of curvature relative to the desired value decreases as the pendulum angle increases. In conclusion, when the inclination of the pendulum is altered, the heading angle deflection and error in the radius of curvature exhibit opposite behaviors. Based on heading deflection tolerance and acceptable error in the radius of curvature, figures \ref{fig:heading_err_vs_beta} and \ref{fig:radius_err_vs_beta} can be used to choose a pendulum angle that corresponds to the desired operational speed.

\section{Conclusion and future work} \label{conclusion}

This article proposes a dynamic model of a pendulum-actuated spherical robot considering the coupling between forward and steering motion. Our model qualitatively captures the system's lateral oscillations, which we refer to as wobbling. Modeling the wobble allows for the investigation of controller design for stabilizing the yoke, which serves as a platform for mounting sensors on a spherical robot. The yoke must remain stable during motion because the sensors mounted on the yoke would otherwise produce inaccurate results. In addition, we present mathematical formulations for the robot's radius of curvature, precession rate, wobble amplitude, and wobble frequency when moving in a wobbly circular motion. Finally, we design a controller for the non-holonomic spherical robot using only two input torques. The controller design aims towards controlling the forward speed $\dot{\psi}$, pendulum angle $\beta$, and limiting wobble (oscillations in the lean angle $\theta$). The control set-points are determined by calculating the desired robot speed and pendulum angle based on the required radius of curvature of the turning maneuver. The results of this work can be applied to the robot's semi-autonomous teleoperation. In this mode, a command is issued for a desired pendulum angle to steer the robot in a specific direction while maintaining a desired forward speed. The wobble controller can assist the robot in moving without wobbling along the desired curve or line. In the future, the results of the controller design can be experimentally validated using actual hardware and used to develop a navigation algorithm. 






\bibliographystyle{unsrt}

\end{document}